\documentclass[journal]{IEEEtran}
\usepackage{amsmath,amsfonts}

\usepackage[linesnumbered,ruled,vlined]{algorithm2e}

\SetCommentSty{mycommfont}
\usepackage{algpseudocode}
\usepackage{amssymb}
\usepackage{colortbl}
\usepackage{array}
\usepackage[caption=false]{subfig}
\usepackage{pstool}
\usepackage{changepage} % for adjustwidth
\usepackage{multirow}
\usepackage[table,dvipsnames]{xcolor}
\usepackage{textcomp}
\usepackage{stfloats}
\usepackage[hyphens]{url}
\usepackage{verbatim}
\usepackage{graphicx}

\usepackage[export]{adjustbox}

\usepackage{cite}
\usepackage[capitalize]{cleveref}
\usepackage{booktabs, makecell}
\usepackage{units}
\newcommand*{\V}{\mathbf}

\colorlet{colorFst}{Green!20}       %
\colorlet{colorSnd}{SpringGreen!45} %
\colorlet{colorTrd}{Yellow!30}      %
\colorlet{colorSep}{blue!5}         %
\newcommand{\fs}{\cellcolor{colorFst}\bf}   %
\newcommand{\nd}{\cellcolor{colorSnd}}      %
\newcommand{\rd}{\cellcolor{colorTrd}}      %

\definecolor{Mypink}{HTML}{A44A78}

\makeatletter
\def\adl@drawiv#1#2#3{%
	\hskip.5\tabcolsep
	\xleaders#3{#2.5\@tempdimb #1{1}#2.5\@tempdimb}%
	#2\z@ plus1fil minus1fil\relax
	\hskip.5\tabcolsep}
\newcommand{\cdashlinelr}[1]{%
	\noalign{\vskip\aboverulesep
		\global\let\@dashdrawstore\adl@draw
		\global\let\adl@draw\adl@drawiv}
	\cdashline{#1}
	\noalign{\global\let\adl@draw\@dashdrawstore
		\vskip\belowrulesep}}
\makeatother

\begin{document}

\title{SceneFactory: A Workflow-centric and Unified Framework for Incremental Scene Modeling}

\author{Yijun Yuan, Michael Bleier, Andreas N\"uchter
\thanks{
Yijun Yuan is with Tsinghua University, Beijing, China. Yijun Yuan, Michael Bleier and Andreas Nüchter are with Julius-Maximilians-Universität Würzburg, Germany. Andreas Nüchter is also with the Zentrum für Telematik e.V., Würzburg and currently International Visiting Chair at U2IS, ENSTA, Institut Polytechnique de Paris, France. The research was done at Julius-Maximilians-Universität Würzburg. Contact: {\tt\small \{yijun.yuan|michael.bleier|andreas.nuechter\}@\\uni-wuerzburg.de}~	

This work was in parts supported by the Federal Ministry for Economic Affairs and Climate Action (BMWK) on the basis of a decision by the German Bundestag und the grant number KK5150104GM1. We also acknowledge the support by the Elite Network Bavaria (ENB) through the ``Satellite Technology'' academic program. 
}
}
  
\maketitle

\begin{abstract}
We present SceneFactory, a workflow-centric and unified framework for incremental scene modeling, that conveniently supports a wide range of applications,
such as (unposed and/or uncalibrated) multi-view depth estimation, LiDAR completion, (dense) RGB-D/RGB-L/Mono/Depth-only reconstruction and SLAM.
The workflow-centric design uses multiple blocks as the basis for constructing different production lines.
The supported applications, i.e., productions avoid redundancy in their designs.
Thus, the focus is placed on each block itself for independent expansion. 
To support all input combinations, our implementation consists of four building blocks that form SceneFactory: (1) tracking, (2) flexion, (3) depth estimation, and (4) scene reconstruction.
The tracking block is based on Mono SLAM and is extended to support RGB-D and RGB-LiDAR (RGB-L) inputs.
Flexion is used to convert the depth image (untrackable) into a trackable image. 
For general-purpose depth estimation, we propose an unposed \& uncalibrated multi-view depth estimation model (U$^2$-MVD) to estimate dense geometry.
U$^2$-MVD exploits dense bundle adjustment to solve for poses, intrinsics, and inverse depth.
A semantic-aware ScaleCov step is then introduced to complete the multi-view depth. 
Relying on U$^2$-MVD, SceneFactory both supports user-friendly 3D creation (with just images) and bridges the applications of Dense RGB-D and Dense Mono.
For high-quality surface and color reconstruction, we propose Dual-purpose Multi-resolutional Neural Points (DM-NPs) for the first surface accessible Surface Color Field design, where we introduce Improved Point Rasterization (IPR) for point cloud based surface query.

We implement and experiment with SceneFactory to demonstrate its broad applicability and high flexibility.
Its quality also competes or exceeds the tightly-coupled state of the art approaches in all tasks. We contribute the code to the community\footnote{\url{https://jarrome.github.io/SceneFactory/} }.
\end{abstract}


\section{Introduction}
\IEEEPARstart{S}{imultaneous} Localization and Mapping (SLAM) plays an important role in robotics.
Previous works have made substantial progress for robust localization of robots~\cite{engel2014lsd,mur2017orb,campos2021orb}, including in challenging environments~\cite{10286080}, while the mapping of the environment is often sparse.
With the aim of producing higher quality maps, dense mapping~\cite{kerl2013dense} and 3D reconstruction~\cite{newcombe2011kinectfusion} have been developed.
Dense mapping provides a point set as a mapping representation but is much denser. 
Alternatively, 3D reconstruction~\cite{newcombe2011kinectfusion} produces meshes for compactness and continuity reasons, which is more user friendly.

However, we observe that most works construct the whole pipeline in a highly tight-compact design.
This restricts the model to a specific application and is difficult to upgrade each submodule.
This leads to our first research question: \emph{Can we design a framework that will support all of the production lines?}

\begin{figure}[t]
	\centering
	\includegraphics[width=1\linewidth]{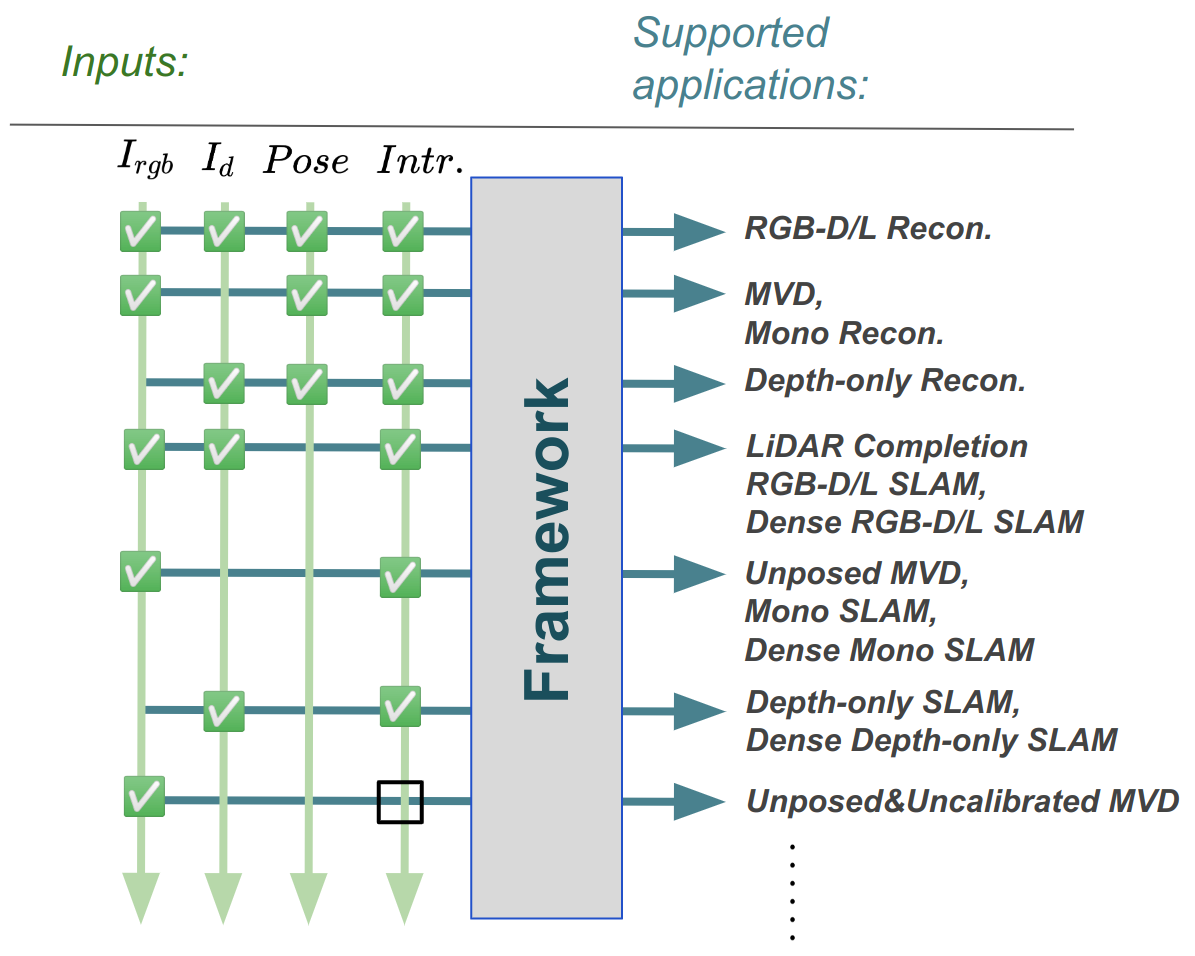}
	\caption{SceneFactory is workflow-centric and supports a wide range of applications given different input combinations of RGB $\V I_\text{rgb}$, depth $\V I_\text{d}$, pose $\V G$ and intrinsics $\V \theta$.}
	\label{fig:cover}
	\vspace{-.4cm}
\end{figure}

First of all, we traverse all possible input cases and look at what the tasks might be.
As shown in~\cref{fig:cover}, given different input combinations of RGB $\V I_\text{rgb}$, depth $\V I_\text{d}$, pose $\V G$ and intrinsics $\V \theta$, a number of applications could be run.
And our goal at the very beginning is to \emph{support them all} in a single framework.

To achieve this goal, we design our framework in a workflow-centric fashion.
The advantage of such a design philosophy is that all tasks are connected by a dependency graph to reduce redundancy.
For example, as in our design~\cref{fig:pipeline}, %fig:dependency},
one final task usually depends on the output from other tasks. 
However, the other tasks can also be the final tasks.
All subtasks work both independently and together, each block must be completed on its own. 
So we have four blocks in the workflow: tracking, flexion, depth estimation, and reconstruction.

On the tracking block, we subtract and find the greatest common divisor, Mono SLAM~\cite{monoslam,orbslam}.
Mono SLAM considers more complicated conditions than RGB-D and Stereo SLAM due to the difficult depth estimation.
Which operates the simplest sensor setup with fewer observation requirements. 
However, Mono SLAM relies on only sparse keypoints, and thus we also introduce a dense estimation task for dense depth-related applications.
The ``loose connection'' concept directs us to build a dense SLAM where each module focuses on its own task. 

When there is no RGB input, but depth input, we exploit the seminal work of ~\cite{Losch2023flexion}, called flexion. 
Flexion image is based on surface curvature, where in each pixel it calculates the difference between two normals from the neighboring pixels.
To make it compatible with our vision-based tracking block, we have to rely on trackable images. 
However, the depth image does not contain SE(3)-invariant features under different viewpoints.
So the flexion is applied then to convert the untrackable/unmatchable depth images into trackable images~\cite{Losch2023flexion}.

Then comes depth estimation, which has been widely studied with monocular and multi-view input.
While monocular depth~\cite{metric3d} naturally loses scale, the reliable methods usually operate multiple views.
Such methods are classified as (1) depth-from-video without known poses~\cite{demon,zhou2018deeptam,deepv2d}, (2) multi-view stereo with known poses~\cite{deepmvs,mvsnet,rmvsnet,cvpmvsnet,casmvsnet,vis-mvsnet} and (3) pointmaps without poses and intrinsics~\cite{dust3r}.
In our depth estimation module, we want to support all of them.
For high-reliability, we rely on a low-level vision model, optical flow~\cite{dkm} for correlation search, and use dense bundle adjustment to estimate the dense structure.
Here, all camera intrinsics, poses and depths are estimated in one optimization problem. 

For use in SLAM sequences, we further introduce \emph{good neighborhood selection} aiming to reduce the above situations.
However, even with this selection, the side effect of producing missing regions still exists.
An example is in~\cref{sec:depthEstimation}, where there is a hole in the wall. 
Inspired by MonoSDF~\cite{yu2022monosdf}, that the monocular depth shares point properties with the real depth, we turn to monodepth and find in our scene case that although monodepth is distorted between structures, in-structure coplanarity is preserved. 
Therefore, we propose to utilize this in-structure property with the ScaleCov model.
It relies on the learned deep covariance to fill the clipped locations with regressed scale image.
In SLAM-related applications, we additionally scale the depth to global to make the mapping consistent.

Although our depth estimation model fits well into mono SLAM, the model on its own supports unposed \& uncalibrated multi-view depth estimation.
This also provides a user-friendly application that requires only RGB images.

% NeRF
Afterwards, we perform the 3D reconstruction. 
Recently, there have been some successful attempts for immersive visualization under 3D reconstruction pipeline, such as ESLAM~\cite{johari2023eslam} and NeRF-SLAM~\cite{rosinol2022nerf}. 
They borrow the NeRF-like training scheme to produce high-quality view synthesis.
However, the ray integration naturally suffers from many samples with large space and time costs.
Therefore, we turn to another related branch, Neural Surface Light Fields (NSLF), without geometry optimization.
% NSLF
Recent research of NSLF has verified the efficiency on SLAM sequences~\cite{yuan2023online}. 
Because NSLF only learns the surface field, each surface position is not modified. % by irrelevant rays.
However, the existing online SLF method~\cite{yuan2023online} requires an external SDF model~\cite{huang2021di} to provide the surface points.
Therefore, we propose here dual-purpose multi-resolutional neural points (DM-NPs) to overcome this limitation.
Following NSLF-OL~\cite{yuan2023online}, which uses a multi-resolutional hash grid for efficient online learning, we create a \emph{multi-resolutional point grid for fast SLF learning}.
Moreover, since our point-based representation naturally provides position on surfaces, a straightforward idea is to use point rasterization.
However, point rasterization is a dysfunction compared to surface rasterization (as introduced in~\cref{sec:IPR}). 
To make it work, we propose an \emph{Improved Point Rasterization} which turns out to be really useful compared to the point rasterization of Pytorch3D~\cite{pytorch3d}.
Our CUDA implementation is 10 to 50 times faster than Pytorch3D's CUDA implementation.

In summary, as shown in~\cref{fig:cover}, we intend to design a modular scene modeling method, where each module supports individual applications, while they can be combined into one.

The contributions of this work are:
\begin{enumerate}
\item
We propose a workflow-centric framework, SceneFactory, that supports \emph{all incremental scene modeling} applications with different combinations under the connection of a dependency graph.	
\item
We introduce a dual-purpose multiresolution neural points representation for both Surface Light Fields (SLF) and Improved Point Rasterization (IPR). This is (1) \emph{the first surface-gettable SLF model}, (2) \emph{the first to make point rasterization as usable as surface rasterization}.
\item
We give a robust depth estimation block with (1) \emph{an unposed \& uncalibrated multiview depth estimation model (U$^2$-MVD)} and (2) \emph{a deep correlation kernel-based depth completion model, ScaleCov}.
\item
We capture the \emph{first dense mono SLAM purpose RGB-X dataset} for high-quality monocular reconstruction.    
\end{enumerate}

In the following, we first describe the related work on dense monocular SLAM, neural rendering in SLAM and multi-view depth estimation. 
Then, in three separate sections, we introduce the overall SceneFactory structure, the dual-purpose multi-resolution neural point representation and the unposed \& uncalibrated depth estimation.
Then, we conduct experiments to thoroughly evaluate the performance of the system.
Finally, we conclude this paper and attach the supplementary.

\section{Related Works}

\subsection{Dense SLAM}
% dense SLAM use sensor depth
Dense SLAMs are the main application of SceneFactory.
Originally for reliable reconstruction,
dense SLAM systems operate the metric dense depth from laser scanners or RGB-D sensors and focus more on the mapping representation, to approach a high-quality surface reconstruction. 
When metric depth is involved, traditional fusion methods~\cite{tsdf} incrementally update the Truncated Signed Distance Field (TSDF) and then extract afterwards a mesh using Marching Cubes~\cite{marchingcubes}.
More recently, neural priors have been used for Neural Implicit Representation~\cite{huang2021di,yuan2022algorithm}.
Uni-Fusion's implicit representation supports even more data properties without any training~\cite{yuan2024uni}.
With the even more recent trend of volumetric rendering, trained neural implicit rendering methods also come into view~\cite{zhu2022nice,yuan2023online}. 

% However, mono is more portable
However, the application scenarios of scanners and depth sensors are still limited by their low affordability, portability and accessibility.
Alternatively, almost everyone has their own monocular camera at hand, e.g., on a mobile phone.
Therefore, a lot of work has been done in recent years to explore dense SLAM with a monocular camera~\cite{bloesch2018codeslam,czarnowski2020deepfactors,koestler2022tandem,teed2021droid,rosinol2023probabilistic}. %there are still some works in this direction in the last years.
% CodeSLAM, DeepFactors
For Dense RGB SLAM, optimizing an entire sequence of depths is not feasible given the large number of variables involved.
To reduce the computational costs of depth estimation, 
CodeSLAM~\cite{bloesch2018codeslam} and DeepFactors~\cite{czarnowski2020deepfactors} optimize the latent codes of depth images.
However, the single-frame depth encoder-decoder needs to be pre-trained with similar data sets.
And mono-depth naturally loses scale and intrinsic information, usually resulting in distorted structures.
% Tandem
Instead of optimizing poses and depths at the same time, Tandem~\cite{koestler2022tandem} first solves frame poses and then uses the pre-trained MVSNet to recover the dense structure.
However, the MVSNets are usually overfitted with the trained camera intrinsic. 
This is suboptimal for custom cameras and new scenes.
% DROID-SLAM
On the other hand, DROID-SLAM~\cite{teed2021droid} ensembles dense optical flow modules into the pipeline, and optimizes poses and downsampled inverse depths using dense bundle adjustment.
% Sigma-Fusion
To cope with DROID-SLAM's noisy depth estimation, Sigma-Fusion~\cite{rosinol2023probabilistic} introduces depth uncertainty into DROID-SLAM's framework and uses TSDF to provide a high-quality dense reconstruction.

% NeRF-SLAM
Starting in 2022, more researchers realized the high potential of using NeRF for high-quality mapping.
Orbeez-SLAM~\cite{orbeez-slam} and NeRF-SLAM~\cite{nerf-slam} first embedded NeRF in SOTA-SLAM frameworks.
For example, Orbeez-SLAM generates a sparse map using OrbSLAM3~\cite{orbslam3} to set up a coarse occupancy grid for on-ray point sampling.
NeRF is then applied directly given the tracked poses.
Orbeez-SLAM gets high-quality rendering from NeRF.
However we find that it also inherits the problem of NeRF, especially for SLAM sequences.
It hardly works with non-around-object sequences.
While based on Sigma-Fusion, which provides dense depth in nature, NeRF-SLAM addresses the above problem by monitoring depth along with color.

From the design of Orbeez-SLAM~\cite{orbeez-slam} and NeRF-SLAM~\cite{nerf-slam}, that they have loosely connected pose-and-depth and depth-and-reconstruction relations respectively, 
we find that the loosely coupling does not just provide high flexibility, but also shows high advances when each sub-task is mature, such as the use of OrbSLAM3 and etc.
This point inspires us to make a workflow-centric design that has high expectations for each submodule.

\subsection{Neural Rendering in SLAM}

In our view, neural rendering is a topic that has the potential to replace reconstruction for high-quality mapping.

Many 3D reconstruction algorithms rely on marching cubes to extract mesh from explicit~\cite{tsdf} or implicit~\cite{huang2021di,yuan2022algorithm,yuan2024uni}.
This is not an efficient update due to the complicated (implicit)-to-explicit-to-mesh steps.
While the alternative view synthesis relying on rendering, shows a more direct way to extract visualization from implicit representations. 
View generation could be super efficient.

The success of neural rendering in SLAM begins with the invention of iMAP~\cite{sucar2021imap} and NICE-SLAM~\cite{zhu2022nice}. 
iMAP applies volumetric rendering to MLPs and optimizes poses and MLPs with photometric loss.
NICE-SLAM improves the rendering base and replaces the MLPs with hierarchical feature grids. 
This further improves the surface quality.
%To avoid the problem we mentioned in Orbeez-SLAM, NICE-SLAM uses RGB-D sequences to provide depth information to avoid the convergence problem.
NICE-SLAM certainly provides a good basis for neural rendering.
However, it still has the problems that 1. it is not real-time capable, 2. the color result is of low quality. 

Orbeez-SLAM~\cite{orbeez-slam} and NeRF-SLAM~\cite{nerf-slam} address these by coupling SOTA-SLAM with SOTA-NeRF, instant-ngp~\cite{instant-ngp}. 
NSLF-OL~\cite{yuan2023online}, on the other hand, focuses only on surface color and produces a real-time neural surface light field model to be used in conjunction with an
external real-time reconstruction model.

%FIXED: The next two sentences are not understandable
%A: the input are from other modules, it assume the inputs are very nice.
In our SceneFactory design, we expect high for each module.
Thus the reconstruction model does not worry about the inputs, e.g. depth.
So NSLF-OL is a good starting point.
However, working alongside other model makes SLF's performance highly dependent on the hosting reconstruction algorithm.
Therefore, in this work, we also propose to make the SLF model supporting surface extraction in the same representation.

\subsection{Multi-view Depth Estimation}
Depth from multi-view is obtainable in multiple ways, i.e., as depth-from-video, multi-view stereo, and pointmaps.

Depth-from-video directly estimates depth images and camera poses from video sequences with known camera intrinsics.
DeMoN~\cite{demon} has firstly utilized deep learning techniques to estimate the depth and motion with a single network.
DeepTAM~\cite{zhou2018deeptam} and DeepV2D~\cite{deepv2d} do not only work on image pairs, but process more images with alternating mapping and tracking modules.
However, according to~\cite{robust_mvd}, such methods are overfitted to the trained scale and camera parameters, which makes it difficult to generalize to arbitrary real-world applications.

Multi-view stereo estimates 3D geometry from unconstrained images with given intrinsic and extrinsic information.
%Here we only focus on the depth image estimation from MVS.
MVS is starting to get a boost with the trend of deep learning. 
DeepMVS~\cite{deepmvs}, as the first deep network-based method, aggregates sampled patchwise deep features to estimate full image depth. 
It has demonstrated the high potential of high-quality depth estimation with deep learning. 
Similarly, MVSNet~\cite{mvsnet} learns a feature map for whole images and fuses multi-view information into one feature map.
Then MVSNet's design serves as mainstream for following deep learning models~\cite{rmvsnet,cvpmvsnet,casmvsnet,vis-mvsnet}.
But, as mentioned before, MVS nets overfit the trained camera and scene.
This also leads to malfunction in new real scenes.
%[BA based]

Pointmaps methods start attracting more attention in 2024.
It uses a dense 2D field of 3D points as geometric representation.
Pointmaps are first used in visual localization~\cite{dsac,sacreg} and monocular 3D reconstruction~\cite{lin2018learning,wang2019mvpnet}.
A recent pointmaps-based work DUSt3R~\cite{dust3r}, designs a flexible stereo 3D reconstruction model.
DUSt3R unifies different 3D tasks in a groundbreaking way.
For the first time, it also provides a user-friendly interface that does not require camera parameters.

Our depth estimation module is also partly inspired by the high-flexibility of the DUSt3R.
We want to provide a user friendly interface like DUSt3R. 
But further, for supporting all above settings.

\begin{figure*}[t!]
	\centering
	\includegraphics[width=.8\linewidth]{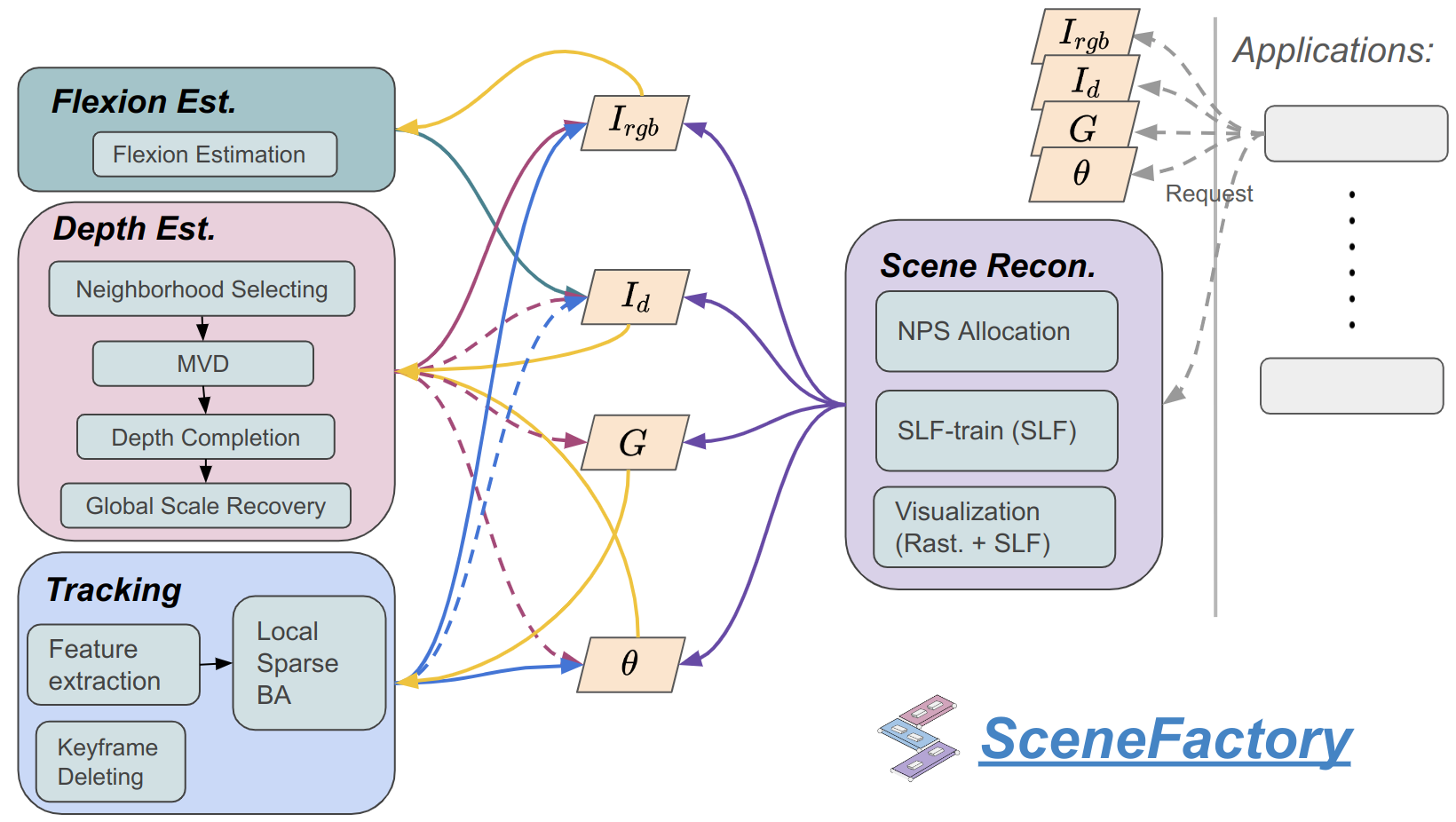}
	\caption{The dependency graph in SceneFactory.
          SceneFactory sends requests to its dependent sub-tasks (inputs/blocks).
          If the dependent sub-tasks are not complete, then each sub-task will call its corresponding dependent sub-tasks.
          The gray line indicates the requirement of a specific application.
          The yellow line shows the dependency of input (RGB $\V I_{rgb}$, depth $\V I_{d}$, pose $\V G$ and intrinsics $\V\theta$), which is triggered when the input value is None.
          The green, purple, blue, and pink lines show the dependencies of the Flexion, Depth, Tracking, and Scene Reconstruction blocks.
          The solid and dotted lines show mandatory and optional dependencies.
          Inside each block, the functions are applied one after each other, as shown by the black arrows.
	}
	\label{fig:pipeline}
\end{figure*}

\section{A Unified Framework for Incremental Scene Modeling}

SceneFactory supports various combinations of RGB $\V I_{rgb}$, depth $\V I_{d}$, pose $\V G$ and intrinsics $ \V \theta$ as input.
We plot the entire pipeline in~\cref{fig:pipeline} to have a brief overview.
All of the applications in~\cref{fig:cover} find their corresponding parts in this diagram.

SceneFactory consists of four building blocks: 
\begin{itemize}[]
	\item tracking block to complete pose,
	\item flexion estimation block to complete the image,
	\item depth estimation block to estimate and complete the depth,
	\item scene reconstructing block.
\end{itemize}

In this section, we introduce each of these blocks respectively~(\cref{sec:flexion_block,sec:depth_block,sec:tracking_block,sec:recon_block}) and combine them into a whole workflow~(\cref{sec:workflow}).

Note that, the last two (depth estimation and scene reconstruction blocks) are our contribution models, we add two more sections (\cref{sec:nps} and \cref{sec:depthEstimation}) after this framework section for a detailed explanation.

\subsection{Tracking Block}
\label{sec:tracking_block}
For the tracking block, we use Mono SLAM because it is a greatest common divisor with minimal demands.
More specifically, our tracking block is based on DPVO~\cite{dpvo}, while we generalize it to also support RGB-D/L input.

DPVO is based on sparse correspondences, while the pose $\V G$ is optimized with sparse bundle adjustment:
\begin{equation}
	\mathcal L_{track}(\V G, \V P) = \sum_{(k,j) \in \mathcal{E}} \parallel \Pi(\V G_{ij}\circ \Pi^{-1}(\V P^{'}_{k}))
	%\hat\omega_{ij}(\V G_j, \V P^{'}_{kj})
	- [\hat{\V P}^{'}_{kj}+\delta_{kj}] \parallel_{\Sigma_{kj}}^2.
\end{equation}
where $\mathcal{E}$ and $\Pi$ denote the edges and projection, $\V P^{'}_{k}$ is the patch $k$ in image $i$,  $\hat{\V P}^{'}_{kj}$ is the center of patch $\V P^{'}_{kj}$ in image $j$, $\delta_{kj}$ is the patch update.

The above formulation is efficiently solved with Gauss-Newton.
When the depth $\V I_d$ is given, we extract patch $\V P$ from the corresponding position.
We fix the patch from a valid depth pixel and optimize only the poses $\V G$ from $\mathcal L_{track}$.
We choose DPVO for the generalization because it does not require explicit extraction of matches, and thus well fits the sparse observation of the LiDAR.
From our experiments we learnt that this generalized version is more accurate by utilizing the metric depth.

\subsection{Flexion Estimation Block}
\label{sec:flexion_block}
\begin{figure}[!]
	\centering
	\subfloat[Colorized depth]
	{	\includegraphics[width=.48\linewidth]{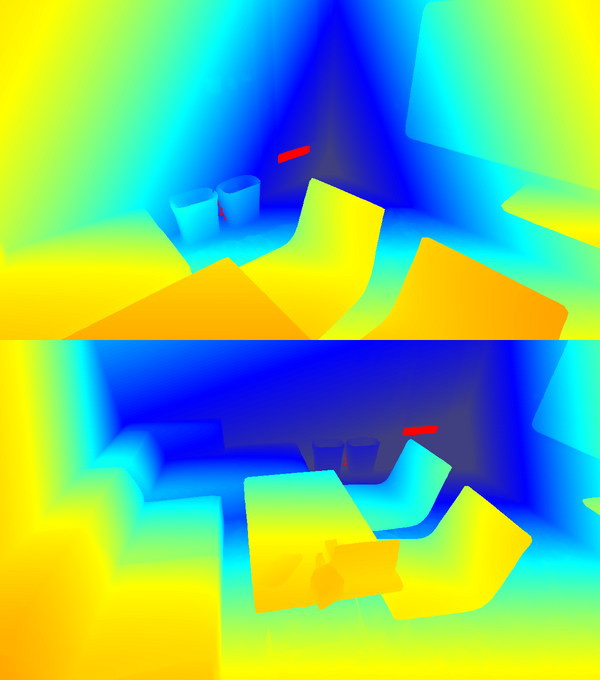}
	}
	\subfloat[Flexion]
	{	\includegraphics[width=.48\linewidth]{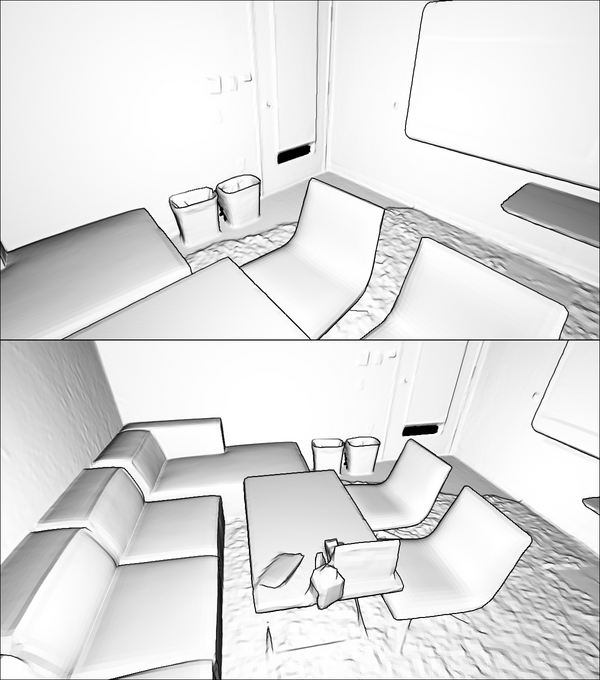}
	}
	\label{fig:flexion_convert}
	\caption{Depth images (left) and their corresponding trackable converted flexion images (right). }
\end{figure}
Flexion estimation is required when $\V I_{rgb}$ is missing, i.e., depth only is given.

To support depth-only tracking in our RGBD SLAM framework, a feature matching operation is required. However, the depth image does not contain SE3-invariant context for descriptor extraction. This is not trackable in our setting.
Therefore, we use the Flexion depth converter~\cite{Losch2023flexion} to convert the depth image into a flexion image that has the SE3-invariant properties.
An example is given in~\cref{fig:flexion_convert}. We colorize the depth only for better visualization. We see that flexion images contain a more consistent value, which is more suitable for feature matching~\cite{Losch2023flexion}.
Hence, depth-only applications are able to treat flexion as RGB and serve as RGBD applications as in~\cref{fig:exp_depthonly}.

\subsection{Depth Estimation Block}
\label{sec:depth_block}

The depth estimation block is mainly our U$^2$-MVD (in~\cref{sec:depthEstimation}) with intrinsics given for pose-free MVD to support incremental application.  
It computes the depth for RGB applications, allowing them to work as RGB-D applications.

Note that when no metric depth is specified, tracking frames only have sparse landmarks.
Depth estimation is then required when keyframe (KF) and good neighbor frames (NFs) are detected or MVD is called for certain application requests.
Once requested, this block will query memory for good neighbor selection (\cref{sec:neib_selection}), pose-free MVD (\cref{sec:depthRecovery}), depth completion (\cref{sec:depthcompletion}) and scale recovery (\cref{sec:scale_recovery}).

We put the depth estimation after the tracking sequences because the tracker optimizes poses in a local window, and the updated landmarks will affect the scale recovery.

\subsection{Reconstruction Block}
\label{sec:recon_block}

The reconstruction block uses our DM-NPs presented in~\cref{sec:nps}.
This block maintains two threads: 1) NPs allocation and online learning, and 2) visualization.
%\begin{algorithm}\subsubsection{NSLF-Train}

\subsubsection{Online-learning thread}
A training thread is used to continuously train our SLF model.
Similar to the online method, NSLF-OL~\cite{yuan2023online}, DM-NPs takes a new keyframe ($\V I_{rgb}^{'}$, $\V I_d^{'}$, $\V G^{'}$) and converts it to a colored point cloud $\{(\V p_n, \V c_n)_{n\in \{1,\cdots\}}\}$.
Based on the positions of point cloud, new neural points (NPs) will be allocated (as in~\cref{sec:alloc}). 
The input color value $\{\V c_n\}_n$ of the point cloud will be utilized to supervise the prediction $\{\V c^*_n\}_n$ (as in~\cref{sec:colorPred}).

\subsubsection{Visualization thread}
Another thread for the renderer relies on our Improved Point Rasterization~(\cref{sec:IPR}) for surface extraction, and color prediction~(\cref{sec:colorPred}) on surface. 
Our interactive GUI receives signals from the user to rotate and move the view camera.
Rendering is done in real-time for a first-person view of the scene~(\cref{sec:vis}). 

\subsection{Main Function}
\label{sec:workflow}

SceneFactory's main functions are $EstablishProductLine$ and $Step$ as in~\cref{algo:task_operation}.
\begin{algorithm}[htbp]
\SetAlgoLined
\DontPrintSemicolon
\SetKwFunction{FMain}{$EstablishProductLine$}
\SetKwProg{Fn}{Function}{:}{}
\Fn{\FMain{$(\V I_{rgb}$, $\V I_d$, $\V G$, $\V \theta$, $app )$}}{
\tcp{Build product line}
$pLine=AssemblingParts(\V I_{rgb}$, $\V I_d$, $\V G$, $\V \theta$, $app )$;\\
\Return $pLine$;
}
\textbf{End Function}

\SetKwFunction{FMain}{$pLine.Step$}
\SetKwProg{Fn}{Function}{:}{}
\Fn{\FMain{$(\V I_{rgb}$, $\V I_d$, $\V G$, $\V \theta$, $app )$}}{
\tcp{Product line start working part-by-part}
$\V V_{inter}\leftarrow pLine.Package(\V I_{rgb}$, $\V I_d$, $\V G$, $\V \theta$, $app )$;\\
\For{$f_{part}$ in pLine.parts}
{$\V V_{inter}=f_{part}(\V V_{inter})$}
\Return $pLine.Unpackage(\V V_{inter})$;
}
\textbf{End Function}
\caption{Main functions}
\label{algo:task_operation}
\end{algorithm}

When a task is triggered, 
SceneFactory successively checks the availability of image $\V I_{rgb}$, pose $\V G$, depth $\V I_d$ and intrinsics $\V \theta$, to prepare the production line.
Then, the operation is conducted following~\cref{fig:pipeline}.

For example, if $\V I_{rgb}$ is missing (depth only), the flexion block~(\cref{sec:flexion_block}) is added to the product line to estimate the trackable image from depth.
If SceneFactory is required to solve for the intrinsics without a $\V \theta$ input, it will add intrinsic-free U$^2$-MVD to the line for $\V \theta$.
If SceneFactory is asked to solve for poses without poses $\V G$ as input, it will add a tracker to the line for predicting the pose $\V G^{'}$.
If the metric depth is missing, the
depth estimator block~(\cref{sec:depth_block}) is added to estimate the depth.
While if the metric depth input is sparse, e.g., from a LiDAR, only the depth completion part will be added.
When reconstruction is requested by the application, the completed frame parameters with image, depth, pose and intrinsics are required to be fed into the reconstruction block~(\cref{sec:recon_block}) for online learning and visualization.
And the corresponding parts are all added to the product line.

Then the production line treats each step function as a task. If a single task is requested (such as MVD, Completion and etc.), the production line will return the step result as a product.
While if sequential application is requested (such as SLAM, Reconstruction and etc.), the production line will conditionally step on each frame for the intermediate and return the product at the end of this production process.

\begin{figure*}[t]
    \centering
    \psfragfig*[width=1.8\linewidth]{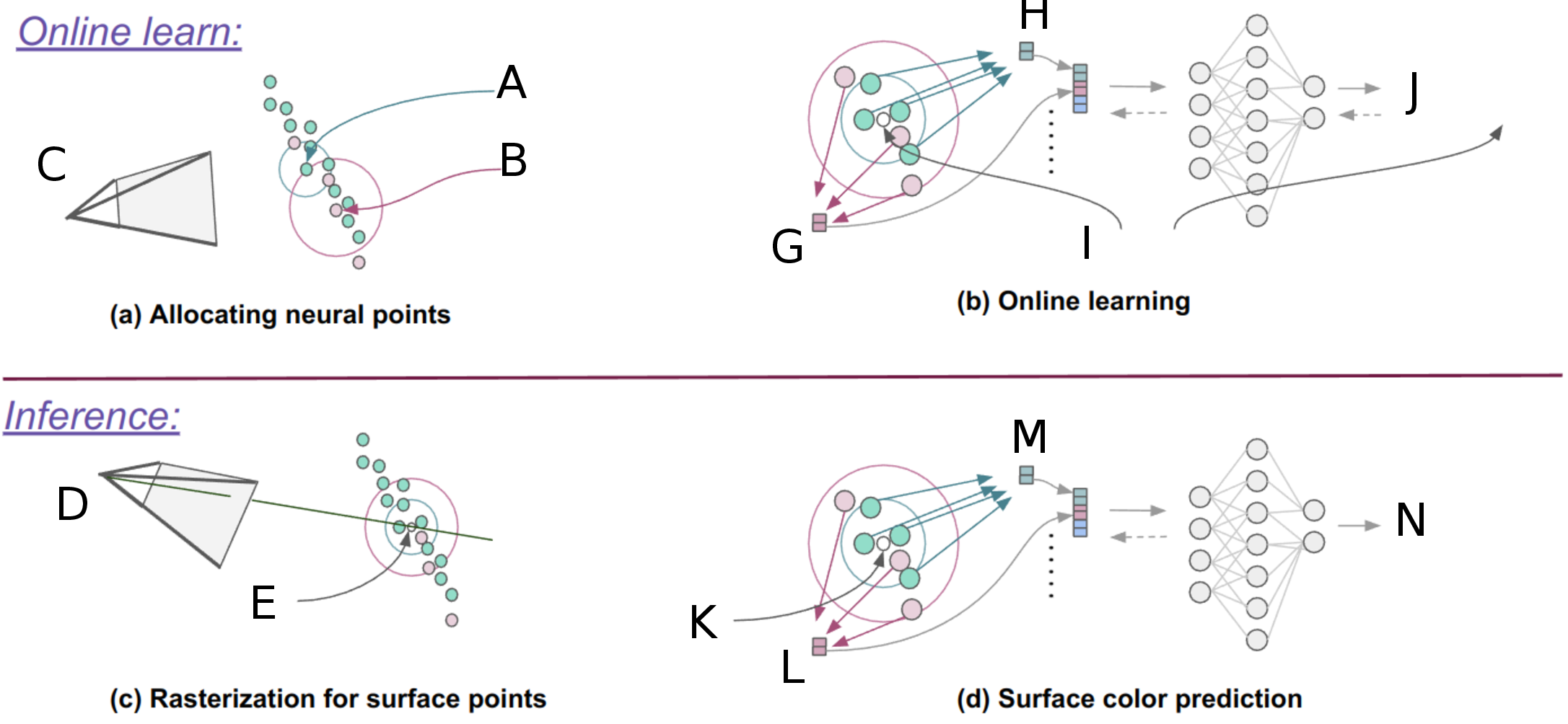}{
        \psfrag{A}{$(\V G\cdot \V p_l, \V F_{\V p_l})$}
        \psfrag{B}{$(\V G\cdot \V p_{l+1}, \V F_{\V p_{l+1}})$}
        \psfrag{C}{$\V G$}
        \psfrag{D}{$\V G$}
        \psfrag{E}{$\V p^{*}$}
        %\psfrag{F}{$ $}
        \psfrag{G}{$\V F_{l+1}$}
        \psfrag{H}{$\V F_l$}
        \psfrag{I}{$(\V p_{gt}, \V c_{gt})$}
        \psfrag{J}{$||\V c - \V c_{gt}||^2_2$}
        \psfrag{K}{$\V p^{*}$}
        \psfrag{L}{$\V F_{l+1}$}
        \psfrag{M}{$\V F_l$}
        \psfrag{N}{$\V c^{*}$}
    }
    \caption{Illustration of the multiresolution neural points in 2D.
        The top row indicates the (a) allocating and (b) training during online learning. The bottom row shows the (c) rasterization and (d) color prediction for visualization.
        We use multiple levels of neural points, for example, {\color{JungleGreen}green dots} for low level and {\color{Mypink}pink dots} for higher level. The corresponding circle indicate the resolution of that levels of points.}
    \label{fig:encode}
\end{figure*}

\section{Dual-purposes Multiresolutional Neural Points}
\label{sec:nps}

In this section, we introduce the dual-purpose representation that simultaneously supports Surface Light Fields (SLF) and point rasterization.
This representation is the first SLF method to support \emph{surface querying}.

\subsection{Multiresolutional Neural Points}

Intuitively, a neural point is a point with a feature.
We denote a neural point as $\V v = (\V F_v, \V p_v)$ with feature $\V F_v\in\mathbb{R}^{m}$ and position $\V p_v\in\mathbb{R}^3$.

Then, inspired by instant-ngp~\cite{instant-ngp}, we use multiresolution neural points to improve learning efficiency.
These are multiple sets of neural points with different densities.
We denote them as $\V V_a$ the set of $\V v$ with resolution/density $a$. 
\cref{fig:encode} shows a illustration.

During training, we follow NSLF-OL~\cite{yuan2023online} to feed the set of colored pairs ${(\V p_i, \V c_i)}_i$ for coding and training.
For each point position, we examine the density in its region and assign neural points. 
Then, the MLP coding is applied.

\subsubsection{Neural Points Allocation}
\label{sec:alloc}

For the feedpoint set $\V Q=\{\V p_i\}_i$, given a set of resolutions $\V a$, we downsample to get $\{\V Q_a\}_{a\in\V a}$.
Then, as depicted in \cref{fig:encode} (a), for each point $\V p_i$ in $\V Q_a$, we check the closest distance to $a$ level neural points, if distance is greater than threshold $t_a$, the set of neural points is extended by adding this point and assigning it initialized with the feature.

\subsubsection{Surface Points Encoding}
\label{sec:encoding}

SLF's inference is on the surface, so this coding is an interpolation of the feature.
For inference point $\V p\in\V Q$, we utilize an efficient $K$-d tree to find its $K$ nearest neighbor ($\{\V v^a_k\}_{k\in\{1,\cdots,K\}}$ with distance $\{d^a_k\}_{k\in\{1,\cdots,K\}}$) of inference point from each level as in as in \cref{fig:encode} (b).
The $a$ level feature is
\begin{equation}
F^a_{\V p}=\sum_{i\in\{1,\cdots,K\}} w_i F^a_{v_i},
\end{equation}
where $w_i= exp(-\frac{{d_i}^2}{\sigma})/ \sum_j exp(-\frac{{d_j}^2}{\sigma})$.

\subsubsection{Color Prediction}
\label{sec:colorPred}

The color prediction is generated by concatenating features along different levels and decoded with MLP: $f_\text{MLP}\circ f_\text{concat}((F^a_{\V p})_{a\in\V a})$.

\subsection{Mapping via Online Learning}
\label{sec:indirect_mapping}

We learn this SLF in an online fashion by continuously feeding data according to~\cite{yuan2023online}.
A main thread of this module is to continuously train the SLF.
As soon as a new posed image with depth is input to the renderer, it is assigned a trained iteration, $it_\text{trained}$, to train in a least-touch strategy.

During the training iteration of a given frame, it randomly samples $n_\text{train}$ pixels and passes the corresponding point pairs $\{(\V p_n, \V c_n)_{n\in \{1,\cdots,n_\text{train}\}}\}$ to the neural allocation~\cref{sec:alloc} and prediction~\cref{sec:colorPred} for the resulting color $\{\V c^*_n\}_n$.

We compute the MSE
\begin{equation}
L_\text{MSE} = \sum_{n\in \{1,\cdots,n_\text{train}\}}\parallel \V c^*_n - \V c_n  \parallel^2
\end{equation}
and use the Adam optimizer for stochastic gradient descent optimization of the features of neural points and MLP.
In addition, we use the following two strategies to speed up the training.

\subsubsection{Jump-start Training Strategy}

From the experiments, we find that our SLF converges slowly on the first frame, while super fast for the following frames if the first frame gets coarse color.
Otherwise, the other frames would also be relatively slow.

We think the problem arises from the chaos of parameters as it is randomly initialized.
So we set a $10\times$ higher learning rate for the first $5$ iterations and recover then. 
This way, even the first image converges in a second.

\subsubsection{Least-trained First Training Strategy}

The second training strategy is that after the first frame, when another frame is input, the previous well-trained frame shouldn't have the same chance to be trained as the new one.
So an intuitive way is to always train the least trained frame.
In this way, all subsequent frames are converged in one second.

\subsection{Improved Point Rasterization (IPR)}
\label{sec:IPR}

\begin{figure}
\centering
\includegraphics[width=.8\linewidth]{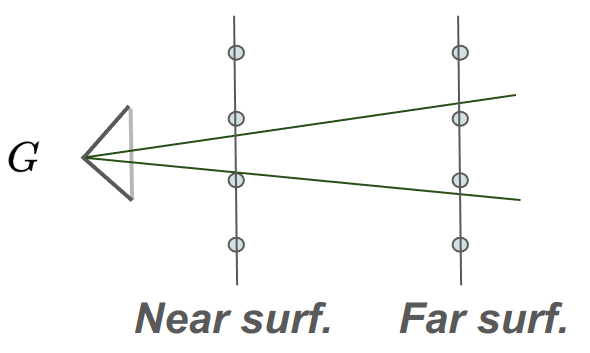}
\caption{Near-far-imbalance. The ray penetrates the near surface.}
\label{fig:near-far-imbalance}
\end{figure}

Above we explained the training process where the depth is given.
During the inference, we have to estimate the depth.
To this end, point rasterization is introduced as in \cref{fig:encode} (c).
For simplicity, we will now operate on points in camera space.
To render an image with a resolution of $H\times W$, we cast a batch of rays $(\V o_i=\V 0, \V d_i\in{\mathbb S^2})_{i\in\{1,\cdots,HW\}}$ with $(\V o, \V d)$ as the ray source and direction.
In implementation, the positions of the neural points are first transformed into NDC space.

Setting camera space point $\V p = [X,Y,Z]^T$ as an example, the projected point in NDC is then $\V p_\text{ndc}=[f_x \frac{X}{Z} + p_x,f_y \frac{ Y} { Z} + p_y, \frac{1} { Z}]^T$, where $(f_x, f_y, p_x,p_y)$ are intrinsic parameters.
In this way, each ray is directed to $+z_\text{ndc}$, i.e., $\V d_\text{ndc} = [0,0,1]^T$. The source of the NDC space ray is obtained by projecting $\V d$ accordingly.

Points within a radius $r_\text{ndc}$ around each ray $(\V o_\text{ndc} = [x_\text{ndc}, y_\text{ndc},z_\text{ndc}]^T, \V d_\text{ndc})$ are then computed using only the $(x,y)$ coordinate.

Thus, for each ray $(\V o, \V d)_i$, we find its $K_\text{ray}$ nearest neural point neighbors and the distance $\{(\V p_{nb,k},d_{nb,k})\}_{k\in\{1,\cdots,K_\text{ray}\}}$ within the radius.
The rasterized point is thus
\begin{equation}
\V p_\text{raster}=\sum_{k\in\{1,\cdots,K\}} w_{nb, k}\V p_{nb,k},
\end{equation}
where $w_{nb,k}= exp(-\frac{{d_k}^2}{\sigma})      / \sum_j exp(-\frac{{d_j}^2}{\sigma})$.
However, unlike mesh rasterization, point rasterization suffers from the following problems:
\begin{enumerate}
\item the point distribution in NDC space is distorted. 
That is, the points near the camera are more sparse while the points far from the camera are more dense compared to Euclidean space.
This results in holes in the near regions.
\item the rasterization may contain several layers of points.  
\end{enumerate}
%A straight forward solution is to rasterize again at hole pixels with a large radius.
To deal with the hole problem, we introduce an adaptive radius to PR. This changes the radius depending on the depth.

Considering that our finest resolution is $r=0.005m$ and the screen space is at $z=1$, we expect that each point with a $0.005m$ gap can completely cover a pixel at $z=1$.
The coverage radius should be $l_\text{coverage}=\sqrt{2}(r\cdot f_{im})$ where $f_im$ is the focal length parameter. 
To simplify the implementation, our CUDA code creates a $l_\text{coverage}\times l_\text{coverage}$ window for this point.

If $z<1$, the scenario in~\cref{fig:near-far-imbalance} occurs because the points are distorted in the screen space.
The ray will then penetrate the near surface.
To solve this problem, we increase the coverage area, the coverage length becomes $l_\text{coverage} = \sqrt{2}(r\cdot f_{im})/z$.

\begin{figure}
\centering
\includegraphics[width=.9\linewidth]{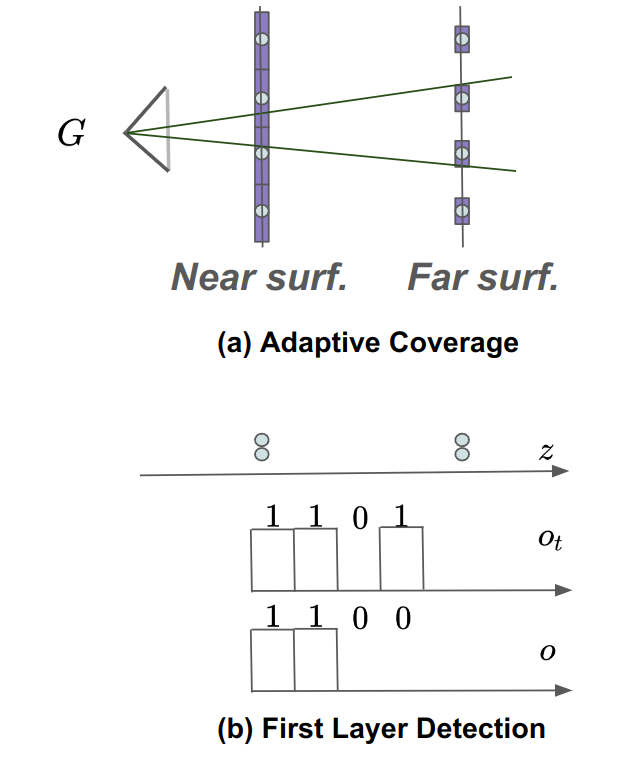}
\caption{Improved Point Rasterization. Increasing the coverage area (above) avoids the Near-far-imbalace, cf. Fig~\cref{fig:near-far-imbalance}} and the first-layer detection keeps only these points (below). 
\end{figure}

For the second problem, i.e., the problem of multiple layers, we propose to use first-layer detection to keep only the points of the first layer.
Given $n$ sorted points along the ascending $z$-axis, which are traced by a given ray, $\{ z_0,\cdots z_n\}$, we compute the occupancy of point $z_t$.

\begin{equation}
o_t=\left\{
\begin{aligned}
	1\ \ \ \ \ \ \ \ & && t=0,\\
	z_{t+1}-z_t < th& &&t>0. \\
\end{aligned}
\right.
\end{equation}
Then, by cumulating production, we get the mask of the first layer points:
\begin{equation}
o_{\{0,\cdots\}} = cumprod(o_{\{0,\cdots\}}).
\end{equation}

We implement our IPR with CUDA, which is about 20 times faster than Pytorch3d's CUDA implementation (\unit[40]{FPS} vs \unit[2]{FPS}). 
We demonstrate the effect of IPR in~Suppl.\ref{sec:suppl:ipr}.

\subsection{Visualization}
\label{sec:vis}

This function plays an important role in rasterizing the depth image and in extracting the surface point colors under certain viewpoints.
First, we rasterize at the level of neural points with the highest resolution.
Then, transforming back to world coordinates with $\V T$, the color prediction is obtained as in \cref{fig:encode} (d), by
\begin{equation}
\V c_{raster} = f_{MLP}\circ f_{concat}((F^a_{\V T \V p_{raster}})_{a\in\V a}).
\end{equation}
We follow~\cite{yuan2023online} to create another thread with an interactive GUI to provide a first-person view of the scene. 

This thread receives a signal from the user to rotate and move the view camera.
The view synthesis is rendered in real-time.

\section{Depth Estimation}
\label{sec:depthEstimation}

\begin{figure*}[t!]
	\centering
	\includegraphics[width=1\linewidth]{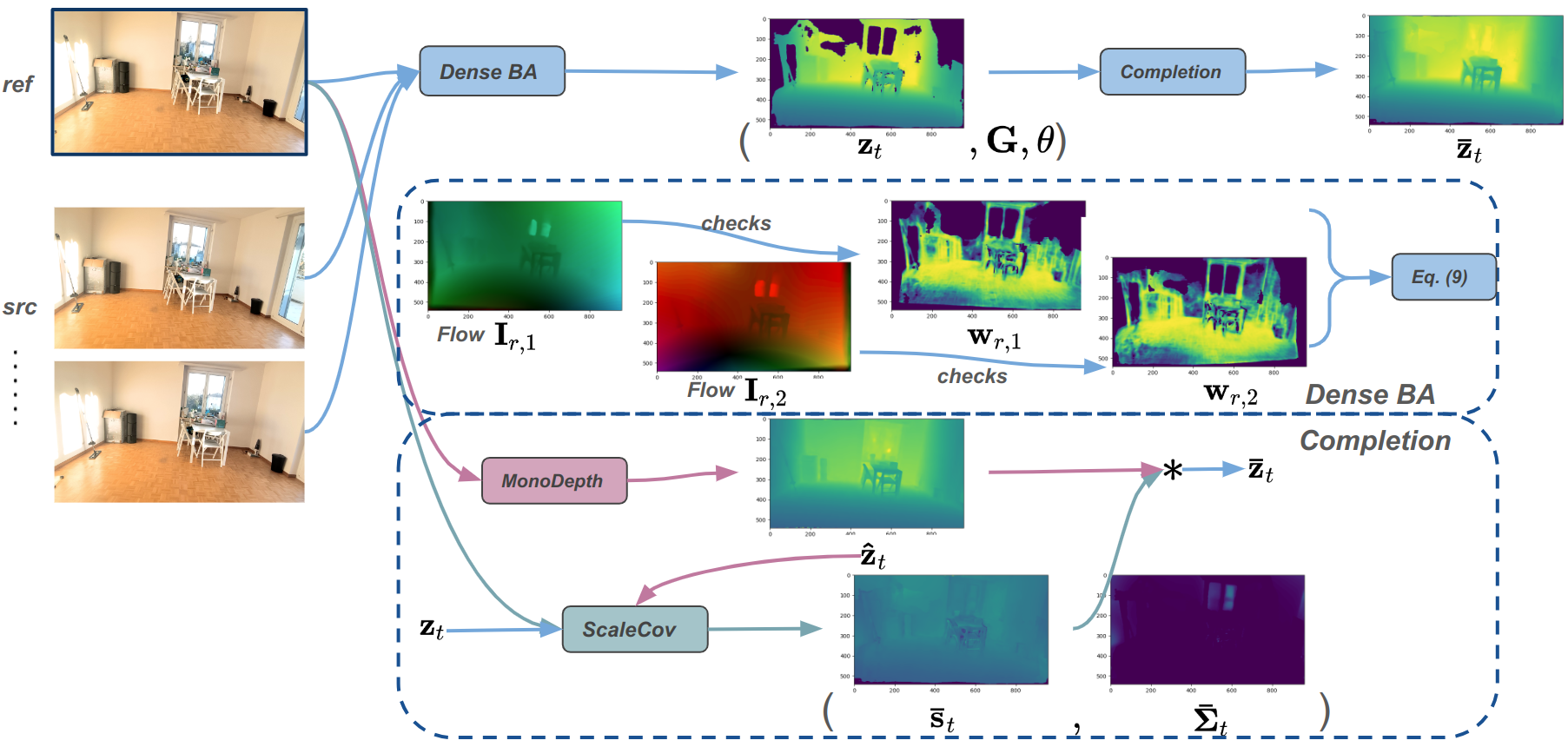}
	\caption{Depth estimation via unposed \& uncalibrated multi-view depth estimation model (U$^2$-MVD). The overall pipeline is shown.}
	\label{fig:mvd}
\end{figure*}

In this section, we introduce our unposed \& uncalibrated multi-view depth estimation model (U$^2$-MVD) that is depicted as in~\cref{fig:mvd}. 
As an adjunct to SLAM, we additionally add good neighbor frame selection for more suitable frames.

\subsection{Depth Recovery with Dense Correlation}
\label{sec:depthRecovery}

We acquire dense correspondences from SOTA optical flow estimation model DKMv3~\cite{dkm} for pixel-wise correspondences $\V l_{i,j}\in\mathbb{R}^{H\times W\times2}$ between the frames $i$ and $j$.
%, high quality dense correspondences are acquired.
Then the expected correspondence pixel would be $\V x_{j}^*=\V l_{i,j}+\V x_{i}$, where $\V x_{i}$ is the pixel coordinate in frame $i$.
However, if false correspondences occur, it will strongly affect the resulting pose and depth.
That is, the optical flow cannot be fully trusted.

\subsubsection{Cross Check}

Because single-source matching is risky, we apply cross-checking of correspondences to ensure a high-quality match:
\begin{align}
	\V w_{i,j,(u,v)}= 
	\begin{cases}
		1               & \text{if } ||\V x_{i,(u,v)} - (\V x^*_{j,(u,v)}+ \V I_{j,i,(u,v)})||_2< 0.5, \\
		0,              & \text{otherwise,}
	\end{cases}
\end{align}
where $(u,v)$ is the pixel coordinate in the image.

\subsubsection{Static Check}

We have assumed that our scene is static.
However, optical flow is not designed for static scenes. 
That is, the flow is not constrained by the rigid body. 
Therefore, unlike previous works~\cite{teed2021droid} that trust the flow, we apply epipolar constraint to filter out the non-rigid flow:
\begin{align}
	\V w_{i,j,(u,v)}= 
	\begin{cases}
		1               & \text{if } dist_{line}(\V L_{j,(u,v)},\V x^*_{j,(u,v)})^2< 3.84, \\
		0,              & \text{otherwise}
	\end{cases}
\end{align}
where $\V L_{j,(u,v)}$ is the epipolar line for $\V x^*_{j,(u,v)}$, $dist_{line}$ is the distance from point to line.
The epipolar line $\V L_{j}$ is obtained by solving the essential matrix with $1000$ randomly chosen correspondences.

\subsubsection{In-image Epipole Check}

However, since the above static check relies on the epipolar constraint, this check does not work near the epipole, i.e., if the epipole is in the image, the near-pixels will all pass the static check.
Therefore, we compute the position of the epipole in the image and filter out the near-pixel correspondences.

\subsubsection{Dense Bundle Adjustment (DBA)}

Although the SOTA optical flow is used, from experiments we learnt, the single flow is still too fragile for dense reconstruction use.
Therefore, more source frames are usually involved in MVD tasks.
In our formulation, we project the depth of the reference frame to all source frames and monitor over dense correspondences between each reference-source pair.
In terms of SLAM, we also collect dense matching from the reference (frame $t$) to neighbor frames ($\mathcal{B}_t$) in a window and compute dense BA to stabilize the inverse depth of reference frame $\V d_t$.

Then we follow DROID~\cite{teed2021droid,droidcalib} to denote the poses, the intrinsics and the projection function as $\V G$, $\V \theta$ and $\Pi$.
The cost function of the Dense Bundle Adjustment (DBA) is the sum of the projection errors over all neighbouring frames:
\begin{equation}
	\label{eq:DBA}
	\mathcal L(\V G, \V d_t, \V \theta) = \sum_{j \in \mathcal{B}_t} \parallel \V x^*_j - \Pi(\V G_{tj}\circ \Pi^{-1}(\V x_{t}, \V d_t, \V \theta),\V \theta) \parallel^2.
\end{equation}
This cost function is solved efficiently with Gaussian-Newton. For the detailed formulation, please refer to DROID-SLAM~\cite{teed2021droid,droidcalib}.
\cref{eq:DBA} directly solves poses, intrinsics and inverse depth at the same time \emph{with only optical flow (dense correspondences) as true value}.

\subsection{Monocular Depth and Depth Completion}
\label{sec:depthcompletion}

In addition to the DBA, we utilize Metric3Dv2~\cite{metric3dv2} for monocular depth $\V {\hat{z}_t}$ on the reference frame $t$. 
$\V {\hat{z}_t}$ guides the DBA from a good start point.
While more importantly, we use $\V {\hat{z}_t}$ to assist in completing the DBA depth with vacancy.

\begin{figure}[t!]
	\centering
	\subfloat[width=.5\text][Frontview]{
		\centering
		\includegraphics[width=.5\linewidth]{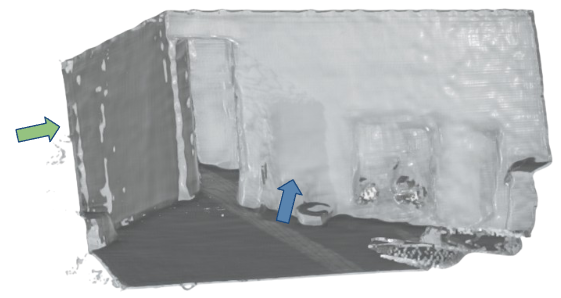}
	}
	\subfloat[width=.5\linewidth][Topview]{
		\centering
		\includegraphics[width=.5\linewidth]{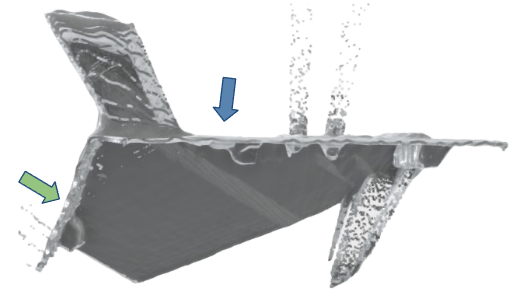}
	}		
	\caption{Example for the monocular depth estimation. {\color{ForestGreen}Green} arrow and {\color{blue}blue} arrow point to two walls.}
	\label{fig:mono_depth}
\end{figure}

Recent monocular depth methods~\cite{metric3d,metric3dv2} provide high-quality depth estimation of structures.
Please find~\cref{fig:mono_depth}. In the frontview, we can find a well-predicted shape of the room.
However, because the estimation from a single image naturally loses scale and intrinsic information, the relative positions between different structures are distorted.
See the top view of~\cref{fig:mono_depth}, the arrows pointing to the walls are not placed correctly.
Besides, monodepth cannot handle vision illusions such as Ames room in~\cref{fig:ames_imgs},
which makes the inter-structure of monodepth theoretically not reliable. 

Nevertheless, we still find useful information: the coplanarity is preserved.
Which reminds us to extract useful information from \textbf{intra-structure}. 

Therefore, we intend to utilize a learned covariance function that semantically identifies intra- and extra-structures, e.g., DepthCov~\cite{depthcov} that learns deep prior. 
However, a problem with DepthCov is that it directly regresses depth image with true sparse depth observation.
This means that if there is no true sample on certain structures, the depth of such structures will be arbitrarily wrong.
An extreme example is with only $1$ sample, then the whole depth image will be the same value.

Conversely, we introduce ScaleCov, which regresses scale image for monodepth $\V {\hat{z}_t}$ given DBA depth $\V {z_t}$ as observation.
ScaleCov does not face this problem, because in the worst case, identical scale image does not ruin the depth structures.

Following DepthCov, we utilize its deep-learned covariance function (kernel function) $k$ to formulate ScaleCov as a Gaussian Process Regression:
\begin{equation}
	\V s_{*} = \V K_{fn}(\V K_{nn} + \sigma_n^2\V I)^{-1}\V {s}_n,
\end{equation}
\begin{equation}
	\V \Sigma_{*} = \V K_{ff} - \V K_{fn}(\V K_{nn} + \sigma_n^2\V I)^{-1}\V K_{nf}.
\end{equation}
Where, we sample non-vacancy depth from DBA depth and compute scales $\V s_n=\V z_{t,n}/\V {\hat{z}}_{t,n}$, and use $\V K_{ff}$, $\V K_{fn}$ and $\V K_{nn}$ to denotes the correlation matrix between query points, query and observations, and observations.
$\V s_{*}$ and $\V \Sigma_{*}$ are the queried scale and variance that would be associated with the depth image.

For the finished image we consider the depth to be complete is more trustworthy and let
\begin{align}
	\V {\bar{s}}_t(\V x)= 
	\begin{cases}
		\V z_{t}(\V x)/\V {\hat z}_{t}(\V x),& \text{if } \V z_t(\V x)> 0\\
		\V s_{*}(\V x),              & \text{otherwise}
	\end{cases}
\end{align}
\begin{align}
	\V {\bar{\Sigma}}_t(\V x)= 
	\begin{cases}
		\V 0,& \text{if } \V z_t(\V x)> 0\\
		\V \Sigma_{*}(\V x),              & \text{otherwise}
	\end{cases}
\end{align}
where $\V x$ is the pixel coordinate.

The final completed depth is
\begin{equation}
	\V {\bar{z}}_t = \V {\bar{s}}_t * \V {\hat{z}}_t,
\end{equation} 
where the corresponding pixel-wise variance is $\V {\bar{\Sigma}}_t$.

In addition to the MVD completion shown in~\cref{fig:mvd}, ScaleCov also supports sparse LiDAR depth completion as shown in~\cref{fig:completion}.

\subsection{Depth Frame Selection}
\label{sec:neib_selection}

We observe that all previous dense mono SLAMs estimate global geometry using keyframes.
It makes sense to estimate depth for keyframes because during the tracking process, the main burden of keyframes is to gain information. 
The goal is to keep the landmarks in view.

However, we find it problematic to use neighbouring keyframes to support depth recovery.
This does not guarantee a good basis for triangulation, especially in dense cases.

Therefore,
the good neighbor frames (NF) may not be keyframes (KF).
In addition, the tracking model stores non-KF poses as a relative pose to the previous KF.
This may not be accurate. 
When using our depth model, we only input intrinsics and images to estimate the depth without using the poses of the NFs.

Next, we ask the question: ``\emph{How to choose a frame that is good for depth estimation?}''
Reliable depth estimation requires a reference frame for triangulation.
Therefore, our answer is: ``\emph{A frame has a good neighborhood for triangulation}''.

Since the tracking model provides poses for all frames, we use the relative poses to the reference frame to help verifing.
We select neighboring frames according to~\cref{algo:depth_frame_selection}, where the relative facing angle and baseline are used. 

\subsection{Recover Scale to Global}
\label{sec:scale_recovery}

Since~\cref{sec:depthRecovery} is applied without poses, a scale recovery operation is required to adjust to the tracking scale. 

While the global scale is hidden in the tracked landmarks or provided poses. 

When landmarks are given, the relative scale from local to global is recovered by using RANSAC regression\footnote{\url{https://scikit-learn.org/stable/modules/generated/sklearn.linear_model.RANSACRegressor.html}}
$s_{t,g}=RANSACRegressor(\{\V z_t^l(\V x)/\V {\bar{z}}_t(\V x)*s=1\}_{\V x\in \mathcal{P}_t})$, where $\V z_t^l$ and $\V {\bar{z}}_t$ are the landmarks from the tracker and the estimated depth from~\cref{sec:depthRecovery}, $\mathcal{P}_t$ and $\V x$ are the set of landmarks in frame $t$ and pixel coordinate.
When landmarks are missing, we adjust depth with the scale difference between solved and provided poses.

\section{Experiments}
The main experiments are conducted on multi-view depth estimation, surface light field and dense SLAM.

\subsection{Settings}
\subsubsection{Implementation Details}

We use $l=3$ levels of neural points with resolution starting at $r_0=0.005m$ with a multiplier $m_{reso} = 4$.
Our reconstruction model uses the Adam optimizer with $lr=1e-3$ for both NPs and MLP.
The k-nearest neighbor search between the query and NPs is done with a $K$-d tree, while the rasterization is done with our CUDA-implemented IPR.
For the depth estimation model, we rely on DKMv3~\cite{dkm} for dense correspondences (optical flow), and Metric3Dv2~\cite{metric3dv2} for monocular depth.
We follow DepthCov~\cite{depthcov} to build our ScaleCov for depth completion.
We extend the monocular tracking model DPVO~\cite{dpvo} to RGB-D and RGB-L.
Due to the optimization of local windowed frames, we only allow out-of-window (fixed) frames for depth and reconstruction.
All experiments are conducted on a PC with an Intel i9-13900KSi9-13900KS CPU and an NVIDIA GeForce 4090 GPU.

\subsubsection{Datasets}
\label{sec:datasets}
Our experiments mainly rely on datasets for scenes:

\paragraph{Replica~\cite{sucar2021imap}}
Replica has recently become the most widely used synthetic dataset for reconstruction and synthesis of views.
It consists of 8 RGB-D sequences for rooms and offices.

\paragraph{ScanNet~\cite{dai2017scannet}}
ScanNet is also widely used for RGB-D reconstruction. 
However, ScanNet is captured with old camera sensors with high motion blur.
Which is harmful for dense matching.

%According to our own data, even the current phone camera [] supports 60FPS.
%Therefore, we consider ScanNet to be outdated for dense mono at the current state of sensor technology

%\paragraph{ScanNet++~\cite{yeshwanth2023scannet++}}
%Unlike the low quality and blurry images of the ScanNet dataset, ScanNet++ aims to provide high quality datasets for the purpose of novel view synthesis.
%However, ScanNet++'s high quality DSLR shots are not sequences.
%Its lower quality iPhone captures are sequences.
%
%Therefore, we use posed frames for their DSLR captures. 
%For the iPhone captures we evaluate Dense-SLAM.
\paragraph{KITTI~\cite{Geiger2013IJRR}}
The KITTI dataset is one of the most famous outdoor datasets with stereo camera, Velodyne laserscanner and GPS.
We use its depth completion data branch for RobustMVD benchmarking.

\paragraph{ETH3D~\cite{schops2017multi}}
ETH3D is a widely used benchmark in the field of 3D reconstruction. 
This dataset provides multi-view images with ground truth poses and mesh reconstruction.
We use its multi-view benchmark data branch for RobustMVD.

\paragraph{DTU~\cite{jensen2014large}}
DTU is the most widely used object-level dataset in deep learning-based multi-view stereo work. 
DTU provides high-quality images with dense point cloud for each object.
Although the tabletop object is not the research focus of this paper, we use DTU for RobustMVD.

\paragraph{Tanks\&Temples~\cite{Knapitsch2017}}
Tanks\&Temples captures real indoor and outdoor scenes with high-quality videos.
It is mostly used alongside the DTU dataset for generalization testing.
We also use it for RobustMVD.

\begin{figure}[t!]
	\centering
	\begin{minipage}{\linewidth}
		\subfloat[width=0.45\linewidth][Handheld RGB-L]{
			\centering
			\includegraphics[width=0.45\linewidth]{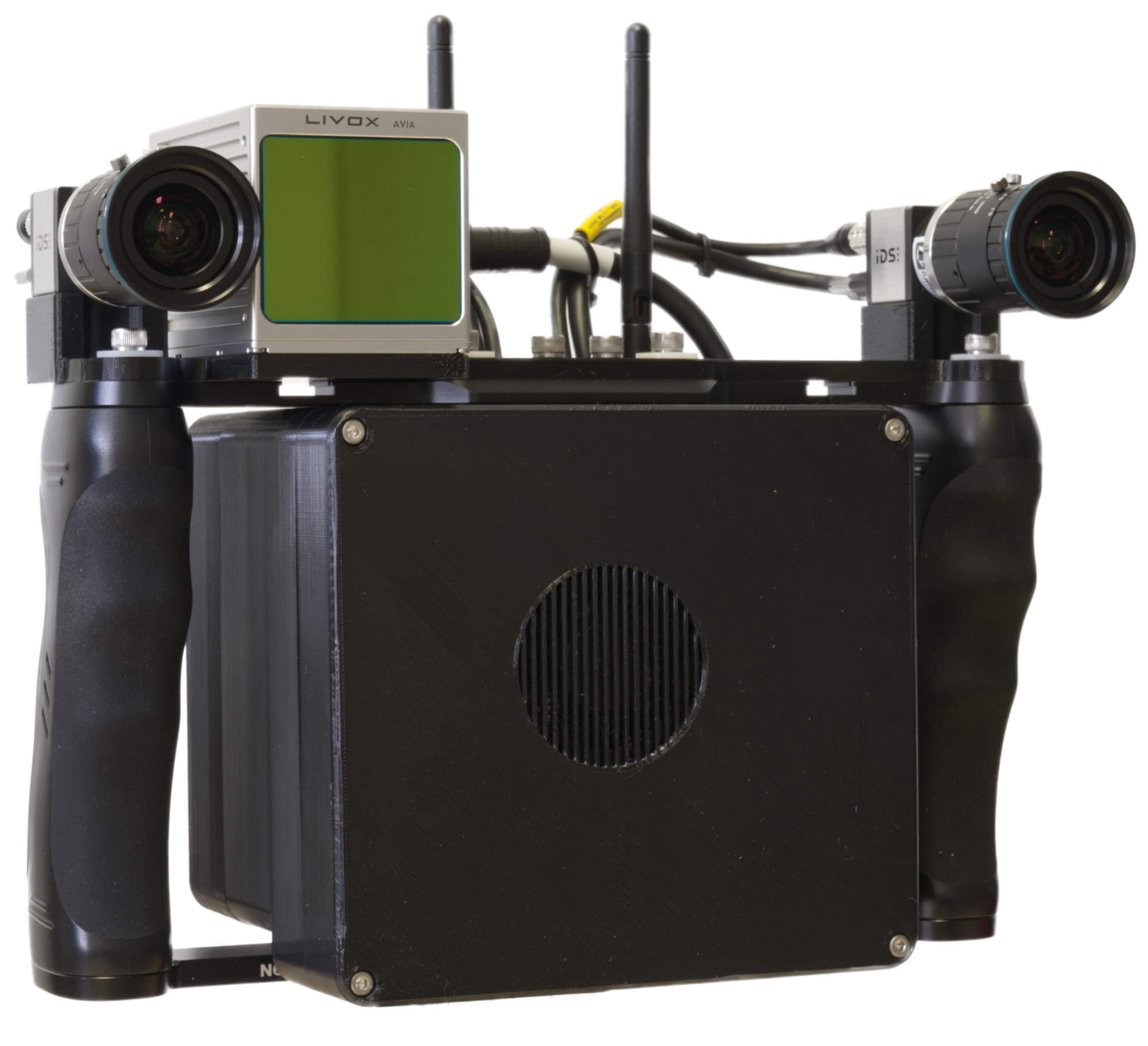}
		}
		\hfill
		\subfloat[width=0.5\linewidth][UAV RGB-L]{
			\centering
			\includegraphics[width=0.5\linewidth]{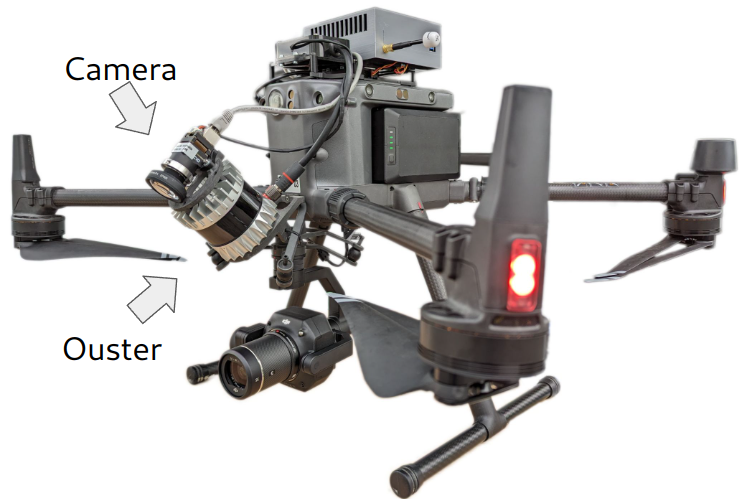}
		}
	\end{minipage}
	
	\begin{minipage}{\linewidth}
		\subfloat[width=0.4\linewidth][Azure Kinect RGB-D]{
			\centering
			\includegraphics[width=0.4\linewidth]{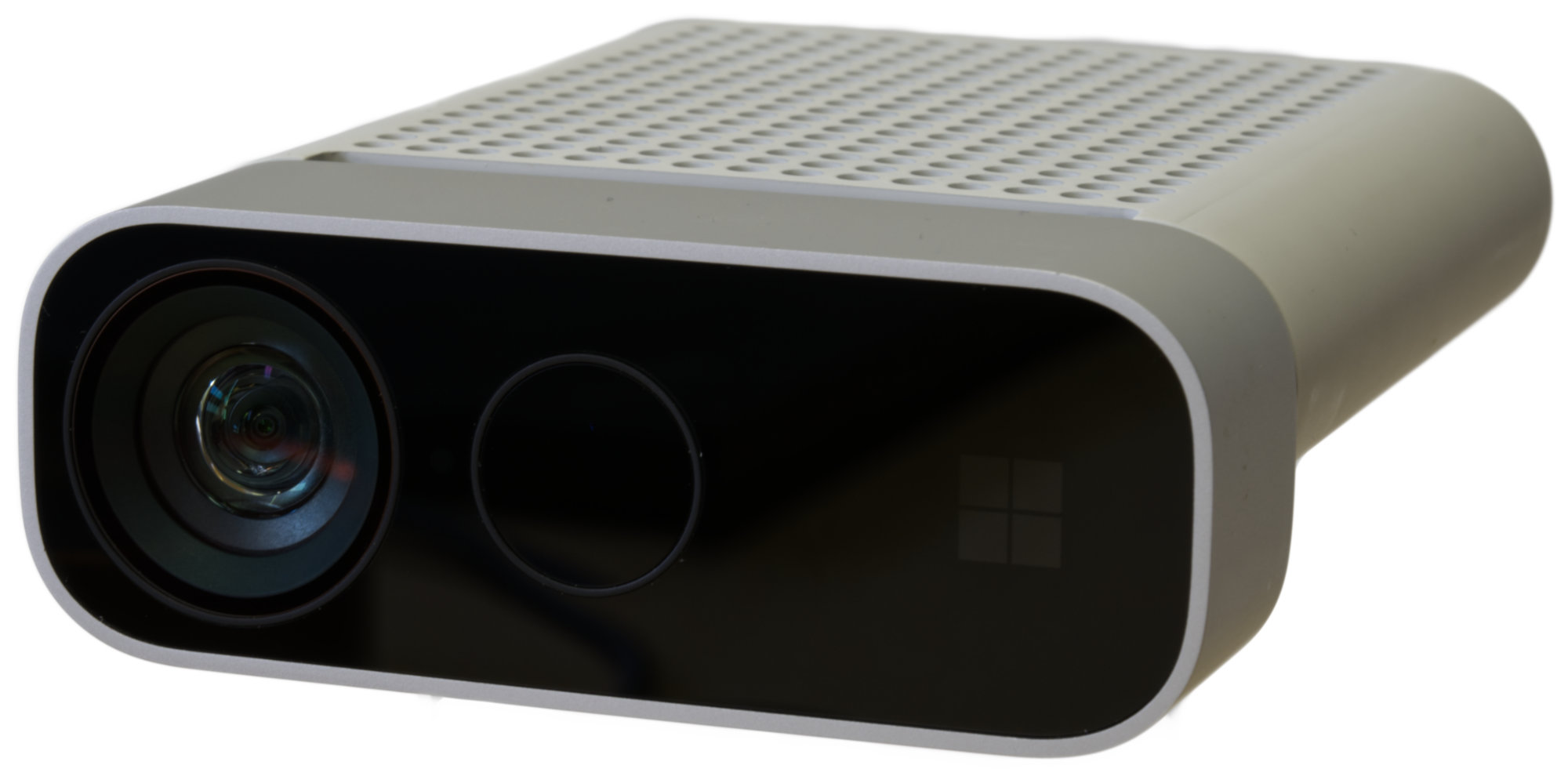}
		}
		\hfill
		\subfloat[width=0.4\linewidth][Redmi Phone RGB]{
			\centering
			\includegraphics[width=0.4\linewidth]{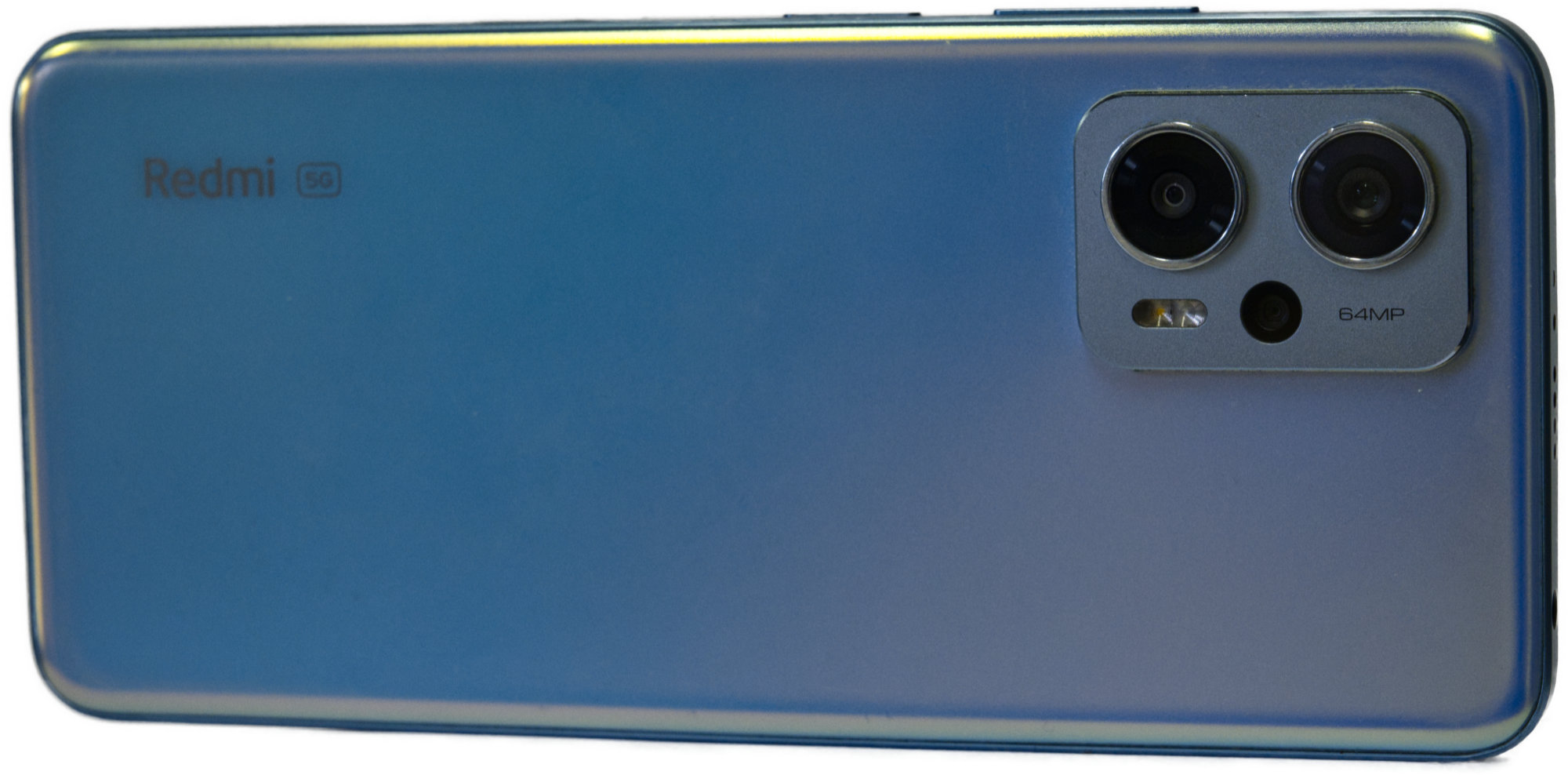}
		}
	\end{minipage}
	
	\caption{Sensors employed for capturing the datasets. See description in~Suppl.\ref{sec:sup:sensor}}
	\label{fig:sensors}
\end{figure}

\paragraph{Our Datasets}
We acquire the first dense mono SLAM purpose RGB-X dataset.
Previous RGB-D reconstruction purpose datasets contain large rotations, which is acceptable for RGB-D SLAM even with blur.
However, for dense monocular purposes, sufficient parallax between consecutive frames is required.
Large rotations and blurry images are detrimental to both tracking and depth estimation.
Moreover, recent SOTA dense mono SLAMs only focus on small room- or object-scale reconstruction.
%Moreover, publicly available RGB-D datasets often focus on room-scale reconstruction due to the limited range of commercial RGB-D sensors.
With our own datasets we also want to show that the proposed method scales to larger scenes. % and is applicable to real-world outdoor data capture with more challenging lighting conditions.
Therefore, we acquire our own dataset using a motion pattern with good parallax and four different sensor systems, which are shown in Fig.~\ref{fig:sensors} (For more details see~Suppl.\ref{sec:sup:sensor}). 
The datasets ordered from small to large scale are:
\begin{itemize}[]
	\item Apartment living room using Xiaomi phone (RGB)
	\item University of W\"urzburg Robotics hall using a Kinect Azure (RGB-D) 
	\item Veitsh\"ochheim Palace captured from a handheld mapping system (RGB-L)
	\item University of W\"uzburg building complex captured from a UAV of the Center for Telematics (RGB-L)
\end{itemize}

\subsubsection{Baselines}
We compare our framework to both multi-view depth estimation (MVD) and incremental scene reconstruction (ISR) models.

For MVD, we compare with famous classic COLMAP~\cite{colmapsfm,colmapmvs} and Deep learning based methods, such as MVSNet~\cite{mvsnet}, Vis-MVSSNet~\cite{vis-mvsnet}, MVS2D~\cite{MVS2D}, DeMon~\cite{demon}, DeepV2D~\cite{deepv2d}, Robust MVD baseline~\cite{robust_mvd}, DUSt3R~\cite{dust3r} and more.
%In particular, the latest fully supervised model DUSt3R~\cite{dust3r} is with the least constraint (no intrinsic input).
%supports the closest to us. 
%While ours is not a deep learning model.

For ISR, we include Dense RGB-D-SLAM SOTAs (NICE-SLAM~\cite{zhu2022nice}, Vox-Fusion), Dense Mono SLAM SOTAs (NeRF-SLAM, DROID-SLAM, NICER-SLAM), Surface Light Field model (NSLF-OL~\cite{yuan2023online}).

\subsubsection{Evaluation Metrics}
In the MVD test, we follow RobustMVD~\cite{robust_mvd} to use Absolute Relative Error (absrel) and the Inlier Ratio to measure the estimation quality.
When ground truth (GT) scale (GT pose) was not given, we follow DUSt3R~\cite{dust3r} to align the depth to GT scale with the median.

In the ISR test, our evaluations are mainly for color, geometry, and pose.
For color, we use PSNR, SSIM, and LPIPS.
For geometry, we use accuracy, completion, and completion ratio.
For trajectory, we use ATE RMSE.
Before evaluating the geometry and trajectory, we follow NICER-SLAM to align the reconstruction and trajectories with the ICP tool of CloudCompare~\cite{cloudcompare}.
For the monocular scenario, scale is also corrected.

\subsection{Effect of Multiview Detph Estimation}
\begin{table*}[!]
		\begin{center}
  			\caption{
				\textbf{Multi-view depth evaluation} with different settings: 
				a) Classical approaches; 
				b) with poses and depth range, without alignment;
				c) absolute scale evaluation with poses, without depth range and alignment; 
				d) without poses and depth range, but with alignment;
				e) without poses, depth range and intrinsics, but with alignment.
				(Parentheses) denote training on data from the same domain. The best results for each setting are in \textbf{bold}. The overall best are underlined and bolded.% in \underline{\textbf{best}}.
			}
			\label{tab:mvd}
		\end{center}
		\vspace*{-6mm}
			\renewcommand\arraystretch{1.2}
			\setlength{\tabcolsep}{1pt} 
			\small
			\hspace{-3mm}
			\resizebox{\textwidth}{!}{
				\begin{tabular}{llcccc|rr|rr|rr|rr|rr}
					\hline
					\specialrule{1.5pt}{0.5pt}{0.5pt}
					\multicolumn{2}{l}{\multirow{2}{*}{Methods}} & GT & GT & GT & Align & \multicolumn{2}{c}{KITTI} & \multicolumn{2}{c}{ScanNet} & \multicolumn{2}{c}{ETH3D} & \multicolumn{2}{c}{DTU} & \multicolumn{2}{c}{T\&T} \\%& \multicolumn{3}{c}{Average} \\
					\cline{7-16}
					&& Pose & Range & Intrinsics &  & rel $\downarrow$ & $\tau \uparrow$ & rel $\downarrow$ & $\tau \uparrow$ & rel $\downarrow$ & $\tau \uparrow$ & rel $\downarrow$ & $\tau \uparrow$ & rel$\downarrow$ & $\tau \uparrow$\\% & rel$\downarrow$ & $\tau \uparrow$ & time (s)$\downarrow$ \\
					\specialrule{1.5pt}{0.5pt}{0.5pt}
					\multirow{2}{*}{(a)} & COLMAP~\cite{colmapsfm,colmapmvs} & $\checkmark$ & $\times$ & $\checkmark$ & $\times$ &{\bf 12.0}&{\bf 58.2} &{\bf 14.6}&{\bf 34.2}&{\bf 16.4}&{\bf 55.1}&{\bf 0.7}&{\bf 96.5}&{\bf 2.7}& 95.0 \\%&{\bf 9.3} & {\bf 67.8} & $\approx$ 3 min  \\
					&COLMAP Dense~\cite{colmapsfm,colmapmvs} & $\checkmark$&$\times$ & $\checkmark$ & $\times$ & 26.9 & 52.7 & 38.0 & 22.5 & 89.8 & 23.2 & 20.8 & 69.3 & 25.7 & 76.4 \\%& 40.2 & 48.8 & $\approx$ 3 min \\
					\hline
					\multirow{5}{*}{(b)} & MVSNet~\cite{mvsnet} & $\checkmark$ & $\checkmark$ &$\checkmark$ & $\times$ & 22.7 & 36.1 & 24.6 & 20.4 & 35.4 & 31.4 & (1.8) & $(86.0)$ & 8.3 & 73.0\\% & 18.6 & 49.4 & 0.07 \\
					& MVSNet Inv. Depth~\cite{mvsnet} & $\checkmark$ & $\checkmark$ &$\checkmark$ & $\times$ & 18.6 & 30.7 & 22.7 & 20.9 & 21.6 & 35.6 & (1.8) & $(86.7)$ & 6.5 & 74.6 \\%& 14.2 & 49.7 & 0.32\\
					& Vis-MVSSNet~\cite{vis-mvsnet} & $\checkmark$ & $\checkmark$ & $\checkmark$ & $\times$ &{\bf 9.5}&{\bf 55.4}& \bf8.9 & \bf33.5 &{\bf 10.8}&{\bf 43.3}&{\bf (1.8)} &{\bf (87.4)} &{\bf 4.1}&{\bf 87.2} \\%&{\bf 7.0} &{\bf 61.4} & 0.70 \\
					& MVS2D ScanNet~\cite{MVS2D} & $\checkmark$ & $\checkmark$ & $\checkmark$ & $\times$ & 21.2 & 8.7 & (27.2) & (5.3) & 27.4 & 4.8 & 17.2 & 9.8 & 29.2 & 4.4 \\%& 24.4 & 6.6 & \textbf{0.04} \\
					& MVS2D DTU~\cite{MVS2D} & $\checkmark$ & $\checkmark$& $\checkmark$ & $\times$ & 226.6 & 0.7 & 32.3 & 11.1 & 99.0 & 11.6 & (3.6) & (64.2) & 25.8 & 28.0 \\%& 77.5 & 23.1 & 0.05  \\
					\hline
					\multirow{8}{*}{(c)} & DeMon~\cite{demon} & $\checkmark$ & $\times$ &$\checkmark$ & $\times$ & 16.7 & 13.4 & 75.0 & 0.0 & 19.0 & 16.2 & 23.7 & 11.5 & 17.6 & 18.3 \\%& 30.4 & 11.9 & 0.08  \\
					& DeepV2D KITTI~\cite{deepv2d} & $\checkmark$ & $\times$ &$\checkmark$ & $\times$ & (20.4) & (16.3) & 25.8 & 8.1 & 30.1 & 9.4 & 24.6 & 8.2 & 38.5 & 9.6 \\%& 27.9 & 10.3  & 1.43\\
					& DeepV2D ScanNet~\cite{deepv2d} & $\checkmark$ & $\times$ &$\checkmark$ & $\times$ & 61.9 & 5.2 & (3.8) & (60.2) & 18.7 & 28.7 & 9.2 & 27.4 & 33.5 & 38.0 \\%& 25.4 & 31.9  & 2.15 \\
					& MVSNet~\cite{mvsnet} & $\checkmark$ & $\times$ &$\checkmark$ & $\times$ & 14.0 & 35.8 & 1568.0 & 5.7 & 507.7 & 8.3 & (4429.1) & (0.1) & 118.2 & 50.7 \\%& 1327.4 & 20.1 & 0.15 \\
					& MVSNet Inv. Depth~\cite{mvsnet} & $\checkmark$ & $\times$ &$\checkmark$ & $\times$ & 29.6 & 8.1 & 65.2 & 28.5 & 60.3 & 5.8 & (28.7) & (48.9) & 51.4 & 14.6 \\%& 47.0 & 21.2 & 0.28  \\
					& Vis-MVSNet \cite{vis-mvsnet} & $\checkmark$ & $\times$ &$\checkmark$ & $\times$ & 10.3 &{\bf 54.4}& 84.9 & 15.6 & 51.5 & 17.4 & (374.2) & (1.7) & 21.1 & 65.6\\% & 108.4 & 31.0 &0.82 \\
					& MVS2D ScanNet~\cite{MVS2D} & $\checkmark$ &$\times$ &$\checkmark$ &$\times$ & 73.4 & 0.0 & (4.5) & (54.1) & 30.7 & 14.4 & 5.0 & 57.9 & 56.4 & 11.1\\% & 34.0 & 27.5 & \textbf{0.05}  \\
					& MVS2D DTU~\cite{MVS2D} &$\checkmark$ & $\times$ & $\checkmark$ & $\times$ & 93.3 & 0.0 & 51.5 & 1.6 & 78.0 & 0.0 & (1.6) & (92.3) & 87.5 & 0.0 \\%& 62.4 & 18.8 & 0.06 \\
					& Robust MVD Baseline~\cite{robust_mvd} &$\checkmark$&$\times$&$\checkmark$&$\times$ &{\bf 7.1} & 41.9 & {\bf 7.4}&{\bf 38.4}&{ 9.0}&{ 42.6}&{\bf 2.7}&{82.0}&{5.0}&{ 75.1}\\%&{\bf 6.3} &{\bf 56.0} & 0.06\\
					& \bf{U$^2$-MVD} (Ours) & $\checkmark$&$\times$&$\checkmark$&$\times$&23.6 &31.8& 27.6 & 21.8 &\bf 2.6&\bf 43.7 &2.82 &\bf 81.1 &\underline{\bf 2.26} &\underline{\bf  89.2}\\
					\hline
					\multirow{4}{*}{(d)} & DeMoN~\cite{demon} &$\times$&$\times$ &$\checkmark$& $\|\mathbf{t}\|$ & 15.5 & 15.2 & 12.0 & 21.0 & 17.4 & 15.4 & 21.8 & 16.6 & 13.0 & 23.2 \\%& 16.0 & 18.3 & 0.08 \\
					& DeepV2D KITTI~\cite{deepv2d} &$\times$&$\times$&$\checkmark$&med& (3.1) & (74.9) & 23.7 & 11.1 & 27.1 & 10.1 & 24.8 & 8.1 & 34.1 & 9.1\\% & 22.6 & 22.7 & 2.07 \\
					& DeepV2D ScanNet~\cite{deepv2d} &$\times$&$\times$&$\checkmark$& med &10.0 & 36.2 &{(4.4)} & ({\bf54.8}) & 11.8 & 29.3 & 7.7 & 33.0 & 8.9 & 46.4 \\%& 8.6 & 39.9 & 3.57  \\
					& \bf{U$^2$-MVD} (Ours) & $\times$&$\times$&$\checkmark$&med& \underline{\bf2.07}&\bf81.9& \underline{\bf3.40} &  41.4 &{\bf 1.89} & \underline{\bf 44.8}& \bf2.77&\bf72.0 & \bf2.30&\bf81.4\\
					\hline
					\multirow{4}{*}{(e)}& {{DUSt3R} 224-NoCroCo~\cite{dust3r}} &$\times$&$\times$&$\times$&med& 15.14&21.16 & 7.54&40.00 & 9.51&40.07 & 3.56&62.83 & 11.12&37.90 \\%& 9.37 & 40.39 & \textbf{0.05} \\
					& {{DUSt3R} 224}~\cite{dust3r}     &$\times$&$\times$&$\times$&med& 15.39&26.69 & (5.86)&(50.84) & 4.71&61.74 &{\bf 2.76}&{\bf 77.32}& 5.54&56.38 \\%& 6.85&54.59 & \textbf{0.05} \\
					& {{DUSt3R} 512}~\cite{dust3r} &$\times$&$\times$&$\times$&med&{ 9.11}&{ 39.49}& ({\bf4.93})&({ \bf60.20}) &{ 2.91}&{\bf 76.91}& 3.52&69.33 &{ 3.17}&{ 76.68}\\%&{\bf 4.73}&{\bf 64.52} & 0.13\\
					& \bf{U$^2$-MVD} (Ours) & $\times$&$\times$&$\times$&med& \bf2.08 &\underline{\bf 82.01} & 7.19 & 22.17 & \underline{\bf 1.87}  & 50.14  & 15.61 & 33.74 & \bf 2.46 &  \bf 81.00 \\
					\specialrule{1.5pt}{0.5pt}{0.5pt}\\
			\end{tabular}}
		\vspace*{-6mm}
			\normalsize
	\end{table*}

One of the most important components of SceneFactory is the depth estimator. 
We follow the Multiview Depth Estimation (MVD) benchmark, RobustMVD~\cite{robust_mvd}, to evaluate the depth result on 5 widely used datasets (KITTI, ScanNet, ETH3D, DTU, and Tanks\&Temple).
%,  we find DUSt3R~\cite{dust3r} and add one more class (e) and one more column (GT Intrinsics) for comparison.

\subsubsection{Quantitative Evaluation}
% benchmark
Please find~\cref{tab:mvd} the benchmarking result.
The metrics are the Absolute Relative Error (rel or absrel) and the Inlier Ratio
($\tau$) with a threshold of 1.03~\cite{robust_mvd}.
Since our model also supports intrinsic-free, which is not listed in~\cite{robust_mvd}, we follow DUSt3R~\cite{dust3r} to add an additional column (GT Intrinsics) to indicate the requirement of intrinsics besides the whole benchmark. Both of ours and DUSt3R do not require GT pose, depth range, and intrinsics (in setting (e)).
To have a better evaluation of our model, we add intrinsics and also test our model in both (c) with GT pose and (d) without GT pose setting.

First of all, in (e) setting (without the input of GT pose, depth range, or intrinsics), on the indoor and outdoor datasets ETH3D, Tanks\&Temple and KITTI, our model can already outperform other approaches over all (a-e) folds.
On the indoor dataset ScanNet, our scores do not exceed DUSt3R.
What's more, our model performs particularly poorly on the object level (DTU dataset). 
We believe that this is due to the fact that the estimation of intrinsic properties adds additional challenges to the MVD tasks. 
%Small differences on intrinsics can easily destroy the final depth.
On ScanNet, the images originally contained strong motion blur, which is also detrimental to the estimation of intrinsics.
While on the object level dataset, for example DTU, small intrinsic errors will cause more noticeable estimation errors, sicne the camera is close to the object.
Please note when we also input intrinsics (in (d) setting), our model easily achieves overall best result on all datasets over all (a-e) folds. 
But, when we further input GT pose (in (c) setting), our performance is dropped on KITTI and ScanNet datasets. 
While on the rest, the performances are still maintained.
This is because ETH3D, DTU and T\&T's GT poses are captured from the scanning system or COLMAP, which are good enough for multiview depth estimation purpose.
However, KITTI's GT pose is from GPS, ScanNet's GT pose is from RGB-D SLAM, BundleFusion~\cite{dai2017bundlefusion}.
Which is more for tracking and mapping purposes.
Therefore, if the GT pose is not available, optimizing both (pose and depth) will certainly exceed optimizing with a non-GT GT pose input.
%This is because ScanNet was originally acquired for RGB-D SLAM, which contains large rotation and small translation.
%The reference and source images do not have a good triangulation relationship.
%The other thing is that ScanNet's image contains strong motion blur, which is harmful to dense matching based models like our model.
%Nevertheless, our model still outperforms the matching-based baseline, RobustMVD.

%We believe this is because in our model, none of the optical flow, monodepth, or deep covariance is trained on the object.

%While differently, DUSt3R requires training on eight indoor\&outdoor dataset\footnote{DUSt3R is trained on Habitat, MegaDepth, ARK-itScenes, MegaDepth, Static Scenes 3D,
	%Blended MVS, ScanNet++, CO3D-v2 and Waymo.}, which includes indoor, outdoor, synthetic, real-world, object-centric and more.

%Please find when we also input the intrinsics (in (d) setting), our model easily achieves the overall best.

%Our model relies on dense bundle adjustment, which we consider is less affected by the deep prior.

From the condition that overall \underline{BEST} in~\cref{tab:mvd} are mainly ours, we find our model can flexibly support all input settings, while in a reliable depth estimation performance. 
This provides us a good basis for our entire SceneFactory pipeline.
%, from a practical point of view, DUSt3R 1) is not guaranteed to work on custom scene dataset (will be shown in~\cref{sec:exp:mvd:custom}), 2) DUSt3R is not able to work with given poses and intrinsics. 
%These are not a problem for our model and allow us to work with our entire SceneFactory pipeline.

% qualatitive
\subsubsection{Qualitative Evaluation}
We are particularly interested in the unconstraint setting (e) and show the estimated depth and error in~\cref{fig:mvd_result}.
Where we have attached the qualitative result of the robustMVD of the $5$ datasets in~\cref{tab:mvd} sequentially.

Please find in figure that our result has better performance on outdoor and indoor dataset, except for ScanNet.
It is consistent with the result in~\cref{tab:mvd} (e) where the intrinsic solving via dense matching of our method is fragile for ScanNet that contains strong motion blur caused by a rather old RGB camera.
%Which we consider is because of the nature of our method of dense matching. while ScanNet contains strong motion blur caused by quite old RGB camera.
However, DUSt3R is more optimized for monodepth (because during the benchmarking, DUSt3R does not benefit from more source views).
So it is not affected by this issue.

Although our average score on object dataset DTU is lower, it is only because our method fails in some cases. 
While the robustMVD selected qualitative figure is still better.

	\newcommand{\MVDImSize}{.2}
	\begin{figure*}[t!]
		\centering
		\setlength{\tabcolsep}{0.1em}
		\renewcommand{\arraystretch}{.1}
		\begin{tabular}{c c c c c}
			\toprule	
			{\large{Scene-RGB}} & {\large{DUSt3R-MVD}} & {\large{DUSt3R-absrel}}&{\large{Ours-MVD}}&{\large{Ours-absrel}} \\ 
			
			\midrule
			\includegraphics[width=\MVDImSize\linewidth]{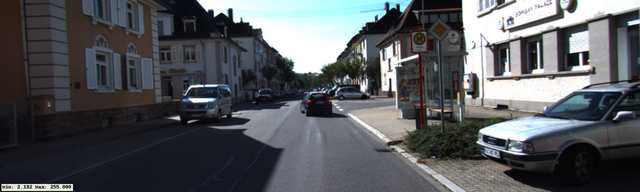}
			&\includegraphics[width=\MVDImSize\linewidth]{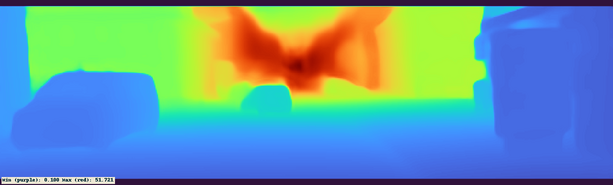}
			&\includegraphics[width=\MVDImSize\linewidth]{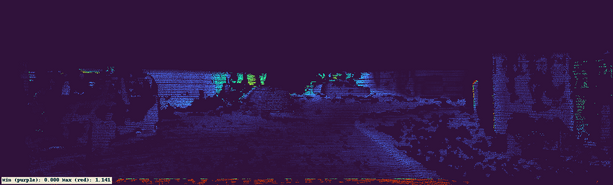}
			&\includegraphics[width=\MVDImSize\linewidth]{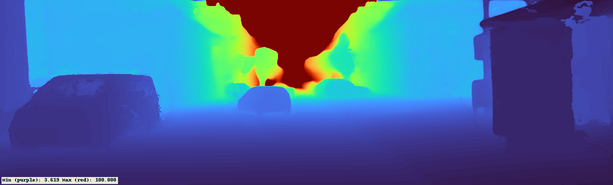}
			&\includegraphics[width=\MVDImSize\linewidth]{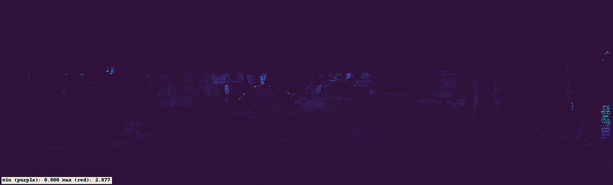}
			\\
			\midrule
			\includegraphics[width=\MVDImSize\linewidth]{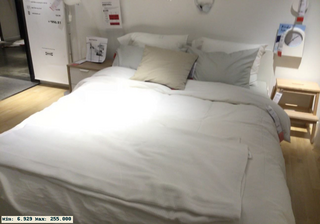}
			&\includegraphics[width=\MVDImSize\linewidth]{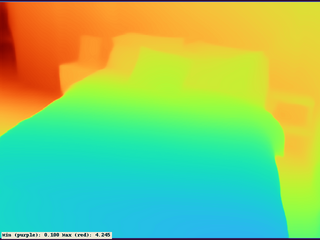}
			&\includegraphics[width=\MVDImSize\linewidth]{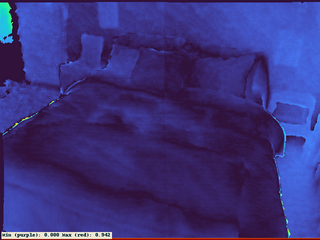}
			&\includegraphics[width=\MVDImSize\linewidth]{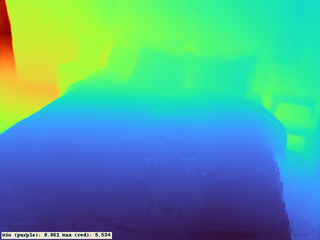}
			&\includegraphics[width=\MVDImSize\linewidth]{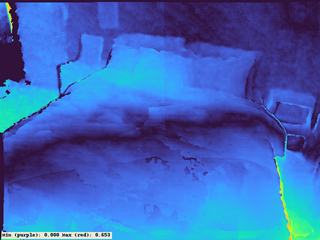}
			\\		
			\midrule
			\includegraphics[width=\MVDImSize\linewidth]{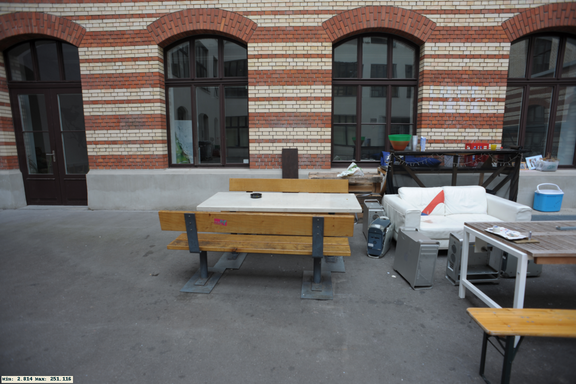}
			&\includegraphics[width=\MVDImSize\linewidth]{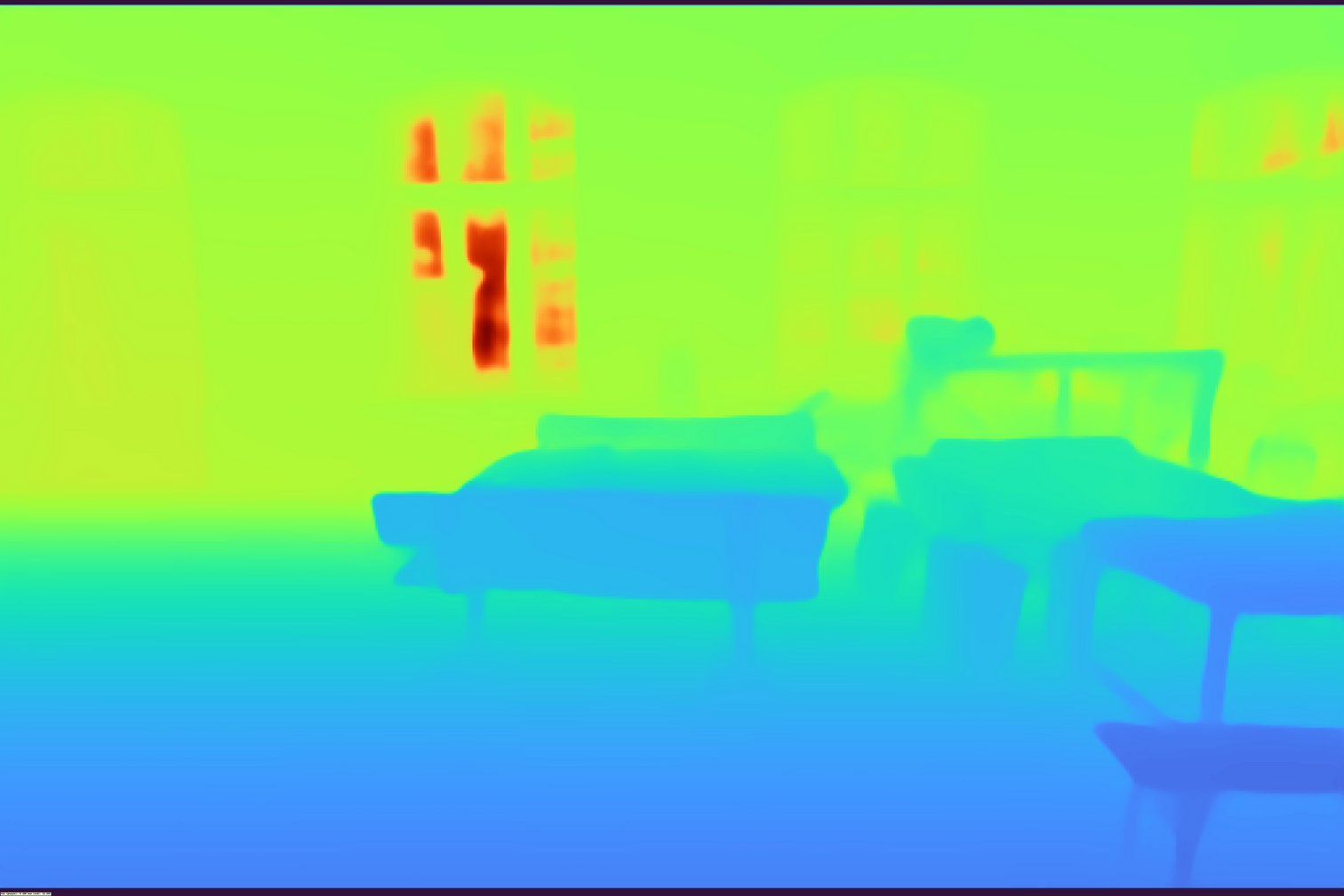}
			&\includegraphics[width=\MVDImSize\linewidth]{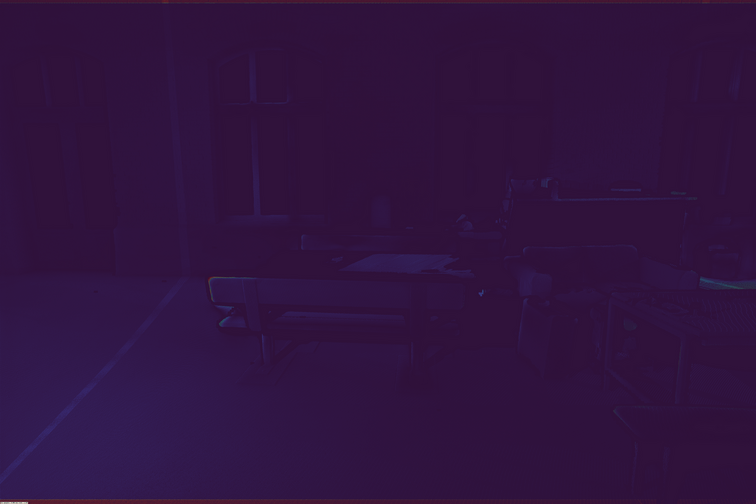}
			&\includegraphics[width=\MVDImSize\linewidth]{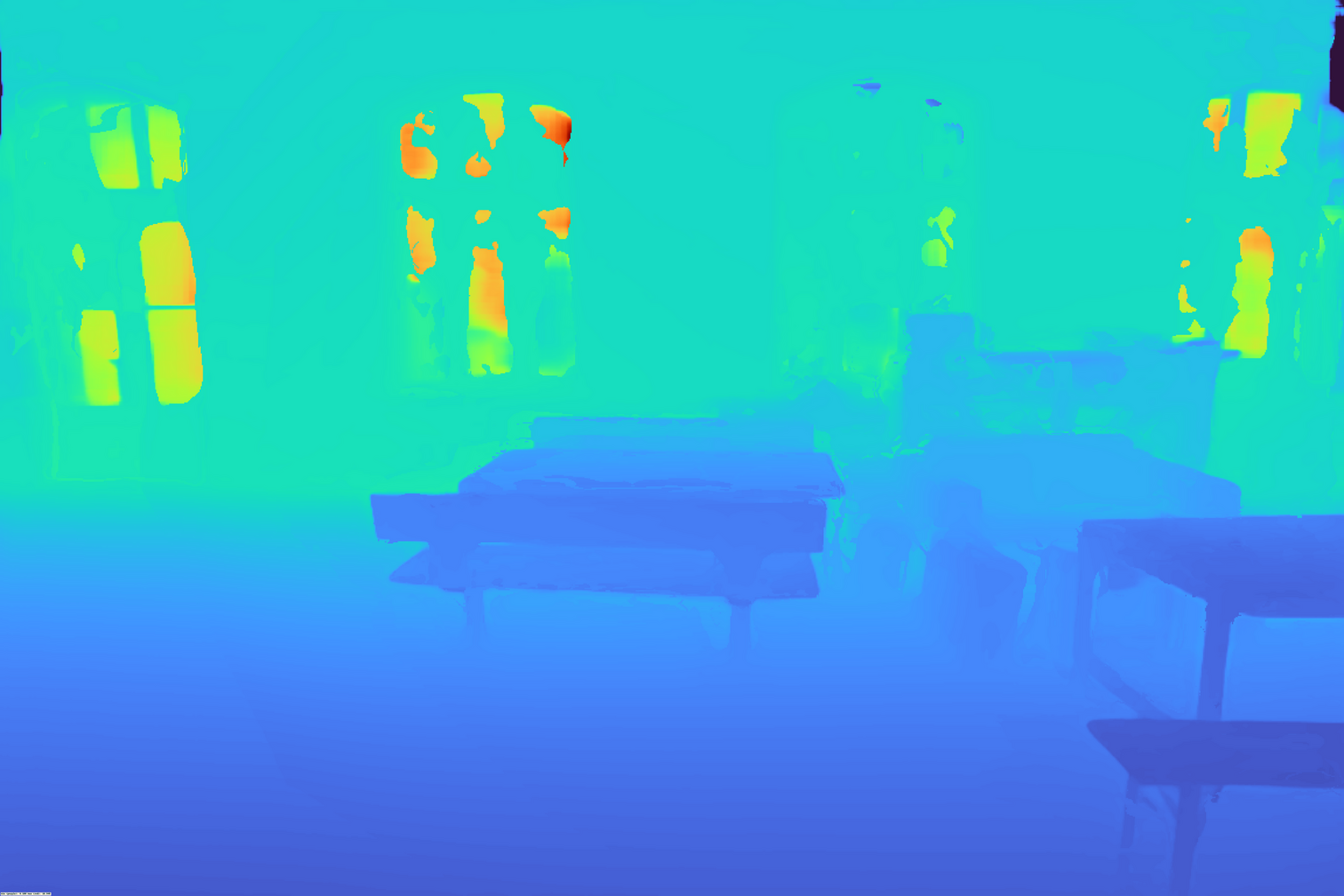}
			&\includegraphics[width=\MVDImSize\linewidth]{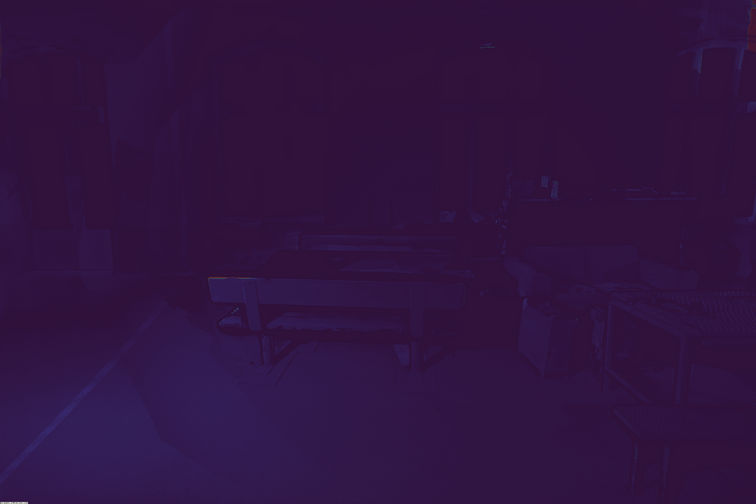}
			\\	
			\midrule
			\includegraphics[width=\MVDImSize\linewidth]{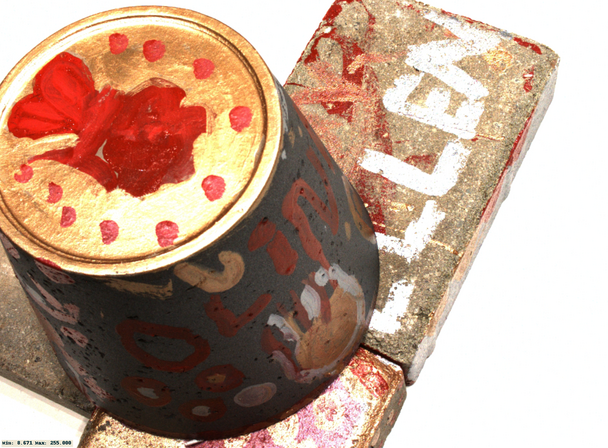}
			&\includegraphics[width=\MVDImSize\linewidth]{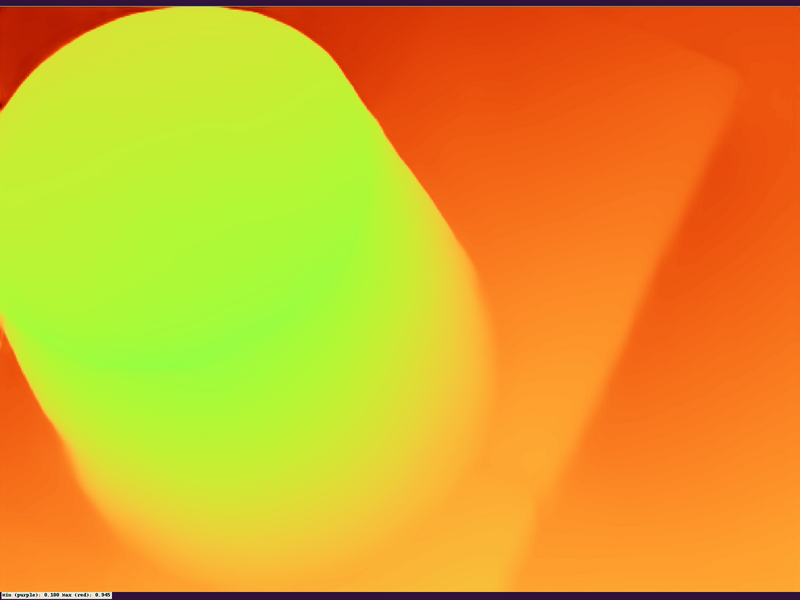}
			&\includegraphics[width=\MVDImSize\linewidth]{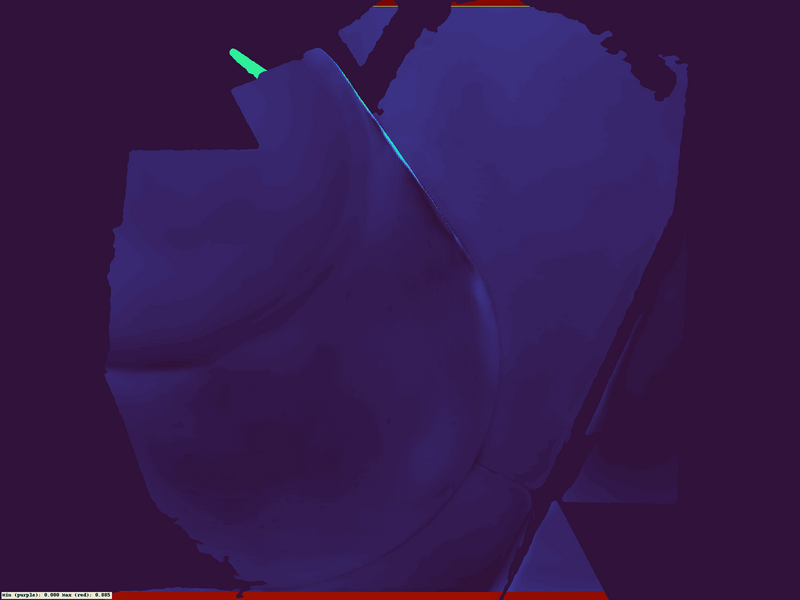}
			&\includegraphics[width=\MVDImSize\linewidth]{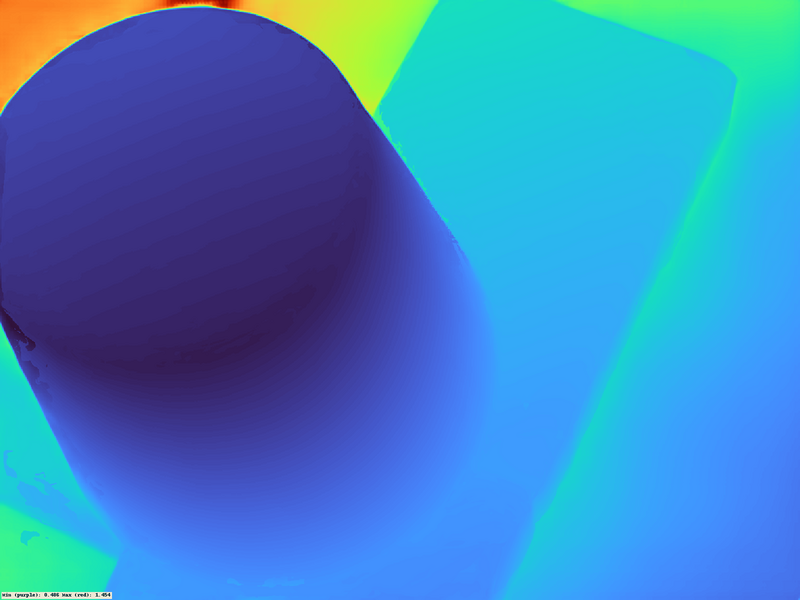}
			&\includegraphics[width=\MVDImSize\linewidth]{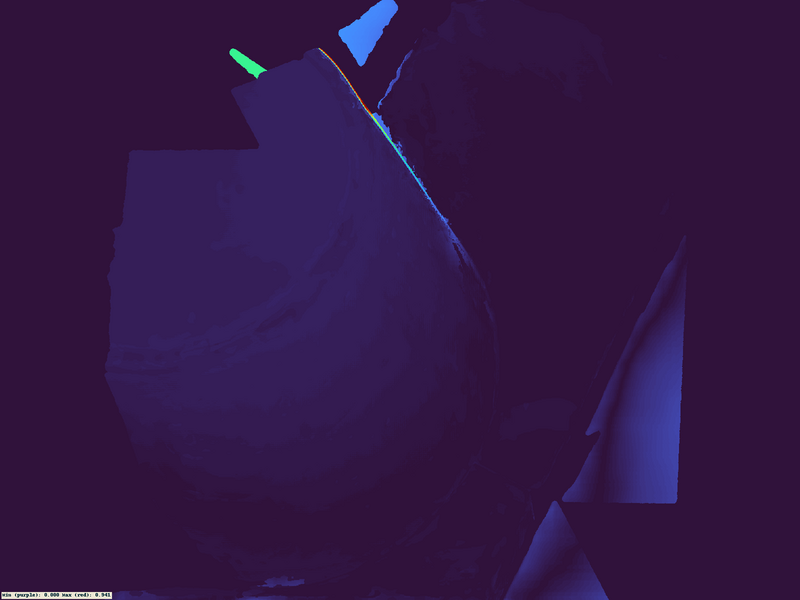}
			\\	
			\midrule
			\includegraphics[width=\MVDImSize\linewidth]{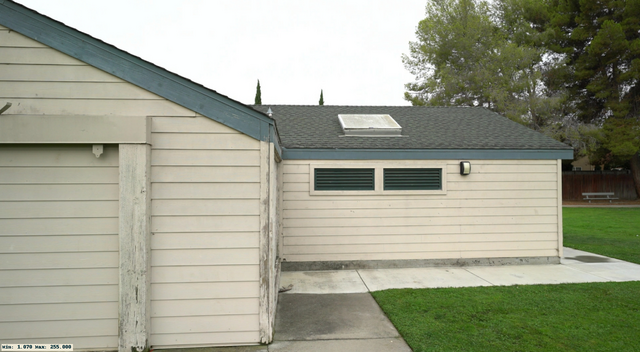}
			&\includegraphics[width=\MVDImSize\linewidth]{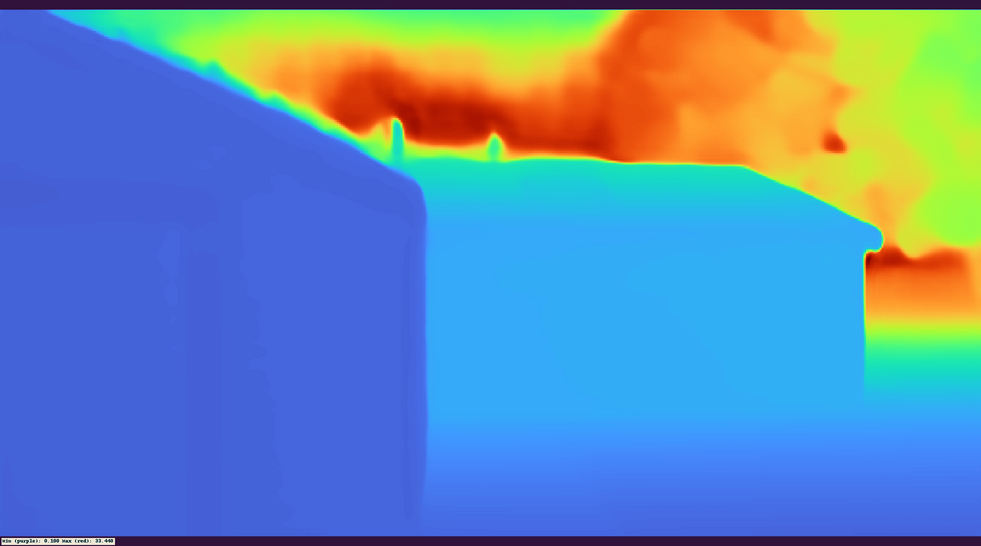}
			&\includegraphics[width=\MVDImSize\linewidth]{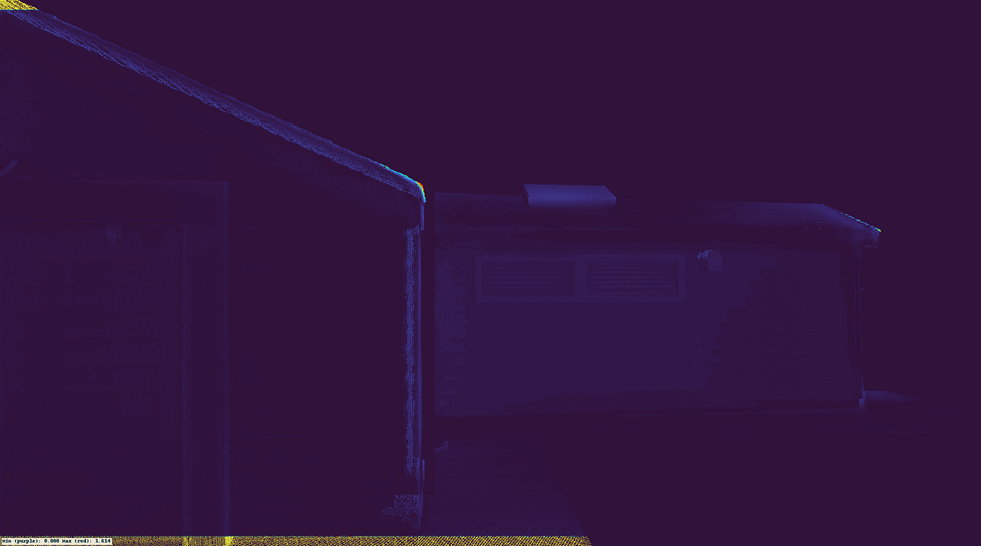}
			&\includegraphics[width=\MVDImSize\linewidth]{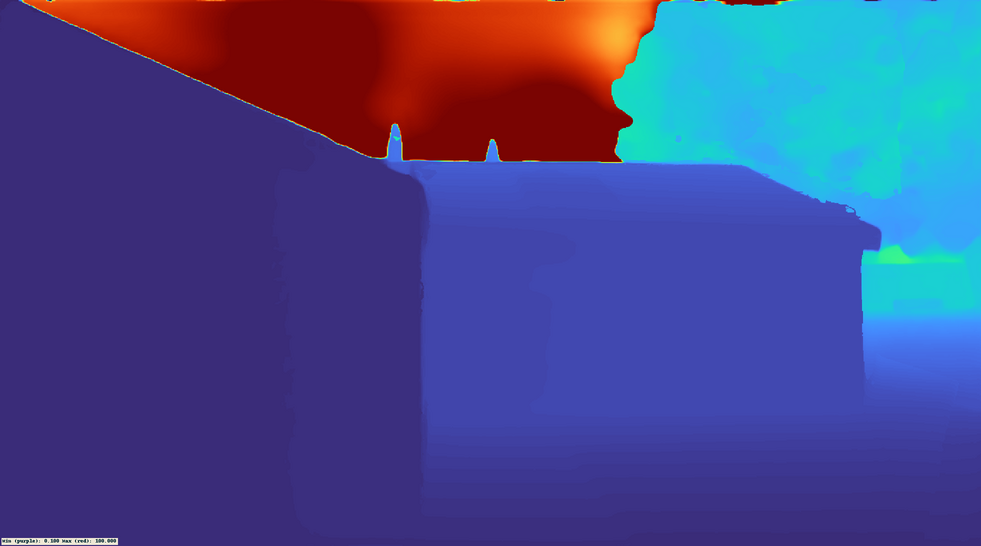}
			&\includegraphics[width=\MVDImSize\linewidth]{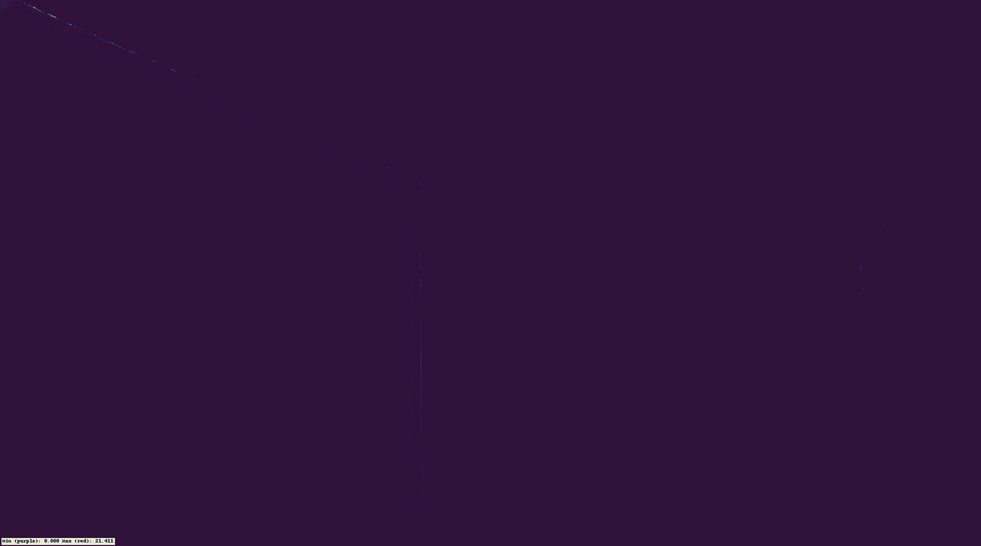}
			\\					
			\bottomrule
		\end{tabular}
		%\captionof{figure}
		\caption{Result of MVD. Examples from top row to bottom are from the KITTI, ScanNet, ETH3D, DTU and T\&T datasets. From left to right are scene image, depth prediction and absolute relative error (absrel) error of DUSt3R, and depth prediction and absrel error of our model.
		For depth and absrel images, higher values are warmer, lower values are colder colors.}
		\label{fig:mvd_result}
		\vspace{-.5cm}
	\end{figure*}

\begin{figure}[htbp!]
	\centering
	
	\subfloat[width=.5\linewidth][\footnotesize	Ames Room (ref frame)]{
		\centering
		\includegraphics[width=.5\linewidth]{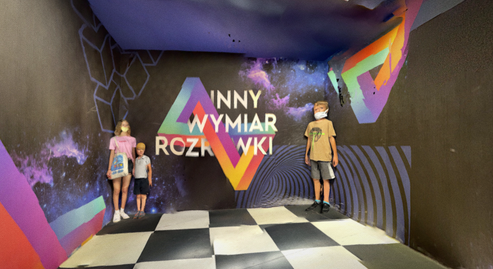}
	}
	\subfloat[width=.5\linewidth][\footnotesize	Ames Room (src frame)]{
		\centering
		\includegraphics[width=.5\linewidth]{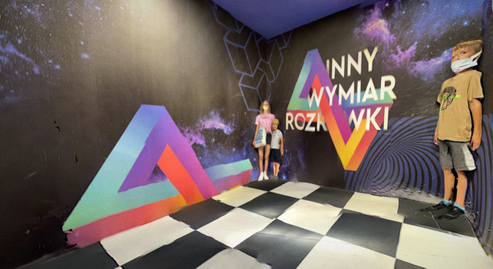}
	}\\	\vspace{-.2cm}
	\caption{MVS input of Ames room. The true shape please find webpage $^4$.}
	\label{fig:ames_imgs}
\end{figure}
\begin{figure}[htbp!]
	\centering
	
	%	\subfloat[width=.5\linewidth][\footnotesize	Ames Room (ref frame)]{
		%		\centering
		%		\includegraphics[width=.5\linewidth]{ims/ames_room/ref}
		%	}
	%	\subfloat[width=.5\linewidth][\footnotesize	Ames Room (src frame)]{
		%		\centering
		%		\includegraphics[width=.5\linewidth]{ims/ames_room/src}
		%	}\\	\vspace{-.2cm}
	\subfloat[width=.5\linewidth][{\footnotesize	DUSt3R's Frontview}]{
		\centering
		\includegraphics[width=.5\linewidth]{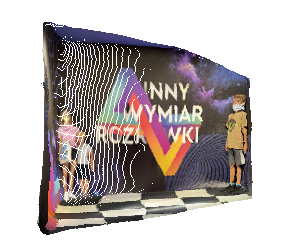}
	}
	\subfloat[width=.4\linewidth][\footnotesize	DUSt3R's Topview]{
		\centering
		\includegraphics[width=.4\linewidth]{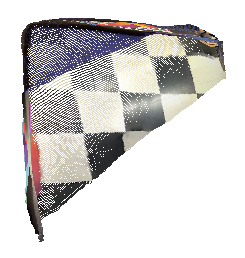}
	}\\	
	\subfloat[width=.5\linewidth][{\footnotesize	Ours Frontview}]{
		\centering
		\includegraphics[width=.5\linewidth]{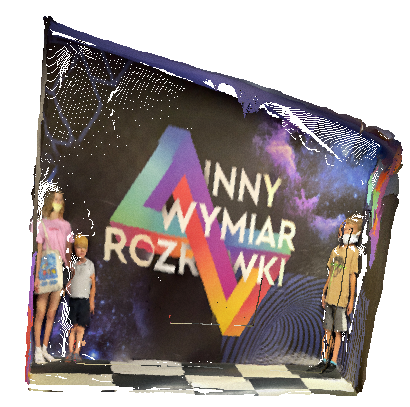}
	}
	\subfloat[width=.5\linewidth][\footnotesize	Ours Topview]{
		\centering
		\includegraphics[width=.5\linewidth]{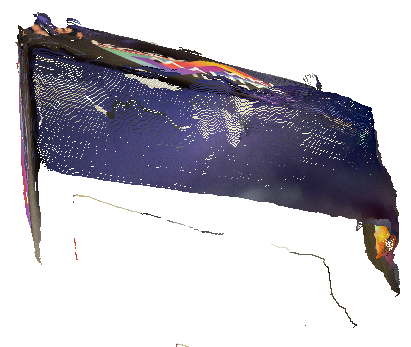}
	}\\		
	
	\caption{MVD on Ames room. (a) and (b) shows the frontview and the topview of DUSt3R's prediction. Which falsely predicts the distorted space as a rectangular room. 
		(c) and (d) shows the frontview and the topview of our prediction. Which well predicts the real shape.}
	\label{fig:ames}
\end{figure}

\subsubsection{Demonstration on Custom Data}
\label{sec:exp:mvd:custom}
% my captured data
As we claimed earlier, DUSt3R is not guaranteed to work due to the limitation of the training dataset.
In addition, due to the fact that 1) DUSt3R does not improve with increasing number of source images in the RobustMVD benchmark and 2) performs similarly with and without source images. 
We suggest that the SOTA DUSt3R is mainly based on its strength in monocular depth.

To verify this, we test the vision illusion that also happens to a human eye vision, the Ames Room\footnote{\url{https://en.wikipedia.org/wiki/Ames_room}}.
Here we show the data of the Ames Room in~\cref{fig:ames_imgs}. 
Where we capture the image from Sketchfab resource\footnote{\url{https://sketchfab.com/3d-models/ames-room-iphone-3d-scan-b59a0dcf49de4df2a50104abf3eab7e4} }
Ames Room to create the illusion of a distorted room.
Where typically shows incorrect relative scales on the left and right corners.
DUSt3R, as we suspected, shows the Ames room as a normal room, as in \cref{fig:ames}.
Our model, on the other hand, truly reproduces the distorted space.

The above tests show that our model not only scores well, but is also more reliable.

%As in the previous SLF test, we mitigate the effect of the tracker and the reconstructor in this depth estimation test. 
%Image and pose $(\V G, \V I_{rgb})$ are all available in the dependency graph (~\cref{fig:dependency}) for a mono recon application.
%For comparison, we transform the depth frames directly into a point cloud.
%
%We compare with our closest work DROID-SLAM under the same setting to us.
%To have a fair comparison, we remove the global optimization of DROID-SLAM at the very end.

%\input{tex/experiment/SLF}
\newcommand{\ml}[2]{\multicolumn{#1}{l}{#2}}
\newcommand{\mc}[2]{\multicolumn{#1}{c}{#2}}
	\begin{table*}[t] %[htbp]
		\caption {Full model comparison on Replica sequences. Header indicates scene names.}
		\begin{adjustwidth}{0pt}{0pt}
			\centering
			{\small
				\begin{tabular}{llcccccccc}
					\toprule
					&& {Office0} & {Office1}  &{Office2} &{Office3} &{Office4} &{Room0} &{Room1} &{Room2}              \\	
					\midrule
					\multirow{3}{*}{NICE-SLAM~\cite{zhu2022nice}} &
					PSNR $\uparrow$
					&28.38&30.68&23.90&24.88&25.18&23.46&23.97&25.94\\
					
					&SSIM $\uparrow$  &0.908&\bf0.935&\bf0.893&\bf0.888&\bf0.902&0.798&0.838&0.882 \\
					&LPIPS $\downarrow$ &0.386&0.278&0.330&0.287&0.326&0.443&0.401&0.315\\
					%&LPIPS$_{vgg}$$ \downarrow$ &0.455&0.403&0.433&0.405&0.430&0.496&0.486&0.451
					\midrule
					\rowcolor{colorSep}
					\ml{3}{\textit{\textbf{Surface Light Field methods}}} & \mc{6}{}\\
					\multirow{3}{*}{NSLF-OL\cite{yuan2023online}+DI-Fusion\cite{huang2021di}} &
					PSNR $\uparrow$
					&28.59&26.70&21.10&21.89&25.74&23.24 &25.68&24.88\\
					&SSIM $\uparrow$  &0.913 &0.879&0.863&0.847&0.893&0.816&0.883&0.888 \\
					&LPIPS $\downarrow$ &0.371 &0.497&0.362&0.368&0.401&0.371&0.308&0.330\\
					%&LPIPS$_{vgg}$$ \downarrow$ &0.395 &0.446&0.421&0.416&0.425&0.417&0.415&0.422\\
					\midrule
					\multirow{3}{*}{\textbf{SceneFactory (Ours)}} &
					PSNR $\uparrow$&\bf 33.38&\bf 31.89 &\bf 24.84& \bf25.39&\bf 31.14&\bf 27.98&\bf 29.51&\bf 30.64\\
					&SSIM $\uparrow$  &\bf0.921&0.893&0.837& 0.873&0.893&\bf0.858&\bf0.883&\bf0.898\\
					&LPIPS $\downarrow$&\bf0.167&\bf0.267&\bf0.208& \bf0.129&\bf0.178&\bf0.111&\bf0.149&\bf0.145\\
					
					\bottomrule
				\end{tabular}
			}
			\label{tab:slf_comp_rep}
		\end{adjustwidth}
                \vspace*{-4mm}
	\end{table*}

	\newcommand{\scannetImSize}{.3}
	\begin{figure*}[htbp!]
		\centering
		\setlength{\tabcolsep}{0.1em}
		\renewcommand{\arraystretch}{.1}
		\begin{tabular}{c  c c }
			\toprule	
			{\large{NSLF-OL}} & {\large{SceneFactory}} &{\large{Ground Truth}}  \\ 
			
			\midrule
			\includegraphics[width=\scannetImSize\linewidth]{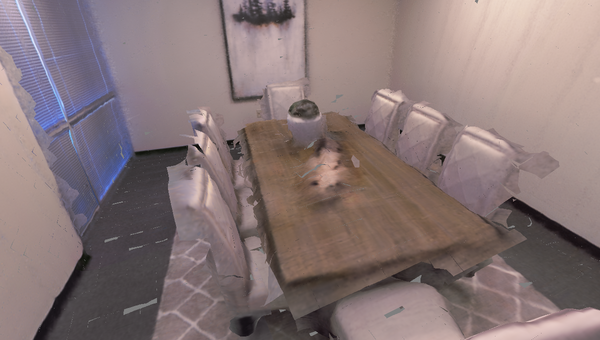}
			&\includegraphics[width=\scannetImSize\linewidth]{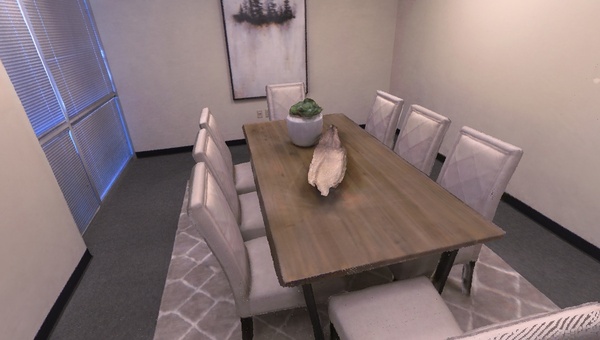}
			&\includegraphics[width=\scannetImSize\linewidth]{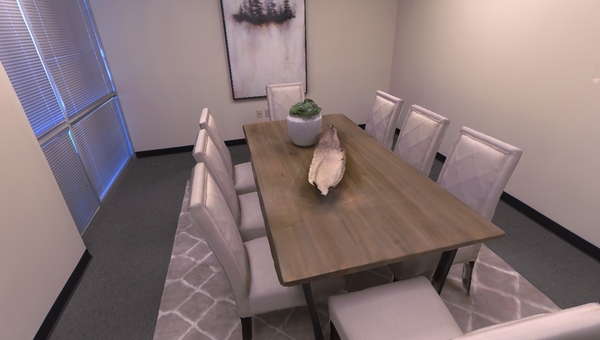}
			\\
			\midrule
			
			\includegraphics[width=\scannetImSize\linewidth]{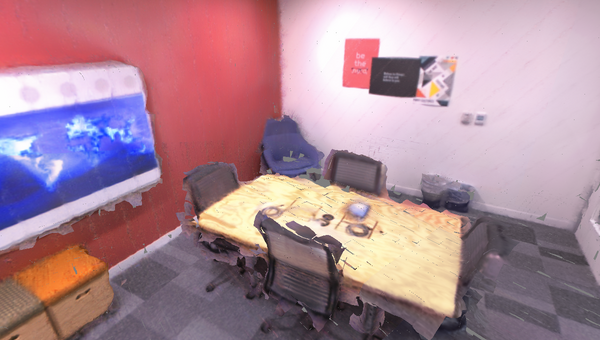}
			&\includegraphics[width=\scannetImSize\linewidth]{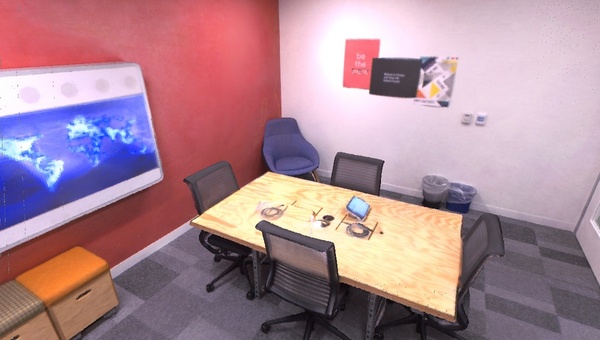}
			&\includegraphics[width=\scannetImSize\linewidth]{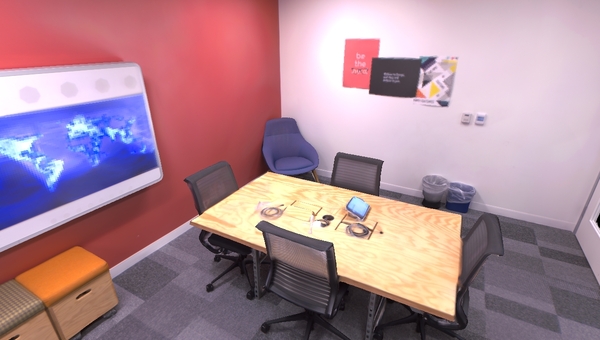}
			\\		
			\midrule
			
			\includegraphics[width=\scannetImSize\linewidth]{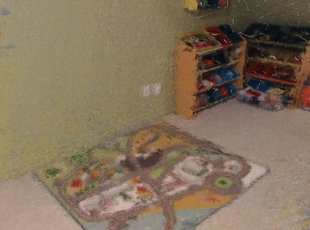}
			&\includegraphics[width=\scannetImSize\linewidth]{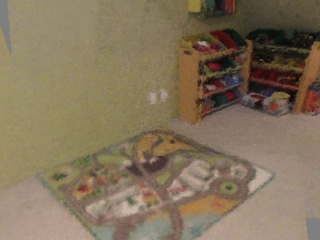}
			&\includegraphics[width=\scannetImSize\linewidth]{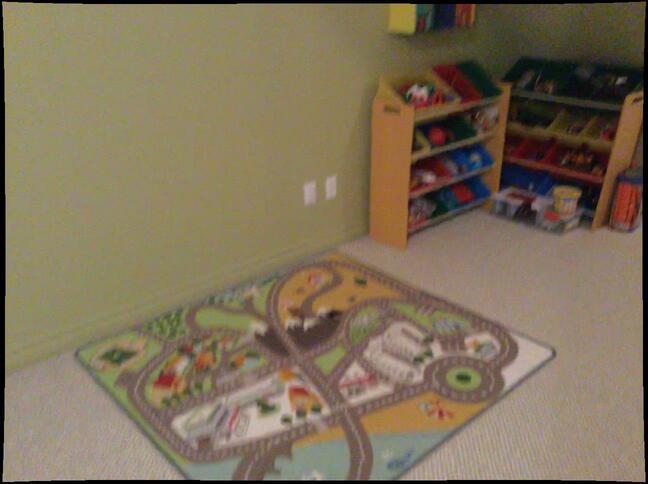}
			\\	
			\midrule
			\includegraphics[width=\scannetImSize\linewidth]{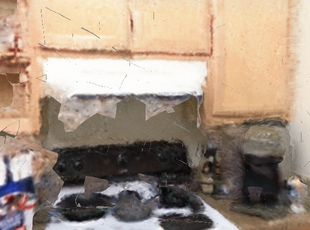}
			&\includegraphics[width=\scannetImSize\linewidth]{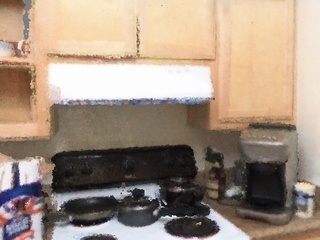}
			&\includegraphics[width=\scannetImSize\linewidth]{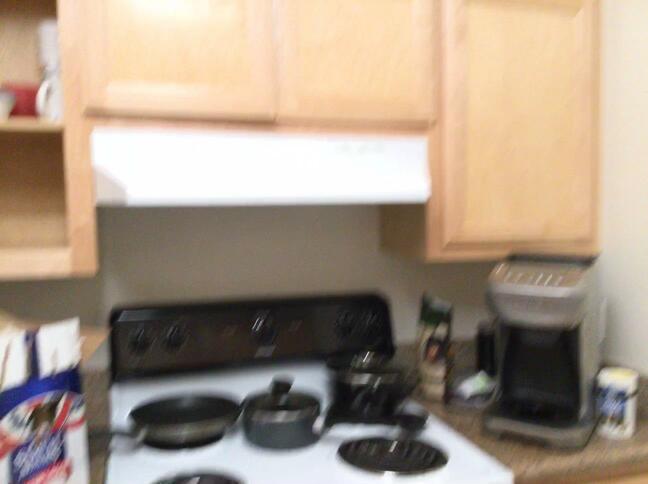}
			\\			
			%		\midrule
			%		
			%		\includegraphics[width=\scannetImSize\linewidth]{ims/SLF/nslfol/8b5caf3398_10.png}
			%		&\includegraphics[width=\scannetImSize\linewidth]{ims/SLF/ours/8b5caf3398_10.jpg}
			%		&\includegraphics[width=\scannetImSize\linewidth]{ims/SLF/gt/8b5caf3398_10.jpg}
			%		\\	
			%		\midrule
			%		
			%		\includegraphics[width=\scannetImSize\linewidth]{ims/SLF/nslfol/b20a261fdf_6.png}
			%		&\includegraphics[width=\scannetImSize\linewidth]{ims/SLF/ours/b20a261fdf_6.jpg}
			%		&\includegraphics[width=\scannetImSize\linewidth]{ims/SLF/gt/b20a261fdf_6.jpg}
			%		\\	

			\bottomrule
		\end{tabular}
		%\captionof{figure}
		\caption{Result of Surface Light Field test on Replica and ScanNet dataset.}
		\label{fig:recons:SLF_demo}
\vspace*{-4mm}
	\end{figure*}

\subsection{Evaluation on Surface Light Fields (SLF)}
In the test above, we generate depth from images. 
In this subsection, we generate a color image and explore the importance of our surface light field model.

\subsubsection{Replica Test}
To mitigate the effects of other SceneFactory blocks, image, depth, pose and intrinsics $( \V I_{rgb}, \V I_d, \V G, \V \theta)$ are all available in the dependency graph (\cref{fig:pipeline}).

First, we follow the scene-level SLF model NSLF-OL~\cite{yuan2023online} to test on the Replica dataset. 
From~\cref{tab:slf_comp_rep}, our model strongly outperforms the NICE-SLAM and SOTA SLF model, NSLF-OL on all PSNR, SSIM, LPIPS metrics.
To make the comparison more obvious, we show images in the~\cref{fig:recons:SLF_demo}, the first two rows are from the Replica dataset.
From this, we see that the main advances of our model are that on the one hand, unlike NSLF-OL, our DM-NPs itself can provide a cleaner surface, and on the other hand, our SLF recovers the color better.
Also, NSLF-OL only works alongside the surface model, while our SLF model can support surface generation by itself.
%We demonstrate the result images in~\cref{fig:recons:SLF_demo} the row $1,2$, 
%where it shows better color rendering.

%\begin{figure*}[t!]
%	\centering
%	\includegraphics[width=.6\textwidth]{example-image-a}
%	\caption{Qualitative evaluation of SLF.}
%	\label{fig:exp_slf}
%\end{figure*}

\subsubsection{ScanNet Test}

	\begin{table}[htbp]
		\caption{Surface Light Field comparison on ScanNet sequences.}
		\centering
		\begin{tabular}{ll cc}
			\toprule
			&&0568 &0164\\
			\midrule
			\multirow{4}{*}{\shortstack{NSLF-OL\\+Di-Fusion}}&
			PSNR $\uparrow$ &19.09& 18.78
			\\
			&SSIM $\uparrow$ &0.500& 0.595 \\
			&LPIPS $\downarrow$ &0.575&0.551\\
			
			\midrule 
			\multirow{4}{*}{SceneFactory}& PSNR $\uparrow$ &\bf 19.90&\bf21.85\\
			&SSIM $\uparrow$  &\bf0.573&\bf0.688\\
			&LPIPS $\downarrow$ &\bf0.536&\bf0.444\\
			
			\bottomrule
		\end{tabular}
		
		\label{tab:SLF:scannet}
	\end{table}

Following the same setup, we continue testing on ScanNet. 
Please find~\cref{fig:recons:SLF_demo}. The ScanNet image has a lot of blur on the image and more noise in the depth (due to the low-quality capture). 

However, our result is still better than NSLF-OL with better quantitative evaluation (\cref{tab:SLF:scannet}).
From the image shown in~\cref{fig:recons:SLF_demo} the row $3,4$,
we see that the structure of NSLF-OL is not well reconstructed.
Our result is much clearer.
%Moreover, it recovers even the fuzziness well.

%\subsubsection{ScanNet++ Test}
%\SLFTableScanNetPP
%
%Unlike ScanNet, ScanNet++ relies on high quality captured sequences.
%From~\cref{tab:SLF:scannetpp}, our model outperforms NSLF-OL on both seen and novel tests.
%
%However, the metrics is not sufficient to reveal the real difference.
%Therefore, we show the result in~\cref{fig:recons:SLF_demo} the row $5.6$.
%Where NSLF-OL suffers from the noisy surface and wrong prediction. 
%On the other hand, our model provides a fairly close result to the human eye.

%Besides of the synthetic result, we also test on real captured RGB-D dataset for novel synthesis, ScanNet++.

%\input{tex/experiment/SLAM}
\subsection{Evaluation of Dense SLAM}

In this section, we evaluate SceneFactory's main application, Dense SLAM, in both RGB-D and monocular settings.% following SOTA NICER SLAM.
%both monocular tracking and Dense monocular reconstruction on both synthetic and real-world scenarios.

\subsubsection{Replica Test}

	\begin{table}[htbp]
		\caption{\textbf{Camera Tracking Results on the Replica Dataset.} ATE RMSE [cm] ($\downarrow$) is used as the evaluation metric.}
		%\centering
		\footnotesize
		\setlength{\tabcolsep}{1.5pt}          %
		\renewcommand{\arraystretch}{1.05}     %
		\resizebox{.17\linewidth}{!}{
			\begin{tabular}{@{\hspace*{1em}}llcccccccccccc}
				\toprule
				& \multicolumn{1}{c}{\makecell{\tt{rm-0}}} & \multicolumn{1}{c}{\makecell{\tt{rm-1}}} &  \multicolumn{1}{c}{\makecell{\tt{rm-2}}} & \multicolumn{1}{c}{\makecell{\tt{off-0}}} & \multicolumn{1}{c}{\makecell{\tt{off-1}}} & \multicolumn{1}{c}{\makecell{\tt{off-2}}}& \multicolumn{1}{c}{\makecell{\tt{off-3}}} & \multicolumn{1}{c}{\makecell{\tt{off-4}}} & Avg. \\
				\midrule
				\rowcolor{colorSep}
				\multicolumn{12}{l}{\textit{\textbf{RGB-D input}}}\\
				NICE-SLAM  & 1.69 & 2.04 & 1.55 & 0.99 & 0.90 & 1.39 & 3.97 & 3.08 & 1.95 \\
				Vox-Fusion &  0.27 &  1.33 & 0.47 &  0.70 & 1.11 &  0.46 & \bf 0.26 &  0.58 &  0.65 \\   
				\textbf{SceneFactory} & \bf 0.20&\bf 0.12& \bf0.14&\bf 0.17&\bf 0.07& \bf0.13& 0.29& \bf0.22 &\bf0.17 \\   
				
				\midrule
				\rowcolor{colorSep}
				\multicolumn{12}{l}{\textit{\textbf{RGB input}}}\\
				COLMAP     & 0.62 & 23.7 &  0.39 &  0.33 &  0.24 & 0.79 &  0.14 & 1.73 & 3.49 \\
				NeRF-SLAM  & 17.26 & 11.94  & 15.76 & 12.75  & 10.34  & 14.52   &20.32  &14.96 & 14.73\\
				%{MonoGS} &9.32&44.85&&27.24\\			
				DIM-SLAM &0.48 &0.78& 0.35 &0.67 &0.37 & 0.36& 0.33 &0.36 &0.46\\
				DROID-SLAM &  0.34 & \bf 0.13 &  0.27 &  0.25 &  0.42 &  0.32 &  0.52 &  0.40 &  0.33 \\
				%DROID-SLAM$^*$ & \rd 0.58 & \nd 0.58 & \nd 0.38 & 1.06 & \nd 0.40 & \rd 0.70 & 0.53 & \rd 1.33 & \rd 0.70 \\
				{NICER-SLAM} & 1.36 & 1.60 & 1.14 & 2.12 & 3.23 & 2.12 & 1.42 & 2.01 & 1.88 \\
				
				\textbf{SceneFactory} &\bf 0.20 & 0.20 & \bf 0.15 & \bf 0.20 &\bf 0.12 & \bf 0.25 & \bf 0.29 &\bf 0.22 &\bf 0.20\\
				\bottomrule
		\end{tabular}}%
		
		\label{tab:replica_tracking}
		\vspace{-5pt}
	\end{table}

First, we follow Dense Mono SLAM SOTA, NICER-SLAM, to have a complete test on both color and geometric on Replica dataset.

We show the tracking performance, from~\cref{tab:replica_tracking}, SceneFactory achieves the best tracking performance overall.
While with RGB input, SceneFactory uses Momo-SLAM DPVO, but with RGB-D input, SceneFactory uses our generalized DPVO with RGB-D, which achieves the best performance.

Notably, with RGB and depth as input, 
%both NSLF-OL and our \textbf{SceneFactory} are classified as Surface Light Field models. 
%However, 
NSLF-OL requires an external reconstruction model to provide the surface prediction.
While in contrast, \textbf{SceneFactory} itself also supports surface.
We render input frames for depth and accumulate a point cloud. 
To fit in the evaluation script of NICER-SLAM, we extract mesh via Screened Poisson Surface Reconstruction for the evaluation.
\cref{tab:replica_rec} shows the performance of the reconstruction.
In addition, to demonstrate the ability to reconstruct surface color, we show the color result in~\cref{tab:replica_rec} in the view synthesis comparison.
The upper rows are with RGB-D input, while the lower rows are with RGB input.

From the tables, we learn that with RGB-D input, \textbf{SceneFactory} outperforms all SOTAs by far in both reconstruction and view generation over all scenes by a wide margin.
To better demonstrate the quality, we show a qualitative evaluation in~\cref{fig:denseslam:replica_demo}. 
Here our method works best. Please note the highly detailed texture on the quilt.

	\begin{table}[htbp]
		%\centering
		\caption{\textbf{Reconstruction Results on the Replica dataset.} Best results are highlighted as \colorbox{colorFst}{\bf first}, \colorbox{colorSnd}{second}, and \colorbox{colorTrd}{third}.
		}
		\footnotesize
		\setlength{\tabcolsep}{1pt}
		\renewcommand{\arraystretch}{1.4}      %
		\resizebox{.48\textwidth}{!}{
			\begin{tabular}{llcccccccccccccccccc}
				\toprule
				& & rm-0 & rm-1 & rm-2 & off-0 & off-1 & off-2 & off-3 & off-4 & Avg.\\
				\midrule    
				\rowcolor{colorSep}
				%\multicolumn{12}{l}{\textit{\textbf{RGB-D input}}}\\
				\ml{3}{\textit{\textbf{RGB-D input}}} & \mc{8}{}\\
				\multirow{3}{*}{\rotatebox[origin=c]{90}{NICE}}
				&  \ml{1}{Acc.[cm]$\downarrow$} &  \rd 3.53 & \rd 3.60 & \nd 3.03 & \rd 5.56 & \rd 3.35 &\rd 4.71 & \nd 3.84 & \rd 3.35 & \rd 3.87 \\
				&  \ml{1}{Comp.[cm]$\downarrow$} & \rd 3.40 & \rd 3.62 & \fs 3.27 & \nd 4.55 & \nd 4.03 & \nd 3.94 & \rd 3.99 & \rd 4.15 & \nd 3.87 \\
				&  \ml{1}{Comp.Ratio[$<\!5$cm\,\%]$\uparrow$} & \rd 86.05 & \rd 80.75 & \fs 87.23 & \rd 79.34 & \nd 82.13 & \rd 80.35 & \rd 80.55 &\rd 82.88 & \rd 82.41 \\
				\midrule
				\multirow{3}{*}{\rotatebox[origin=c]{90}{Vox-Fusion}} 
				& Acc.[cm]$\downarrow$ &  \nd 2.53 & \nd 1.69 & \rd 3.33 & \nd 2.20 &  \nd 2.21 & \nd 2.72 & \rd 4.16 & \nd 2.48 &  \nd 2.67 \\
				& Comp.[cm]$\downarrow$ &  \fs 2.81 & \fs 2.51 & \nd 4.03 & \rd 8.75 & \rd 7.36 & \rd 4.19 & \fs 3.26 & \fs 3.49 & \rd 4.55 \\
				& Comp.Ratio[$<\!5$cm\,\%]$\uparrow$ & \fs 91.52 & \fs 91.34 & \rd 86.78 & \fs 81.99 & \rd 82.03 & \fs 85.45 & \fs 87.13 & \fs 86.53 & \nd 86.59 \\
				\midrule

				\multirow{3}{*}{\rotatebox[origin=c]{90}{\textbf{Ours}}} 
				& Acc.[cm]$\downarrow$ & \fs 1.49 & \fs 1.16& \fs 1.24  
				& \fs 1.11 & \fs 0.91 & \fs 1.37 & \fs 1.62 & \fs 1.52 & \fs 1.30 \\
				
				& Comp.[cm]$\downarrow$ & \rd 3.65 & \nd 2.88 & \rd 4.31 
				& \fs 1.68 &\fs 2.23 & \fs 3.59 & \nd 3.59 & \nd 4.02 & \fs 3.24 \\
				& Comp.Ratio[$<\!5$cm\,\%]$\uparrow$ & \nd 87.61 &  \nd 90.02 & \nd 86.83 
				&  \fs93.43 & \fs90.39 &  \fs86.24 & \nd 84.98& \nd  85.07 & \fs 88.07\\
				\midrule
				\rowcolor{colorSep}
				\ml{3}{\textit{\textbf{RGB input}}} & \mc{8}{}\\
				%\multicolumn{12}{l}{\textit{\textbf{RGB input}}}\\

				\multirow{3}{*}{\rotatebox[origin=c]{90}{NeRF-S.}} & \ml{1}{Acc. [cm]$\downarrow$} &\rd11.84& 10.62& 11.86& 9.32& 14.40& 11.54& 16.31& 11.11& 12.13& \\
				& \ml{1}{Comp. [cm]$\downarrow$}&
				\nd5.63& \nd5.88& \nd9.22& \rd13.29&\rd 10.17& \nd6.95& \rd7.81& \nd5.26& \nd8.03 \\
				& Comp. Ratio [\%]$\uparrow$ &
				\rd61.13& 68.19& 47.85& 37.64& 56.17&\nd 66.20& 55.67& 61.86& 56.84\\
				
				\midrule			
				\multirow{3}{*}{\rotatebox[origin=c]{90}{DROID-S.}} & \ml{1}{Acc. [cm]$\downarrow$} &12.18& \rd8.35&\fs 3.26& \nd{3.01}& \nd{2.39}&\rd 5.66& \rd4.49& \rd4.65& \rd5.50\\
				& \ml{1}{Comp. [cm]$\downarrow$} & 8.96& \rd6.07& 16.01& 16.19& 16.20& 15.56& 9.73& \rd9.63& 12.29\\
				& Comp. Ratio [\%]$\uparrow$ & 60.07&\nd 76.20&\rd 61.62& \rd64.19&\rd 60.63& 56.78& \rd61.95& \rd 67.51&\rd 63.60\\
				\midrule

				\multirow{3}{*}{\rotatebox[origin=c]{90}{NICER-S.}} & \ml{1}{Acc. [cm]$\downarrow$} &\fs 2.53&\fs 3.93& \nd3.40& \rd5.49& \rd3.45& \fs{4.02}& \fs{3.34}& \fs{3.03}&\fs 3.65\\
				& \ml{1}{Comp. [cm]$\downarrow$} & \fs{3.04}&\fs 4.10&\fs 3.42& \fs6.09& \fs4.42&\fs \fs4.29& \fs{4.03}&\fs 3.87& \fs4.16\\
				& Comp. Ratio [\%]$\uparrow$ & \fs88.75& \fs76.61& \fs{86.1}&\nd 65.19&\fs 77.84& \fs 74.51& \fs{82.01}& \fs{83.98}& \fs79.37\\
				\midrule
				
				%\multirow{3}{*}{\rotatebox[origin=c]{90}{GO-SLAM}} 
				%%& \ml{1}{Depth L1 [cm]$\downarrow$} & -&-&-&-&-&-&-&-&4.39\\
				%& \ml{1}{Acc. [cm]$\downarrow$} &4.60&\nd{3.31}&3.97& 3.05&2.74&4.61&4.32&3.91&3.81\\
				%& \ml{1}{Comp. [cm]$\downarrow$} &5.56&3.48&6.90&\textbf{3.31}&\textbf{3.46}&5.16&5.40&5.01&4.79\\
				%& Comp. Ratio [\%]$\uparrow$ &73.35&82.86&74.23&\textbf{82.56}&\textbf{86.19}&75.76&72.63&76.61&78.00\\
				%\midrule
				
				%\multirow{3}{*}{\rotatebox[origin=c]{90}{HI-SLAM}} 
				%%& \ml{1}{Depth L1 [cm]$\downarrow$} & 3.61& 2.07& 4.63& 3.66& 1.91& 3.39& 5.07& 4.68& \textbf{3.63}\\
				%& \ml{1}{Acc. [cm]$\downarrow$} & \rd3.33 &\rd3.50 &\fs{3.11} &3.77 &2.46 &4.86 &3.92 &3.53&{3.56}\\
				%& \ml{1}{Comp. [cm]$\downarrow$} & 3.29 &\textbf{3.20} &\textbf{3.39} &3.65 &3.61 &\textbf{3.68} &4.13 &\textbf{3.82}&\textbf{3.60}\\
				%& Comp. Ratio [\%]$\uparrow$ & 86.35 &\textbf{85.79} &82.95 &80.72 &82.35 &\textbf{82.89} &80.29 &82.29&\textbf{82.95}\\
				%\midrule
				
				\multirow{3}{*}{\rotatebox[origin=c]{90}{\textbf{Ours}}} 
				& \ml{1}{Acc. [cm]$\downarrow$} &\nd3.61&\nd4.02&\rd 5.53& \fs2.71&\fs2.17&\nd 4.09&\nd4.23& \nd 3.69&\nd 3.76\\
				& \ml{1}{Comp. [cm]$\downarrow$} &\rd 6.98&6.76&\rd 12.24&\nd 6.46&\nd 5.59&\rd 10.31&\nd7.53& 10.46&\rd8.29\\
				& Comp. Ratio [\%]$\uparrow$ &\nd 74.06&\rd 72.59&\nd 63.85&\fs77.80&\nd75.26&\rd65.56&\nd68.89& \nd69.10&\nd70.89\\
				\bottomrule
		\end{tabular} }

		\label{tab:replica_rec}
	\end{table}

	\begin{table}[htbp]
		\caption{\textbf{Novel View Synthesis Evaluation on Replica Dataset.} Best results are highlighted as \colorbox{colorFst}{\bf first}, \colorbox{colorSnd}{second}, and \colorbox{colorTrd}{third}.
		}
		\centering
		\footnotesize
		\setlength{\tabcolsep}{2.2pt}          %
		\renewcommand{\arraystretch}{1.1}      %
		\resizebox{.48\textwidth}{!}{
			\begin{tabular}{llccccccccc}
				\toprule
				& & rm-0 & rm-1 & rm-2 & off-0 & off-1 & off-2 & off-3 & off-4 & Avg.\\
				\midrule    
				\rowcolor{colorSep}
				\ml{3}{\textit{\textbf{RGB-D input}}} & \mc{8}{}\\
				\multirow{3}{*}{\rotatebox[origin=c]{90}{NICE-S.}}
				& PSNR $\uparrow$ 
				&  22.12 &  22.47 &  24.52 &  29.07 &  30.34 & 19.66 &  22.23 &  24.94 &  24.42
				\\
				& SSIM $\uparrow$ 
				& 0.689 &  0.757 &  0.814 & 0.874 &  0.886 &  0.797 & 0.801 &  0.856 &  0.809
				\\
				& LPIPS $\downarrow$ 
				& 0.330 & 0.271 &  0.208 &  0.229 &  0.181 &  0.235 &  0.209 &  0.198 &  0.233
				\\
				\midrule
				\multirow{3}{*}{\rotatebox[origin=c]{90}{Vox-F.}} 
				& PSNR $\uparrow$ 
				&  22.39 &  22.36 &  23.92 &  27.79 &  29.83 &  20.33 &  23.47 &  25.21 &  24.41
				\\
				& SSIM $\uparrow$ 
				& 0.683 & 0.751 & 0.798 &  0.857 &  0.876 & 0.794 &  0.803 &  0.847 & 0.801
				\\
				& LPIPS $\downarrow$ 
				&  0.303 &  0.269 & 0.234 & 0.241 &  0.184 & 0.243 &  0.213 &  0.199 &  0.236
				\\
				\midrule
				%				\multirow{4}{*}{\rotatebox[origin=c]{90}{NSLF-OL}} 
				%				&PSNR $\uparrow$&
				%				23.24 &25.68&24.88&28.59&26.70&21.10&21.89&25.74&\\
				%				&SSIM $\uparrow$ & 0.816&0.883&0.888&0.913 &0.879&0.863&0.847&0.893& \\
				%				&LPIPS$_{alex}$ $\downarrow$& 0.371&0.308&0.330&0.371 &0.497&0.362&0.368&0.401&\\
				%				&LPIPS$_{vgg}$$ \downarrow$& 0.417&0.415&0.422&0.395 &0.446&0.421&0.416&0.425&\\
				%				\midrule

				\multirow{3}{*}{\rotatebox[origin=c]{90}{\textbf{Ours}}} 
				&PSNR $\uparrow$&\fs 27.72& \fs28.86&\fs30.17&\fs32.59&\fs31.39&\fs24.44&\fs25.34&\fs30.47 &\fs 28.87\\
				&SSIM $\uparrow$&\fs0.850&\fs0.872&\fs0.899&\fs0.915&\fs0.892&\fs0.837&\fs0.867&\fs0.887 &\fs 0.877\\
				&LPIPS $\downarrow$&\fs0.124&\fs0.166&\fs0.143&\fs0.165&\fs0.262&\fs0.208&\fs0.132&\fs0.184 &\fs 0.173\\
				\midrule
				\rowcolor{colorSep}
				\ml{3}{\textit{\textbf{RGB input}}} & \mc{8}{}\\
				%				\multirow{3}{*}{\rotatebox[origin=c]{90}{COLMAP}}
				%				& PSNR $\uparrow$ 
				%				& 20.31 & 12.33 & 14.53 & 9.39 & 7.26 & 15.42 & 15.28 & 16.98 & 13.94 
				%				\\
				%				 & SSIM $\uparrow$ 
				%				&  0.710 & 0.607 & 0.771 & 0.744 & 0.676 & 0.767 & 0.754 & 0.765 & 0.724
				%				\\
				%				& LPIPS $\downarrow$ 
				%				& 0.332 & 0.531 & 0.318 & 0.377 & 0.306 & 0.341 & 0.330 & 0.290 & 0.353
				%				\\
				%				\midrule
				%	
				\multirow{3}{*}{\rotatebox[origin=c]{90}{NeRF-S.}} 
				& PSNR $\uparrow$  &16.45& 19.62 &21.17 &21.44 &20.86&  15.49 & 15.11&   18.96& 18.64\\
				& SSIM $\uparrow$ &0.576& 0.700 &0.754& 0.773 &0.747 &0.731 &0.688  &0.790 &0.720 \\
				& LPIPS $\downarrow$ &0.330 &\fs 0.177 &\fs 0.170& 0.335 &0.229 & 0.251& 0.282 &0.241&\rd 0.252 \\
				
				%				\midrule
				%				\multirow{3}{*}{\rotatebox[origin=c]{90}{DIM-S.$^*$}} 
				%				& PSNR $\uparrow$& 18.48& 26.19& 24.95 &30.16& 31.75& 21.36&  21.22& 23.65 &24.72 \\
				%				& SSIM $\uparrow$ & 0.622 &0.765 &0.788 &0.856 &0.882 &0.744& 0.751& 0.797& 0.776 \\
				%				& LPIPS $\downarrow$ &0.422 &0.283& 0.291& 0.234&   0.185&  0.304&  0.293 &  0.256 & 0.284			
				%				\\
				%	

				\midrule			
				\multirow{3}{*}{\rotatebox[origin=c]{90}{DROID-S.}} 
				& PSNR $\uparrow$ 
				& \rd 21.41 &  \nd 24.04 & \rd22.08 &\rd  23.59 &\rd 21.29 &  \rd 20.64 &\rd 20.22 & \rd20.22 & \rd 21.69
				\\
				& SSIM $\uparrow$ 
				& \nd 0.693 & \fs 0.786 & \nd 0.826 & \fs 0.868 & \fs 0.863 & \fs 0.828 & \nd 0.808 &\nd 0.819 & \nd 0.812
				\\
				& LPIPS $\downarrow$ 
				&  \nd0.329 & \rd 0.270 & \rd 0.228 & \nd 0.232 & \nd 0.207 & \nd 0.231 & \nd 0.234 &\nd 0.237 & \nd0.246
				\\
				\midrule
				
				\multirow{3}{*}{\rotatebox[origin=c]{90}{NICER-S.}} 
				
				& PSNR $\uparrow$ 
				&  \fs 25.33 & \rd 23.92 & \fs 26.12 & \fs 28.54 & \fs 25.86 &\fs  21.95 & \fs 26.13 &  \fs25.47 &\fs  25.41 \\
				& SSIM $\uparrow$ 
				& \fs 0.751 & \nd 0.771 &\nd  0.831 & \nd 0.866 & \nd0.852 & \nd 0.820 &  \fs0.856 & \fs 0.865 & \fs 0.827 \\
				& LPIPS $\downarrow$ 
				& \fs 0.250 &  \nd0.215 & \nd 0.176 &  \fs 0.172 &\fs  0.178 & \fs 0.195 &  \fs0.162 &  \fs0.177 & \fs 0.191 \\
				\midrule				
				\multirow{3}{*}{\rotatebox[origin=c]{90}{\textbf{Ours}}} 
				& PSNR $\uparrow$ 
				& \nd 23.12 & \fs 24.10 & \nd 24.11&  \nd 26.15 & \nd 24.68 & \nd 21.55&\nd 22.44 & \nd 24.42 &\nd 23.82\\
				& SSIM $\uparrow$ 
				& \rd 0.686  &  \rd 0.749 & \rd 0.765 &\rd  0.816 &\rd  0.852 & \rd 0.731 & \rd 0.739& \rd 0.789&\rd0.766\\
				& LPIPS $\downarrow$ 
				&  \rd 0.353 &  0.307 &  0.366 & \rd 0.338 & \rd 0.291 & \rd 0.381 &\rd 0.346&\rd 0.375&0.344\\
				%& LPIPS$_{vgg}$ $\downarrow$ 
				%&0.460& 0.464 & 0.468 & 0.355& 0.339 & 0.401&0.390& 0.414\\
				\bottomrule
		\end{tabular}}%
		
		\label{tab:replica_rendering}
	\end{table}

In the monocular setting with RGB input only, our model cannot outperform the NeRF-like SOTA NICER-SLAM.
Please find~\cref{fig:denseslam:replica_demo}, our model 1) does not support completion, 2) directly fuses depth images without optimizing the geometry that causes false surface rendering.
Which we consider the reason that cannot surpass the NICER-SLAM.
However, NICER-SLAM takes \textbf{$\sim10$ hours train per scene} excluding the preprocessing time that costs more, while our model takes only \textbf{minutes}.
But as a DBA-based method, compared to our closest model DROID-SLAM, our model still achieves much better scores to a level as useful as the NeRF-like method.
We believe this is due to the post-optimization nature of the global geometry for NeRF-like models.
While DBA-based methods rely more on a separate solution of the local inverse depth. There is no post-optimization involved.

In addition, the DBA-based method mainly relies on triangulation. 
However, the replica data set is originally acquired for the purpose of dense RGB-D reconstruction, but not for dense mono.
This means that the trajectory of the replica sequences inherently does not consider the effect of the interrelation of frames for depth estimation.

	\newcommand{\replicaImSize}{.19}
	\begin{figure*}[]
		\centering
		\setlength{\tabcolsep}{0.1em}
		\renewcommand{\arraystretch}{.1}
		\begin{tabular}{c  c | c c |  c }	
			
			\includegraphics[width=\replicaImSize\linewidth]{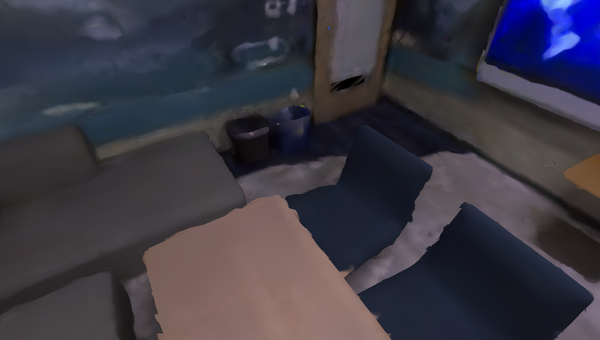}
			&\includegraphics[width=\replicaImSize\linewidth]{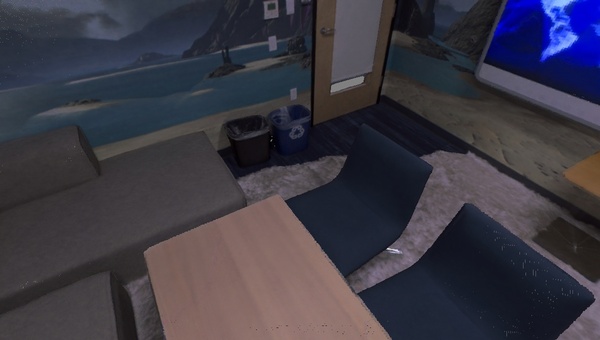}
			&\includegraphics[width=\replicaImSize\linewidth]{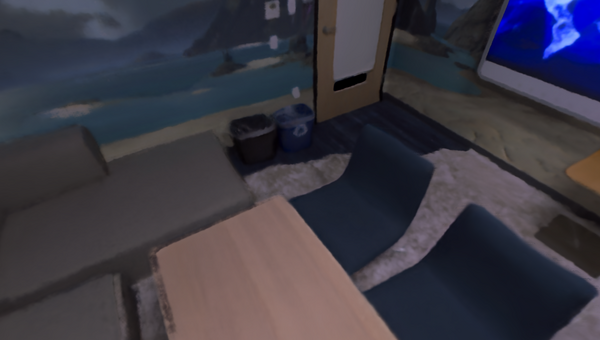}
			&\includegraphics[width=\replicaImSize\linewidth]{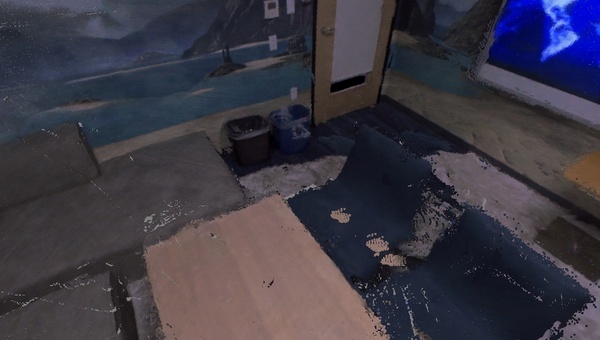}
			&\includegraphics[width=\replicaImSize\linewidth]{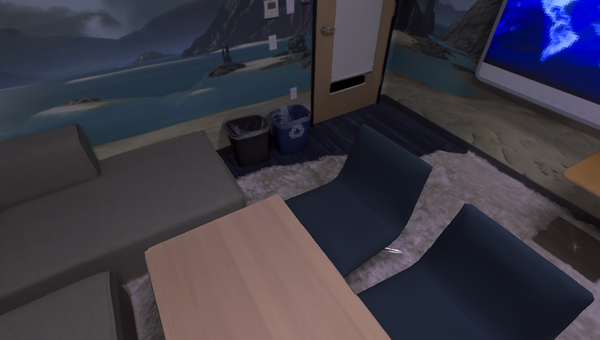}
			\\
			\hline
			\includegraphics[width=\replicaImSize\linewidth]{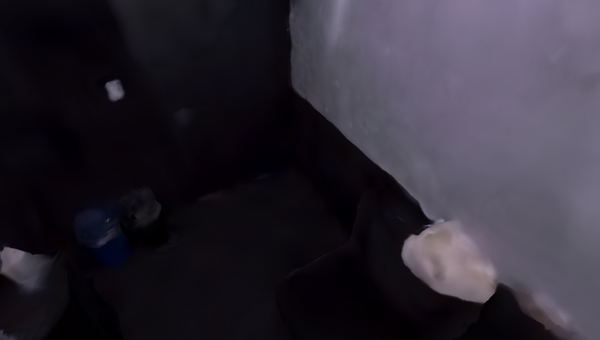}
			&\includegraphics[width=\replicaImSize\linewidth]{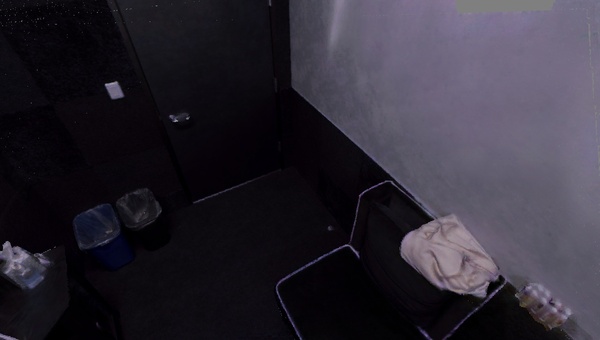}
			&\includegraphics[width=\replicaImSize\linewidth]{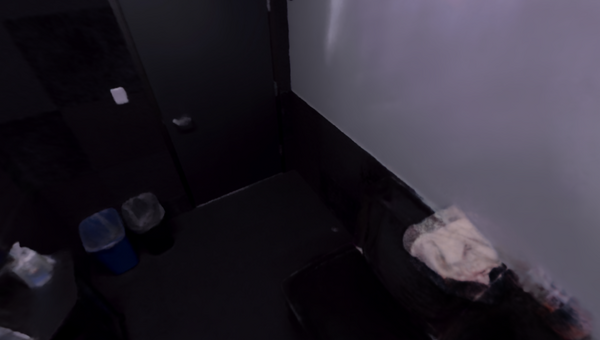}
			&\includegraphics[width=\replicaImSize\linewidth]{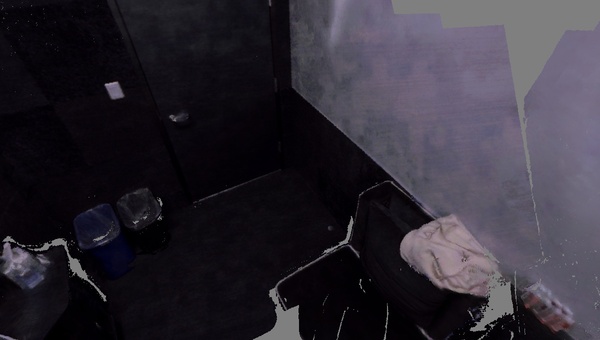}
			&\includegraphics[width=\replicaImSize\linewidth]{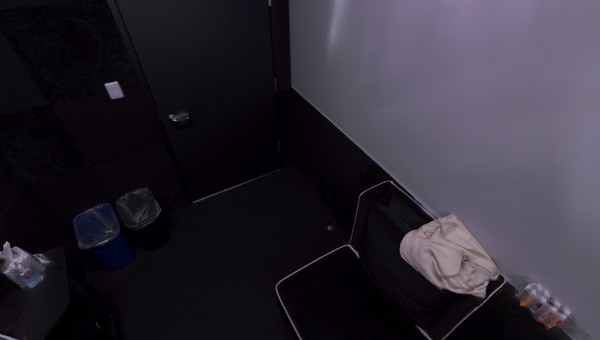}\\		
			
			\hline
			\includegraphics[width=\replicaImSize\linewidth]{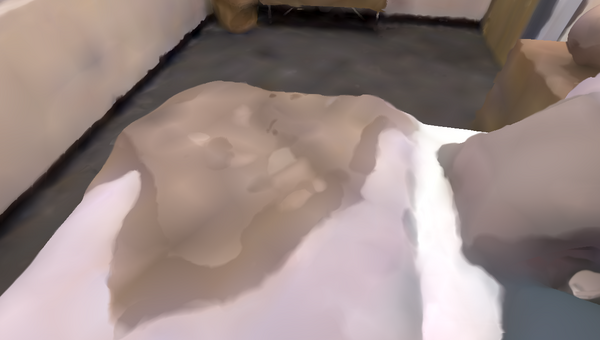}
			&\includegraphics[width=\replicaImSize\linewidth]{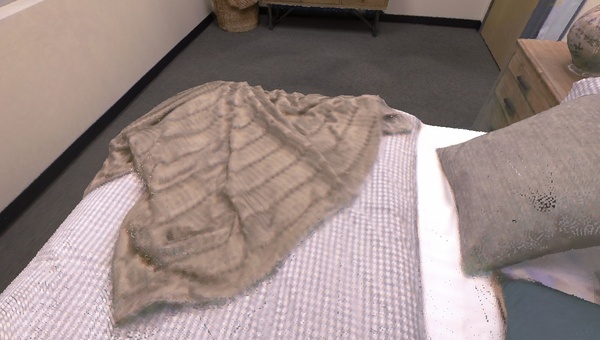}
			&\includegraphics[width=\replicaImSize\linewidth]{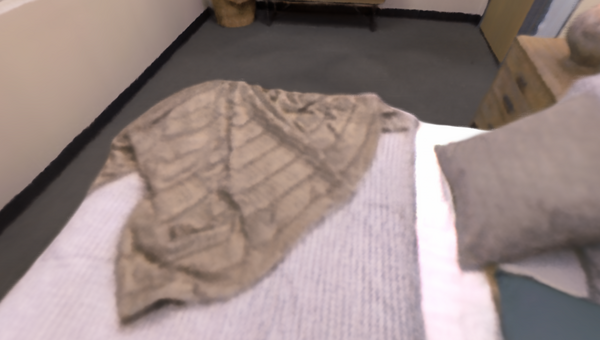}
			&\includegraphics[width=\replicaImSize\linewidth]{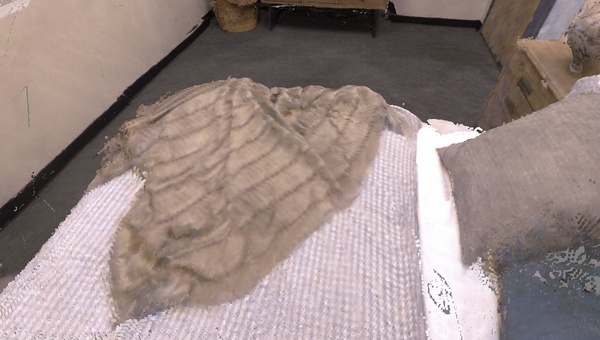}
			&\includegraphics[width=\replicaImSize\linewidth]{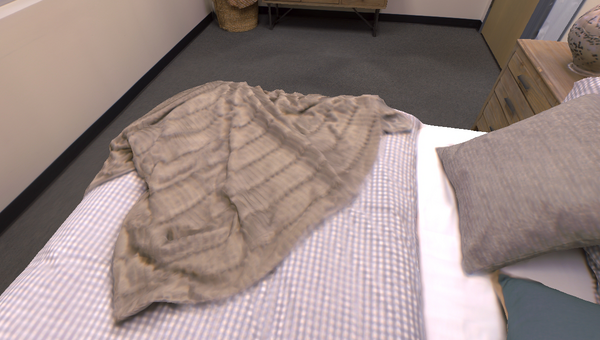}\\ 
			%		\hline\\
			%		\includegraphics[width=\replicaImSize\linewidth]{ims/SLAM/NICE/room1_1802}
			%&\includegraphics[width=\replicaImSize\linewidth]{ims/SLAM/Our_RGBD/room1_1802}
			%&\includegraphics[width=\replicaImSize\linewidth]{ims/SLAM/NICER/room1_1802}
			%&\includegraphics[width=\replicaImSize\linewidth]{ims/SLAM/Our_MONO/room1_1802}
			%&\includegraphics[width=\replicaImSize\linewidth]{ims/SLAM/GT/room1_1802}\\ \\
			%\hline\\
			
			{\large{NICE-SLAM}} & {\large{Ours}}&{\large{NICER-SLAM}} & {\large{Ours}}  &{\large{GT}}  \\ 
			\hline
			\multicolumn{2}{c|}{RGB-D Input} & \multicolumn{2}{c|}{RGB Input} &\\

		\end{tabular}
		%\captionof{figure}
		\caption{Qualitative evaluation on the Replica dataset for selected views.}
		\label{fig:denseslam:replica_demo}
                \vspace*{-4mm}
	\end{figure*}

To better illustrate this issue, we test on our own dataset for Dense Mono SLAM purpose in large scene, with mostly $xy$-directional motion.

The recent SOTA NeRF-like Dense SLAM, NICER-SLAM, highly focuses on the repeated encircling capturing of an object/scene.
This is because NICER-SLAM theoretically expires on large scale scene: 1) NICER-SLAM works in bounded model, the poses are required to bound the scene (from their open release); 2) NICER-SLAM's hashgrid resolution is preset to $2048$ with about 24 GB usage of GPU memory. But this is far too low for a large scene, and  it is hard to increase the resolution further.
(This is also the reason why the unbounded multi-hashgrid method, NSLF-OL~\cite{yuan2023online}, uses SLF instead of NeRF).
So as the scale increases, NICER-SLAM's tracking and then mapping will be dysfunctional.
Even the new branch, Gaussian blending based dense mono SLAM, MonoGS, is rarely tested at large scale.
While SLAM in robotics is usually excircle-like capturing the scene.

Therefore, to better explore the performance, our dataset mainly contains middle- and large-range scenes which are rarely involved by them. 
%\subsubsection{Tanks \& Temples Test}
%Tanks \& Temples dataset releases both captured RGB video and selected frames. 
%\TanksandTempleTable
%% Barn
%\TNTFigure

\subsection{Dense Mono SLAM Purpose dataset}
The above tests give an overview of the performance of the tracker and the reconstructor.

We collect this dataset because we find that the current RGB-D sequences are all intended for RGB-D tracking or dense RGB-D reconstruction in small scenes.
Due to the fact that RGB-D sequences contain real 3D metric depth, the sequence can simply move arbitrarily.
For example, in ScanNet and Replica there is a lot of almost purely rotational motion.

On the other hand, monocular sequences datasets are only used for tracking and SfM purposes because no depth ground truth is available.
For dense mono SLAM, however, the challenge goes beyond tracking and mapping. 
Unlike mono SLAM that only estimates sparse points with matching pairs,
dense mono SLAM requires a much larger region to cover.

Furthermore, in real-world robotics applications, the scene is usually captured on large scenes and non-repeat capturing which is rarely involved in recent dense mono SLAM SOTAs.

\begin{figure}[htbp!]
	\centering
	
	\subfloat[width=.5\text][Robotics hall (RGB-D)]{
		\centering
		\includegraphics[width=.5\linewidth]{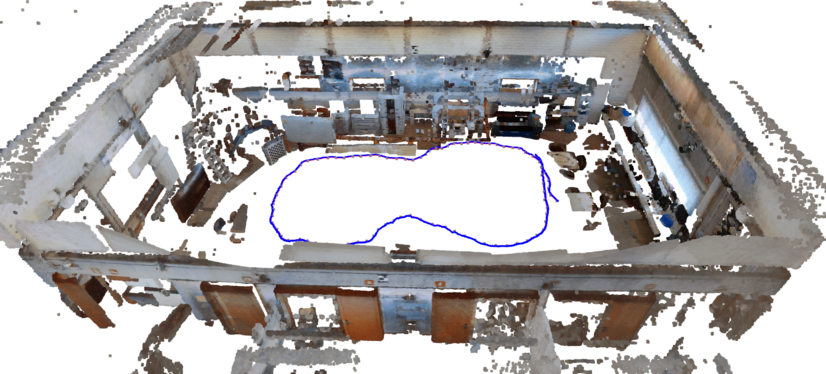}
		\label{fig:ourdata:rh}
	}
	\subfloat[width=.5\linewidth][Veitsh\"ochheim Palace (RGB-L)]{
		\centering
		\includegraphics[width=.5\linewidth]{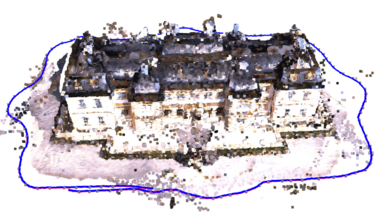}
		\label{fig:ourdata:castle}
	}\\	
	\subfloat[width=.3\linewidth][{\footnotesize Living room (RGB)}]{
		\centering
		\includegraphics[width=.3\linewidth]{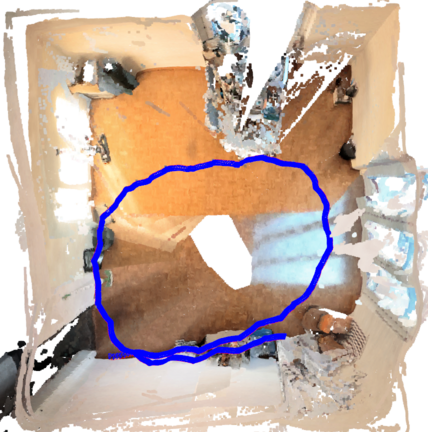}
		\label{fig:ourdata:room}
	}
	\subfloat[width=.5\linewidth][University campus (RGB-L)]{
		\centering
		\includegraphics[width=.5\linewidth]{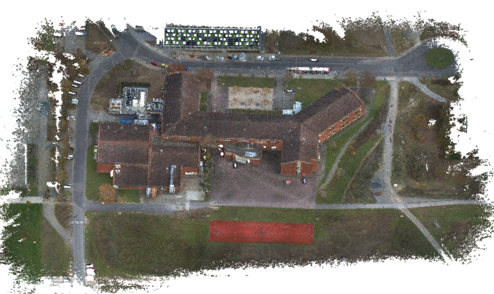}
		\label{fig:ourdata:zft}
	}\\	
	\caption{Our four datasets featuring different ranges.}
	\label{fig:ourdata}
\end{figure}

Therefore, we have not found any published dataset that is primarily for Dense Mono SLAM purposes in robotics.
To fill this gap, we hereby present the first large-scale RGB-D/L dataset for dense mono SLAM purposes:
%Considering that most dense methods require matching and triangulation, we capture the sequences with mainly $xy$-directional translation.
%Considering that most  requires matching and triangulation, the captures mainly contains rough-perpendicular translation to the view direction.
%Therefore, we capture four data for this purpose:
\begin{itemize}[]
	\item Mid-scale ($\sim20\times10m$) scene (RGB-D) as~\cref{fig:ourdata:rh},
	\item Large-scale ($\sim80\times50m$) scene (RGB-L) as~\cref{fig:ourdata:castle},
	\item Small-scale ($\sim5\times5m$) scene (RGB) as~\cref{fig:ourdata:room},
	\item Very-large-scale ($\sim300\times200m$) scene (RGB-L) as~\cref{fig:ourdata:zft}
\end{itemize}
We capture our mid-range data, Robotics Hall, with an Azure Kinect RGB-D camera. 
Because of the range limitation of Kinect Depth ($1m-5m$), we turn to RGB-L for large-range data, Veitsh\"ochheim Palace.
For small-range data, living room, we use cell phone for just quick quality demonstration.
To explore the extreme, we capture very large-range data, university campus (language center) with an aerial view RGB-L capture.

\cref{fig:ourdata:rh,fig:ourdata:castle,fig:ourdata:room}'s trajectory is a circle because the target is on the same level.
\cref{fig:ourdata:zft}'s drone trajectory is parallel to the ground but the aerial view also well fits the $xy$-directional translation.

In the previous test, we aligned the surface result with ICP, which becomes challenging in the large scene due to the large partial non-overlap.
To standardize the comparison, in our dataset, we use the alignment matrix from trajectory comparison to transform the result reconstruction to ground truth. 

\subsubsection{Mid-scale data}
\label{sec:exp:mid-scale}

	\begin{table}[htbp!]
		\caption{Test on robotics hall of our dataset.}
		\centering
		\begin{adjustbox}{max width=.5\textwidth}
			\begin{tabular}{lcccc}
				\toprule
				&NICER. &MonoGS &Ours &Ours (RGB-D)\\
				\midrule
				%\multirow{7}{*}{Robotics Hall}
				ATE [m] $\downarrow$  &2.44&2.17 &\bf0.59&\bf 0.22\\ 
				Acc. [m]$\downarrow$&1.009&0.724&\bf0.360&\bf 0.138\\
				Comp. [m]$\downarrow$&1.246&0.837&\bf0.368&\bf 0.141\\
				Comp. Ratio [\%]$\uparrow$&33.11&49.58&\bf74.84&\bf 99.13\\ 
				PSNR $\uparrow$ &\bf19.20&-& 15.13& 14.65\\
				SSIM $\uparrow$ &\bf0.690&-&0.442&  0.468\\
				LPIPS$_{alex}$ $\downarrow$&\bf0.449&-&0.628& 0.615 \\
				Time  & 8 hours & 25 min. &10 min. & 5 min.\\
				%			\midrule 
				%			\multirow{7}{*}{Castle}
				%			&ATE $\downarrow$ \\ 
				%			&Acc. [cm]$\downarrow$\\
				%			&Comp. [cm]$\downarrow$\\
				%			&Comp. Ratio [\%]$\uparrow$\\
				%			&PSNR $\uparrow$\\
				%			&SSIM $\uparrow$  \\
				%			&LPIPS$_{alex}$ $\downarrow$ \\
				%			
				\bottomrule
			\end{tabular}
		\end{adjustbox}
		\label{tab:SLAM:robotics_hall}
		
	\end{table}

\begin{figure}[htbp!]
	\centering
	\subfloat[width=.5\text][NICER-SLAM]{
		\centering
		\includegraphics[width=.5\linewidth]{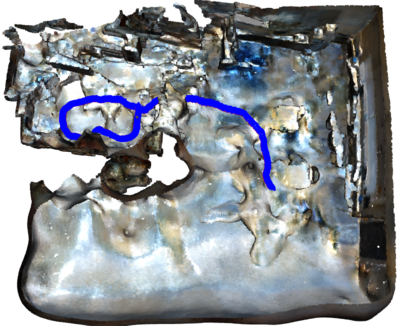}
	}
	\subfloat[width=.5\text][MonoGS]{
		\centering
		\includegraphics[width=.5\linewidth]{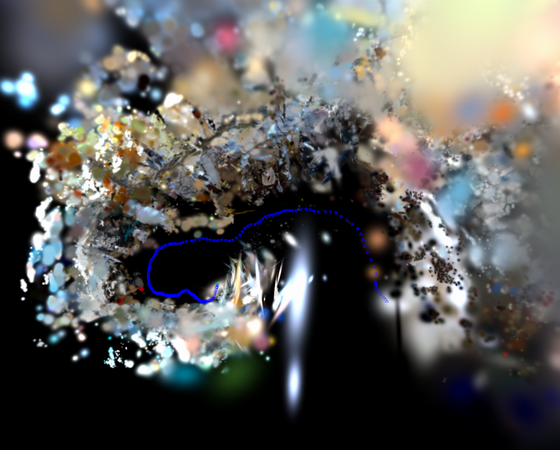}
	}\\
	\subfloat[width=.5\text][Ours (RGB)]{
		\centering
		\includegraphics[width=.5\linewidth]{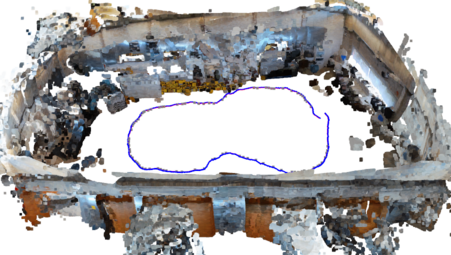}
	}
	\subfloat[width=.5\text][Ours (RGB-D)]{
		\centering
		\includegraphics[width=.5\linewidth]{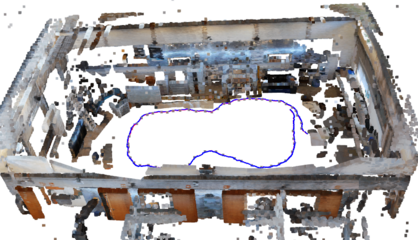}
	}	
	\caption{Test result (3D maps) on our mid-range scene, i.e., robotics hall. Trajectories are plotted blue.}
	\label{fig:robotics_hall}
\end{figure}
For a handheld camera, since our task is mainly with scene modeling, our facing direction of the camera is inside-out.
Therefore, our capture scheme should follow a certain rule, as shown in~\cref{fig:ourdata}:
\begin{itemize}[]
	\item move in circle in the scene
	\item viewing direction is perpendicular to the moving direction.
\end{itemize}

%\begin{figure}
%	\centering
%	\includegraphics[width=1\linewidth]{ims/dataset/ZfT_pcd}
%	\caption{ZfT dataset.}
%	\label{fig:phone}
%\end{figure}

%Then in this test, further compare with our most closed, open-released Dense Mono-SLAM, GO-SLAM on our captured dataset.

%Because NICER-SLAM's and MonoGS's result cannot be directly aligned with ICP, we thus use their trajectory's alignment matrix instead to scale and then use Non-scaled ICP to align to the ground truth for evaluation.
%\OurTestFigure

Please find~\cref{tab:SLAM:robotics_hall} the result on robotics hall.
SceneFactory can easily achieve best tracking and reconstruction than two most new Dense Mono SLAM SOTA, NICER-SLAM and MonoGS.
Moreover, we find NICER-SLAM that works better on image rendering even with much worse tracking and inconsistent scale.
To reveal the truth, we visualize the trajectory with reconstruction.
Please find~\cref{fig:robotics_hall}, both NICER-SLAM and MonoGS severely suffer severely from scale.
This further confirms our speculation: The tightly coupled Dense Mono SLAM SOTAs (NICER-SLAM, MonoGS) models \textit{attend to one thing and lose sight of another}.

Also, the open version of NICER-SLAM requires preprocessing of the data sequence (COLMAP and more), which takes many more hours than the $8$h in the table.
MonoGS is relatively faster, but the result can hardly be viewed in any non-trained view.
Which is, you can only view the mid-range and large-range (next test) in the same trained view.
As claimed in NSLF-OL~\cite{yuan2023online}, this is not usable in SLAM because SLAM captures contain very sparse view directions.

\subsubsection{Large-scale data}

	\begin{table}[htbp!]
		\caption{Test on schloss of our dataset.}
		\centering
		\begin{adjustbox}{max width=.5\textwidth}
			\begin{tabular}{lcccc}
				\toprule
				&NICER. &MonoGS &Ours &Ours (RGB-L)\\
				\midrule
				%\multirow{7}{*}{Robotics Hall}
				ATE [m] $\downarrow$  &-&12.90&0.31&0.37\\ 
				Acc. [m]$\downarrow$&- &7.050&5.074&0.423\\
				Comp. [m]$\downarrow$&-&19.956&0.215&0.321\\
				Comp. Ratio [\%]$\uparrow$&-&2.6435&95.67&86.54\\
				Time  & 4 (28\%) hours  & 2 hours  & 13 min. & 9 min.\\
				
				\bottomrule
			\end{tabular}
		\end{adjustbox}
		\label{tab:SLAM:castle}
		
	\end{table}

\begin{figure}[htbp!]
	\centering
	
	\subfloat[width=.5\text][NICER-SLAM]{
		\centering
		{\raisebox{8mm}{\includegraphics[width=.5\linewidth]{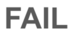}}}
	}
	\subfloat[width=.5\text][MonoGS]{
		\centering
		\includegraphics[width=.5\linewidth]{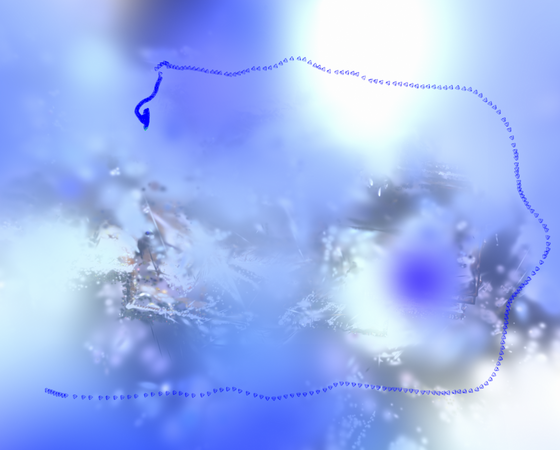}
	}	\\
	\subfloat[width=.5\text][Ours (RGB)]{
		\centering
		\includegraphics[width=.5\linewidth]{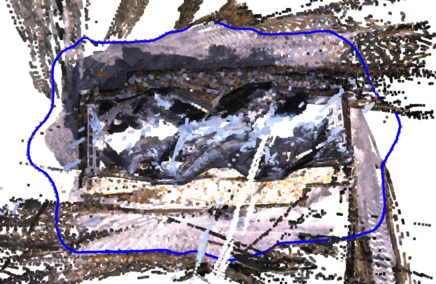}
	}
	\subfloat[width=.5\text][Ours (RGB-L)]{
		\centering
		\includegraphics[width=.5\linewidth]{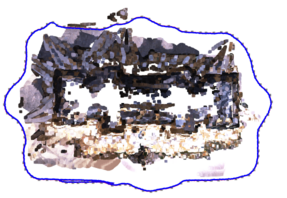}
		\label{fig:castle:rgbl}
	}	
	\caption{Test result (3D map) on our large-range scene, i.e., Veitsh\"ochheim Palace.}
	\label{fig:castle}
\end{figure}
For even larger scenes, RGB-D cameras are not applicable, therefore, we turn to Livox LiDAR to provide the metric depth.

%FIXED: Suestion: Shall we cite the LowCost3D paper here?
%A: maybe not in this script?
We use our handheld RGB-L camera to capture scenes.
Since the range is large, our target should be captured outside-in following the same rule as~\cref{sec:exp:mid-scale}.

For the larger range scene, from~\cref{tab:SLAM:castle}, NICER-SLAM fails at $28\%$, MonoGS shows very large tracking error. 
This is also revealed by the~\cref{fig:castle}. MonoGS losses the scale in the middle.
In addition the better performance on large-range, the time cost should also be considered.
NICER-SLAM takes $4$ hours for only $28\%$ without the counting of preprocessing. 
MonoGS takes $2$ hours for the scene.
%FIXED: What is hours? More precise here!
% A: for a scene
While ours are still in an acceptable range.

\subsubsection{Aerial-view data test}
To better demonstrate the potential, we further expand the test scale to very large.
This is too large to capture with simple handheld equipment. 
So we turn to an airborne device.

The data is collected with a custom-built UAV (DJI Matrice 300 RTK) equipped with a LUCID camera and a co-calibrated OUSTER OS1 laser scanner, flying over a university building that is a former high school.
\begin{figure*}[htbp!]
	\centering
	\centering
	\subfloat[width=.5\textwidth][Aerial image, oblique view ($\sim300m\times200m$)]{
		\includegraphics[width=.5\textwidth]{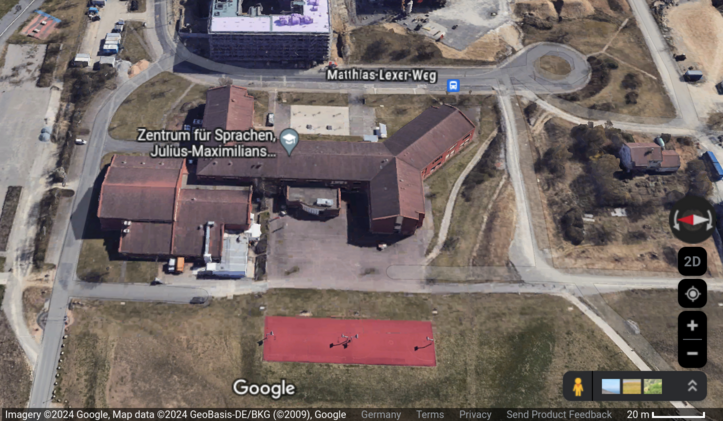}
	}
	\subfloat[width=.5\textwidth][Our result (RGB)]{
		\includegraphics[width=.5\textwidth]{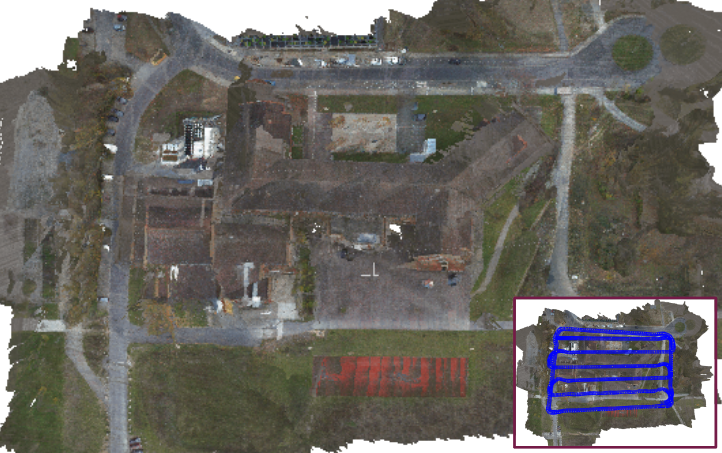}
	}
	
	\caption{SceneFactory's Dense Mono SLAM on our very-large range scene, i.e., University campus.}
	\vspace{-.4cm}
	\label{fig:zft}
\end{figure*}

The Dense SLAM result is demonstrated in~\cref{fig:zft}. Without using of metric depth. 
SceneFactory provides a high-quality reconstruction of this very large scene.

%\subsubsection{Phone data}
%Camera data is captured with a Xiaomi cell phone camera.

% unlike NeRF, SLF strongly relies on metric/predicted depth and make no optimization on geometry.
%\subsection{Evaluation of Depth Estimation}
% follow unimatch's depth test, our model first predict depth, then use pose to recover the scale
%Then we evaluate the estimation of depth.
%RGB images and poses are provided for this application.
%Because our depth model doesnot requires poses, the estimated depth requires a depth recovery to global.
%To have a fair comparison to baselines, different from SceneFactory's Mono-SLAM that use keypoints, we use GT poses to recover the scale.

%\newpage

%\input{tex/additional}

\subsection{Exclusive Applications}
In addition to the previously demonstrated hot tasks, SceneFactory also supports other usages.

\subsubsection{LiDAR Completion}
\begin{figure}[htbp!]
	\centering
	\subfloat[width=.5\text][RGB]{
		\centering
		\includegraphics[width=.5\linewidth]{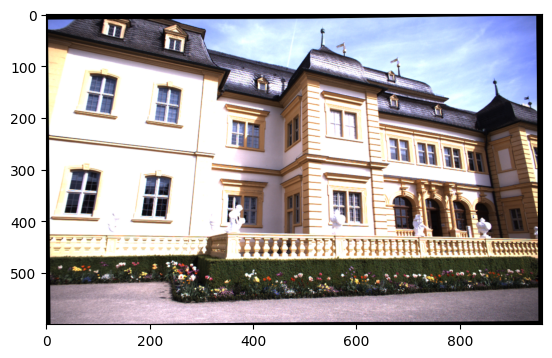}
	}
	\subfloat[width=.5\linewidth][Livox depth]{
		\centering
		\includegraphics[width=.5\linewidth]{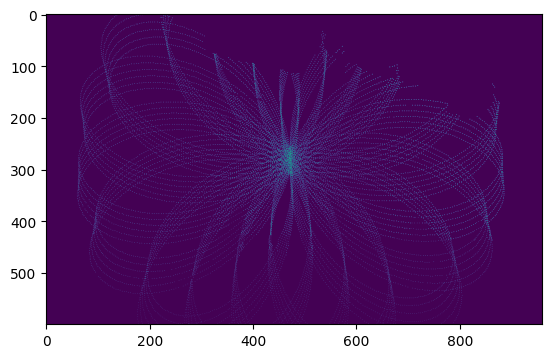}
	}		\\	\vspace{-.2cm}
	\subfloat[width=.5\text][Completed depth]{
		\centering
		\includegraphics[width=.5\linewidth]{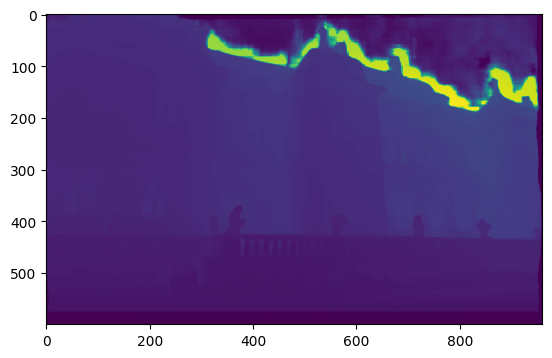}
	}
	\subfloat[width=.5\linewidth][Depth variance]{
		\centering
		\includegraphics[width=.5\linewidth]{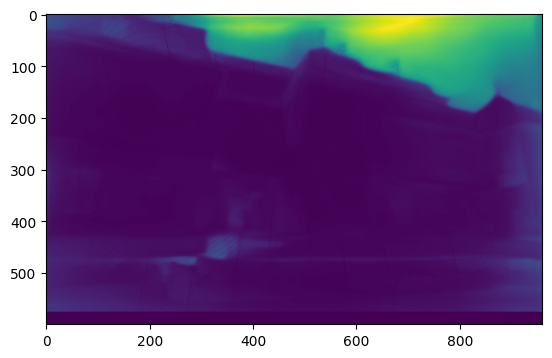}
	}			%\vspace{-.2cm}
	\caption{Example for the completed LiDAR depth with ScaleCov.}
	\vspace{-.4cm}
	\label{fig:completion}
\end{figure}

Our ScaleCov model supports completion of metric LiDAR depth.
Given an RGB and sparse depth image from Livox, ScaleCov regresses the full depth and variance to the client.
We provide an example in~\cref{fig:completion}.

This technique is also utilized in previous test~\cref{fig:castle:rgbl}.

\subsubsection{Dense Depth-only SLAM}
Color sensors tend to be more sensitive to the environment.
If a color sensor fails, e.g. due to poor lighting conditions or at night, the depth sensor could play a role instead.

SceneFactory implements depth-only SLAM under the factory lines of depth-flexion and RGB-D SLAM.
This first transforms the depth image into a trackable RGB image with depth-flexion. 
Then the application is used in the same way as our RGB-D SLAM application.
\begin{figure}[htbp!]
	\centering
	\includegraphics[width=.8\linewidth]{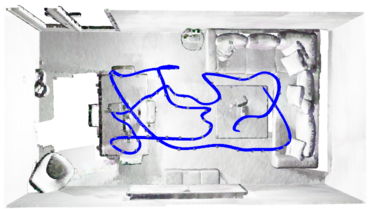}
	\caption{Depth-only SLAM result on the Replica dataset. Shown without the ceiling for better visualization.}
	\vspace{-.4cm}
	\label{fig:exp_depthonly}
\end{figure}
Please find~\cref{fig:exp_depthonly} our demo of office2 in Replica dataset, with depth only as input,
The reconstruction result is of high quality and smoothness.

\section{Conclusion}
In this paper, we have introduced a workflow-centric and unified framework for incremental scene modeling, called SceneFactory.
Following the structure of a ``factory'', we have designed ``assembly lines'' for a wide range of applications, to achieve high flexibility, adaptability and production diversification.
%Besides, this design philosophy can well control the resources and mitigate the risk of fault on certain line.

In addition, within the framework, we propose an unposed \& uncalibrated multi-view depth estimation model for highly flexible use.
We introduce a surface accessible light field design along with an improved point rasterization to enable surface query for the first time.

Our experiments show that SceneFactory is highly competitive or even better than compact SOTAs in such applications.
The high quality and broad applicability of the design further enhances the progress of our design.

Needless to say, a lot of work remains to be done.
In future work, we will focus on adding more applications to the production lines, such as deformable reconstruction, active SLAM, or scene understanding~\cite{ming2024benchmarking}.

{\small
  \bibliographystyle{IEEEtran.bst}
  \bibliography{ref}
}
\begin{appendices}
\section*{Supplementary}

\subsection{Good Neighbor Selection}
The dense bundle adjustment relies on the selection of good neighbors.
We select neighbor frames by filtering the relative poses according to~\cref{algo:depth_frame_selection}.
\begin{algorithm}[htbp]
	\SetAlgoLined
	\DontPrintSemicolon
	%\KwIn{$(\V I_{rgb}$, $\V I_d$, $\V G$, $i )$}%\Comment*[r]{List of Sensitive Terms}}    
%\KwOut{$ S^{*} $}% \Comment*[r]{Negation Excluded List}}
%	\SetKwFunction{FMain}{CheckGoodNeighborhood}
%	\SetKwProg{Fn}{Function}{:}{}
%	\Fn{\FMain{$[\cdots,\V G_i,\cdots])$}}{
%		\tcp{fetch good neighbors}
%		$ids_{pre}\leftarrow FindGoodNeighbors([\cdots,\V G_{i-1}],\V G_i);$\\
%		$ids_{post}\leftarrow FindGoodNeighbors([\V G_{i+1},\cdots],\V G_i);$
%		{
	%			
	%			\eIf{$ids_{pre}\neq \emptyset$ \textbf{and} $ids_{post}\neq \emptyset$}
	%			{
		%				{\Return $ids_{prev}\cup ids_{post};$}
		%			}
	%			{
		%				{\Return $\emptyset;$}
		%			}
	%			
	%		}
%	}
%	\textbf{End Function}
%	\\
%	\hrulefill
%	\\ 
\SetKwFunction{FMain}{FindGoodNeighbors}
\SetKwProg{Fn}{Function}{:}{}
\Fn{\FMain{$[\V G_{nb,1},\cdots],\V G_{i}),\tau_{nb}=2$}}{
\tcp{select frame by relative pose}
$ids=[];Ts=[]$\\
\For{$\V G_j\in [\V G_{nb,1},\cdots]$}{
	\tcp{transformation from i to j}
	$\V T_{ji}=\V G_i^{-1}\V G_j;$\\
	$\V R_{ji}, \V t_{ji}=\V T_{ji};$\\
	$l_{baseline} = || \V t_{ji}||_2;$\\
	$\theta_{facing} = \arccos([0,0,1]\V R_{ji}[0,0,1]^T);$\\
	\If{$l_{baseline}>\tau_{baseline}$ \textbf{and} $\theta_{facing}>\tau_{facing}$ }{
		$ids.append(j);$
		$Ts.append(\V T_{ji})$
	}
}
\eIf{$len(ids)<\tau_{nb}$}
{
	\Return $\emptyset;$
}
{
	\tcp{return the best $\tau_{nb}$ XY directional baselines}
	\Return $SortByBaselineXY(ids,Ts)[:\tau_{nb}];$
}

}

\textbf{End Function}
\caption{Depth neighbor frame selection.}
\label{algo:depth_frame_selection}
\end{algorithm}

%\input{tex/experiment/ablation}
%\subsection{Ablation Study}
%SceneFactory designs each model independently, therefore, we can make ablation study under each block.

\subsection{Improved Point Rasterization Ablation Test}
\label{sec:suppl:ipr}
\paragraph{Effect of Adaptive Radius}

\begin{figure}[htbp]
	\centering
	\subfloat[width=.48\text][Pytorch3d Rast. ($\sim0.57s$)]{
		\centering
		\includegraphics[width=.48\linewidth]{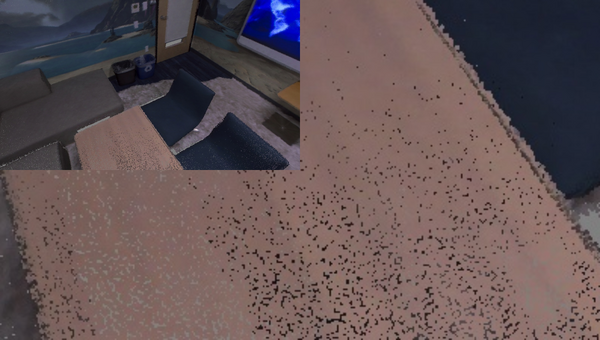}
	}
	\subfloat[width=.48\linewidth][Ours Rast. ($\sim0.025s$)]{
		\centering
		\includegraphics[width=.48\linewidth]{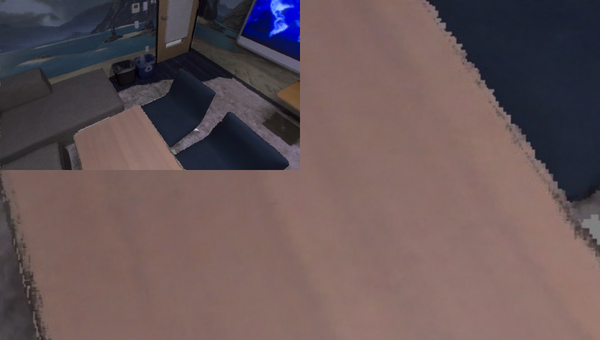}
	}		
	\caption{The effect of the Adaptive Radius (left pytorch3d) and our implementation (right) in an arbitrary view.}
	\label{fig:IPR:adaptive}
\end{figure}

Please note~\cref{fig:IPR:adaptive}, there is a black hole on the desktop. 
This is due to the fixed radius of pytorch3d's point rasterization in NDC space. 
Which would be much more severe if the camera goes even closer. 
While our IPR does not have this problem.

\begin{figure}[htbp]
	\centering
	\subfloat[width=.48\linewidth][w/o Layer filter]{
		\centering
		\includegraphics[width=.48\linewidth]{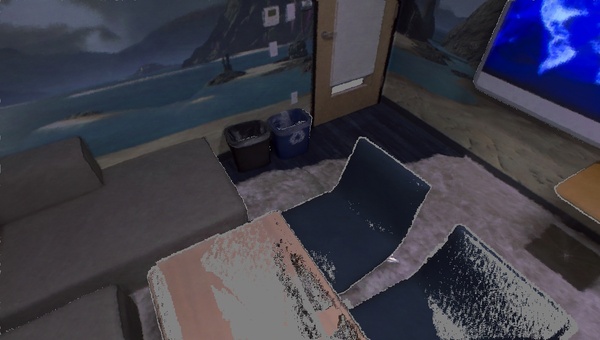}
	}
	\subfloat[width=.48\linewidth][w/ Layer Filter]{
		\centering
		\includegraphics[width=.48\linewidth]{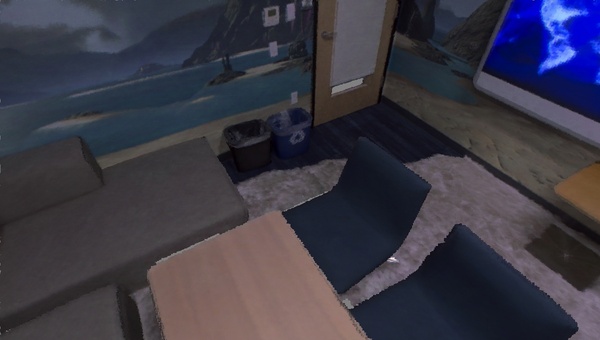}
	}		
	\caption{The application of the Layer Filter (right) removed blank space (left).}
	\label{fig:IPR:layer}
\end{figure}

\paragraph{Effect of First-layer filter}
Apart from the adaptive radius, our IPR also gets the surface with a layer filter as in~\cref{fig:IPR:layer}.
Without the layer filter, the resulting surface points will be in the middle of multiple surface layers, resulting in empty space during color rendering.

%\subsubsection{Depth Estimation Block Ablation Test}
%\paragraph{Effect of Depth Frame Selections and Neighborhood Selections}
%\paragraph{Effect of Static Check}
%\paragraph{Effect of Mono-depth Prior}
%\subsubsection{Scene Modeling Block Ablation Test}
%\paragraph{Effect of NPS Resolutions and Levels}

\subsection{Description of the Sensor Systems}
\label{sec:sup:sensor}
For collecting our own RGB-X dataset, described in Sec.~\ref{sec:datasets}, two commercial sensors are employed: the Microsoft Azure Kinect RGB-D sensor and the built-in camera of a Xiaomi Redmi smartphone. Furthermore, two custom-built sensor systems are used for capturing the datasets:

\paragraph{Handheld mapping system}
Our custom-built handheld mapping device is show in~\cref{fig:sensors}(a).
It is based on a commercial camera rig with the sensors mounted to the top bar.
The sensors are a Livox AVIA Lidar and two IDS U3-30C0CP global-shutter cameras with a Sony IMX392 \unit[2.35]{MPixel} RGB CMOS sensor.
The Lidar sensor has a Field of View (FoV) of $70.4^\circ \times 77.2^\circ$.
The cameras are equipped with \unit[4]{mm} lenses, which results in a similar FoV of $77.3^\circ \times 61.9^\circ$.
Below the sensors a 3D printed enclosure is mounted with an embedded PC for data recording and power supply electronics.
The cameras and lidar are co-calibrated using a calibration board.

From the system we extract a synchronized RGB-L data stream with Lidar scans and camera images with \unit[10]{Hz}.
While the system features stereo cameras, in this work we focus on mono SLAM.
Therefore, only the camera closest to the Lidar sensor is used.
The data is captured with the system handheld and an average walking speed of \unit[0.84]{$\frac{m}{s}$}.
During data collection the sensors always points towards the captured object.
The trajectory around Veitsh\"ochheim Palace is \unit[251]{m} long and was captured in \unit[5]{min}.
It consists of 3000 Lidar scans and 3000 RGB images.

\paragraph{UAV mapping system}
The UAV mapping system is based on the DJI Matrice 300 RTK and is shown in~\cref{fig:sensors}(b).
It carries an Ouster OS1-128 Lidar and a LUCID Vision Labs Phoenix PHX032S-CC global-shutter camera with a Sony IMX265 \unit[3.2]{MPixel} RGB CMOS sensor.
The Lidar sensor has a FoV of $360^\circ \times 45^\circ$.
The camera is used with a \unit[4.5]{mm} fixed lens with a horizontal FoV of $84.7^\circ$.
The co-calibrated camera is mounted directly on top of the Lidar sensor.
For the data collection an embedded PC is mounted on top of the UAV.
The aerial imagery is captured with an oblique angle and a lawnmower pattern flight trajectory.

%\begin{figure}
%	\centering
%	\begin{minipage}{\linewidth}
%	\subfloat[width=0.45\linewidth][Handheld RGB-L]{
%		\centering
%		\includegraphics[width=0.45\linewidth]{ims/sensors/handheld}
%	}
%	\hfill
%	\subfloat[width=0.5\linewidth][UAV RGB-L]{
%		\centering
%		\includegraphics[width=0.5\linewidth]{ims/sensors/drone.png}
%	}
%	\end{minipage}
%	
%	\caption{Handheld and UAV mapping systems used for data collection.}
%	\label{fig:custom_sensor_systems}
%\end{figure}
\end{appendices}
\end{document}